\documentclass[12pt,letterpaper]{article}
\usepackage[top=0.85in,left=1in,right=1in,footskip=0.75in,marginparwidth=0in]{geometry}
\pdfoutput=1

\usepackage[utf8]{inputenc}

\usepackage{cite}

\usepackage{nameref,hyperref}

\usepackage[right]{lineno}

\usepackage{mathtools}

\usepackage{graphicx}
\usepackage[font=scriptsize,labelfont=bf]{caption}
\usepackage{subcaption}
\usepackage{algpseudocode}
\usepackage{algorithm}
\usepackage{multirow}
\usepackage{color}
\usepackage{enumitem}
\usepackage{amsfonts}
\usepackage{bm}
\usepackage{mathtools}

\usepackage{microtype}
\DisableLigatures[f]{encoding = *, family = * }

\raggedright
\usepackage{changepage}

\usepackage[aboveskip=1pt,labelfont=bf,labelsep=period,singlelinecheck=off]{caption}

\makeatletter
\renewcommand{\@biblabel}[1]{\quad#1.}
\makeatother

\usepackage{lastpage,fancyhdr,graphicx}
\usepackage{epstopdf}
\fancyheadoffset[L]{2.25in}
\fancyfootoffset[L]{2.25in}

\usepackage{color}

\definecolor{Gray}{gray}{.25}

\usepackage{graphicx}

\usepackage{sidecap}

\usepackage{wrapfig}
\usepackage[pscoord]{eso-pic}
\usepackage[fulladjust]{marginnote}
\reversemarginpar

\newcommand{\spatscale}{$\bm{S}_{\bm{x}}^{\varepsilon}$}
\newcommand{\mspatscale}{\bm{S}_{\bm{x}}^{\varepsilon}}
\newcommand{\mspatscalei}{\bm{S}_{\bm{x}_i}^{\varepsilon}}

\newcommand{\strainscale}{$\bm{S}^{\varepsilon}$}
\newcommand{\mstrainscale}{\bm{S}^{\varepsilon}}
\newcommand{\stressscale}{$\bm{S}^{\sigma}$}
\newcommand{\mstressscale}{\bm{S}^{\sigma}}
\newcommand{\strainvec}{$\bm{\varepsilon}({\bm{x}})$} 
\newcommand{\mstrainvec}{\bm{\varepsilon}({\bm{x}})} 
\newcommand{\stressvec}{$\bm{\sigma}({\bm{x}})$} 
\newcommand{\mstressvec}{\bm{\sigma}({\bm{x}})} 
\newcommand{\cann}{CaNNCM}
\newcommand{\canns}{CaNNCMs}

\newcommand{\SK}{S_{\bm{x}}^{\varepsilon_k}}
\newcommand{\SKi}{S_{\bm{x}_i}^{\varepsilon_k}}
\newcommand{\SJ}{S_{\bm{x}}^{\varepsilon_j}}
\newcommand{\SJi}{S_{\bm{x}_i}^{\varepsilon_j}}
\newcommand{\partialSK}{\frac{\partial}{\partial\SK}}

\DeclareMathOperator*{\argmin}{argmin}

\begin{document}
\vspace*{0.35in}

\begin{flushleft}
{\Large
\textbf\newline{Cartesian Neural Network Constitutive Models for Data-driven Elasticity Imaging}
}
\newline
\\
Cameron Hoerig\textsuperscript{1,2,*},
Jamshid Ghaboussi\textsuperscript{3},
Michael F. Insana\textsuperscript{1,2},
\\
\bigskip
\bf{1} Department of Bioengineering, University of Illinois at Urbana-Champaign, Urbana, IL 61801 USA
\\
\bf{2} Beckman Institute of Advanced Science and Technology, University of Illinois at Urbana-Champaign, Urbana, IL 61801 USA
\\
\bf{3} Department of Civil and Environmental Engineering, University of Illinois at Urbana-Champaign, Urbana, IL 61801 USA
\\
\bigskip
* hoerig2@illinois.edu

\end{flushleft}

\section*{Abstract}
Elasticity images map biomechanical properties of soft tissues to aid in the detection and diagnosis of pathological states. In particular, quasi-static ultrasonic (US) elastography techniques use force-displacement measurements acquired during an US scan to parameterize the spatio-temporal stress-strain behavior. Current methods use a model-based inverse approach to estimate the parameters associated with a chosen constitutive model. However, model-based methods rely on simplifying assumptions of tissue biomechanical properties, often limiting elastography to imaging one or two linear-elastic parameters.

We previously described a data-driven method for building neural network constitutive models (NNCMs) that learn stress-strain relationships from force-displacement data. Using measurements acquired on gelatin phantoms, we demonstrated the ability of NNCMs to characterize linear-elastic mechanical properties without an initial model assumption and thus circumvent the mathematical constraints typically encountered in classic model-based approaches to the inverse problem. While successful, we were required to use \textit{a priori} knowledge of the internal object shape to define the spatial distribution of regions exhibiting different material properties.

Here, we introduce Cartesian neural network constitutive models (\canns) that are capable of using data to model both linear-elastic mechanical properties and their distribution in space. We demonstrate the ability of \canns\ to capture arbitrary material property distributions using stress-strain data from simulated phantoms. Furthermore, we show that a trained \cann\ can be used to \textit{reconstruct} a Young's modulus image. \canns\ are an important step toward data-driven modeling and imaging the complex mechanical properties of soft tissues.

\nolinenumbers

\section{Introduction}
%
%
%
%

Elasticity imaging methods reconstruct a map of mechanical properties by observing tissue motion in response to a weak mechanical stimulus. For quasi-static ultrasonic elastography, measurement data are forces and displacements as an ultrasound (US) probe is slowly pressed into the tissue surface. Recorded displacements may include both probe motion and internal tissue deformation, the latter being estimated via speckle-tracking algorithms operating on pre- and post-deformation echo data (e.g.,\cite{lubinski1999,hashemi2017}). These time-varying force-displacement measurements are governed by the geometric and mechanical properties of the tissue and can provide diagnostic data relevant to cancer detection in the breast~\cite{evans2010}, liver~\cite{fujimoto2013,castera2005}, and prostate~\cite{ahmad2013}, identifying atherosclerotic plaques~\cite{widman2015}, or treatment monitoring during high-intensity focused ultrasound or RF ablation~\cite{vanvledder2010,souchon2003}.

Estimating mechanical properties from measurement data constitutes the inverse problem in elastography. The goal for quasi-static elastography can be stated simply: given a set of force-displacement estimates and overall object shape, reconstruct the spatial distribution of mechanical properties. Current solutions take a model-based (or knowledge-driven) approach, where the mechanical properties of tissues are defined by parameters of a constitutive model relating stresses and strains. This problem is ill-posed in part due to the presence of measurement noise and limited force-displacement sampling from which stress-strain behavior is determined. Some strain information can be computed as spatial derivatives of the displacements, but stresses cannot be calculated from force data without knowing the object geometry, boundary conditions, and material properties. 

Simplifying assumptions are adopted in model-based techniques to help overcome the ill-posed nature of the inverse problem. Most often the tissue is assumed to be linear-elastic, isotropic, (nearly) incompressible, and under small strain, limiting the parameter space to only the Young's modulus (or shear modulus). However, biological tissues are bi-phasic media exhibiting nonlinear and viscoelastic properties not summarized by a single linear parameter. Recent work in imaging nonlinear~\cite{goenezen2012,sridhar2017} and viscoelastic~\cite{insana2004,bayat2017} material properties are promising, but still rely on initial constitutive model assumptions. And while a chosen model may be appropriate for certain tissue types, it may be incorrect for other tissues in the same field of view. The problem is that the constitutive model sets the parameters to be estimated and if an inappropriate model is chosen, parametric errors are made, corrupting the final elastogram. Also, the most diagnostic model parameters for each disease state have yet to be determined. 

\begin{figure*}
	\centering
	\includegraphics[width=0.7\textwidth]{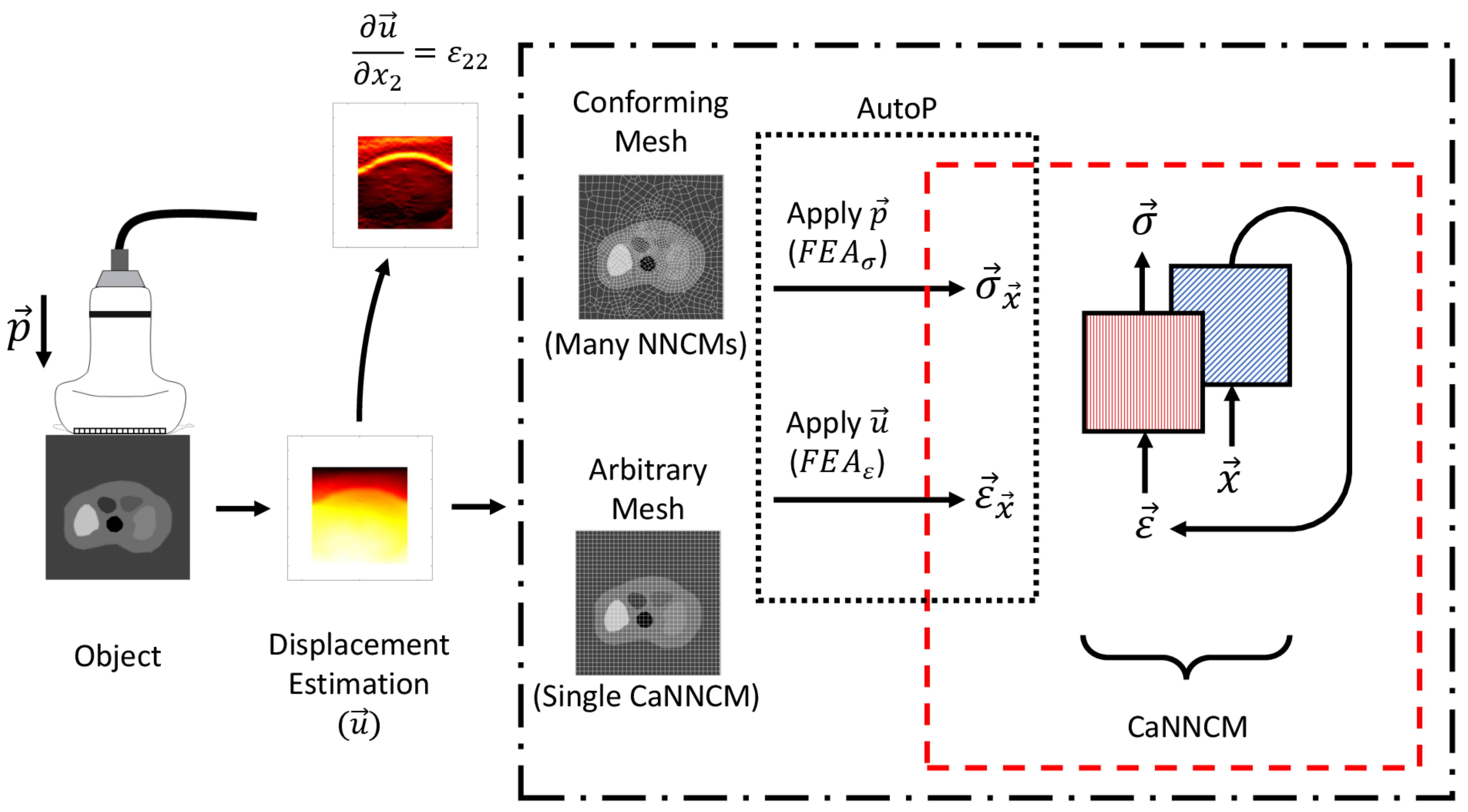}
	\caption{An US probe is slowly pressed into an object with force $\bm{p}$ while pre- and post-deformation RF frames are acquired. Speckle-tracking methods applied to the RF frames estimate displacements $\bm{u}$ within the image. The axial strain computed along the direction of US beam propagation is often used to reconstruct a stiffness image, although current model-based methods are far more sophisticated and can provide more quantitative Young's modulus estimates. Our machine learning method (black dash-dot box) originally relied on a FE mesh conforming to the object geometry and a different NNCM for each region exhibiting unique material properties. The new \canns\ shown learn both material property and geometric information, allowing a single network structure to be used with an arbitrary mesh. AutoP is the method whereby force-displacement measurements are transformed into the stress-strain training data for NNCMs/\canns\ (black dotted box). In this report, we focus only on training \canns\ (red dashed box) and assume the spatially varying stresses \stressvec\ and strains \strainvec\ are known.}
	\label{fig:pathway}
\end{figure*}

Our solution was to implement a \textit{data-driven} approach using artificial neural networks (NNs) in place of a pre-defined constitutive model~\cite{hoerig2017}. These neural network constitutive models (NNCMs) learn stress-strain behavior from force-displacement measurements without any initial assumptions of mechanical behavior. The benefit is that after training, NNCMs can be used to compute all relevant stresses and strains, from which material parameters from any constitutive model can be estimated. A block diagram of our method is shown in Fig.~\ref{fig:pathway}. After acquiring force-displacement measurements, we create a finite element (FE) mesh that conforms to both the internal and external geometries of the object from prior knowledge or manual segmentation of the US images. The mesh and measurement data are used in the Autoprogressive Method (AutoP) to train NNCMs for each region exhibiting unique material properties.  

We demonstrated that our method could train NNCMs to accurately characterize the linear-elastic properties of gelatin phantoms in 2-D and 3-D, and obtain Young's modulus estimates of rabbit kidneys similar to those reported in the literature~\cite{hoerig2017b}. However, prior knowledge of object geometry will not be available in a clinical setting, nor can it be assumed that tissue boundaries observed in a US image correspond to actual material property boundaries, precluding the use of segmentation for defining internal structures. Furthermore, the NNCMs could only capture discrete material property distributions. A limitation of the NNCMs is their inability to account for geometric information; segmentation defined the spatial distribution of material properties at the time of FE mesh creation. We can greatly enhance the utility of NNCMs and AutoP for elastography by altering the network architecture to incorporate learning of spatial information.

Cartesian neural network constitutive models (\canns) address this issue. \canns\ accept Cartesian coordinate information as additional inputs to simultaneously learn material property and geometric information independent of the internal structure represented by the FE mesh. In this paper, we introduce \canns\ and describe how this novel architecture learns spatial distributions of material properties from stresses and strains. We leave the details of implementing \canns\ in AutoP to a separate report. Here we use stress-strain data acquired from simulated phantoms to demonstrate the ability of \canns\ to model material property distributions on an arbitrary FE mesh without affecting the learned mechanical behavior. Because \canns\ capture spatial information, we will show how an elastogram can be \textit{reconstructed} directly from a trained model. Finally, we will also demonstrate that, in the presence of noise or changes to the FE mesh, \canns\ are remain able to accurately model linear-elastic mechanical behavior. 

\section{Methods}

\subsection{Overview of the Autoprogressive Method}\label{sec:autop}

AutoP has been primarily developed and used in civil and geotechnical applications to model the mechanical properties of various materials and structures \cite{sidarta1998,hashash2006,hashash2006b,jung2006,jung2010,kim2012,yun2012,ghaboussi2011}. The AutoP approach to constitutive modeling is to apply force-displacement measurements to two finite element analyses associated with a mesh of the imaged object. Forces and displacements applied in separate finite element analyses (FEAs) are connected through NNCMs. Training is to develop NNCMs that consistently relate measured forces and displacements. Meaning, application of measured forces in a FEA results in the measured displacements and vice-versa.

All examples of AutoP prior to our initial report of its use in elasticity imaging have relied solely on surface measurements. The addition of ultrasonic imaging provides internal displacement estimates which are imposed in AutoP along with surface displacements. We exploited the extra displacement data to develop NNCMs that learned the linear-elastic behavior of gelatin phantoms; however, because the FE mesh matched both the internal and external phantom geometry, only a sparse sampling of the internal data was necessary.

Internal displacements under a quasi-static load provide an enormous amount of information regarding internal structure. Because the force stimulus has time to propagate throughout the entire object before force-displacement measurements are acquired, displacements at one location are affected by deformation at all other points in the object. Previously, the NNCMs ignored the spatial aspects of measurements when relating stress to strain. We are now changing the NNCM architecture to incorporate spatial information to make it possible to relax the geometric constraints on the FE mesh. We introduce Cartesian NNCMs that are capable of learning both material and internal geometric properties for the task of forming elasticity images.

\begin{figure}[!h]
	\captionsetup[subfigure]{justification=centering}
	\centering
	\begin{subfigure}[b]{0.45\textwidth}
		\includegraphics[width=\textwidth]{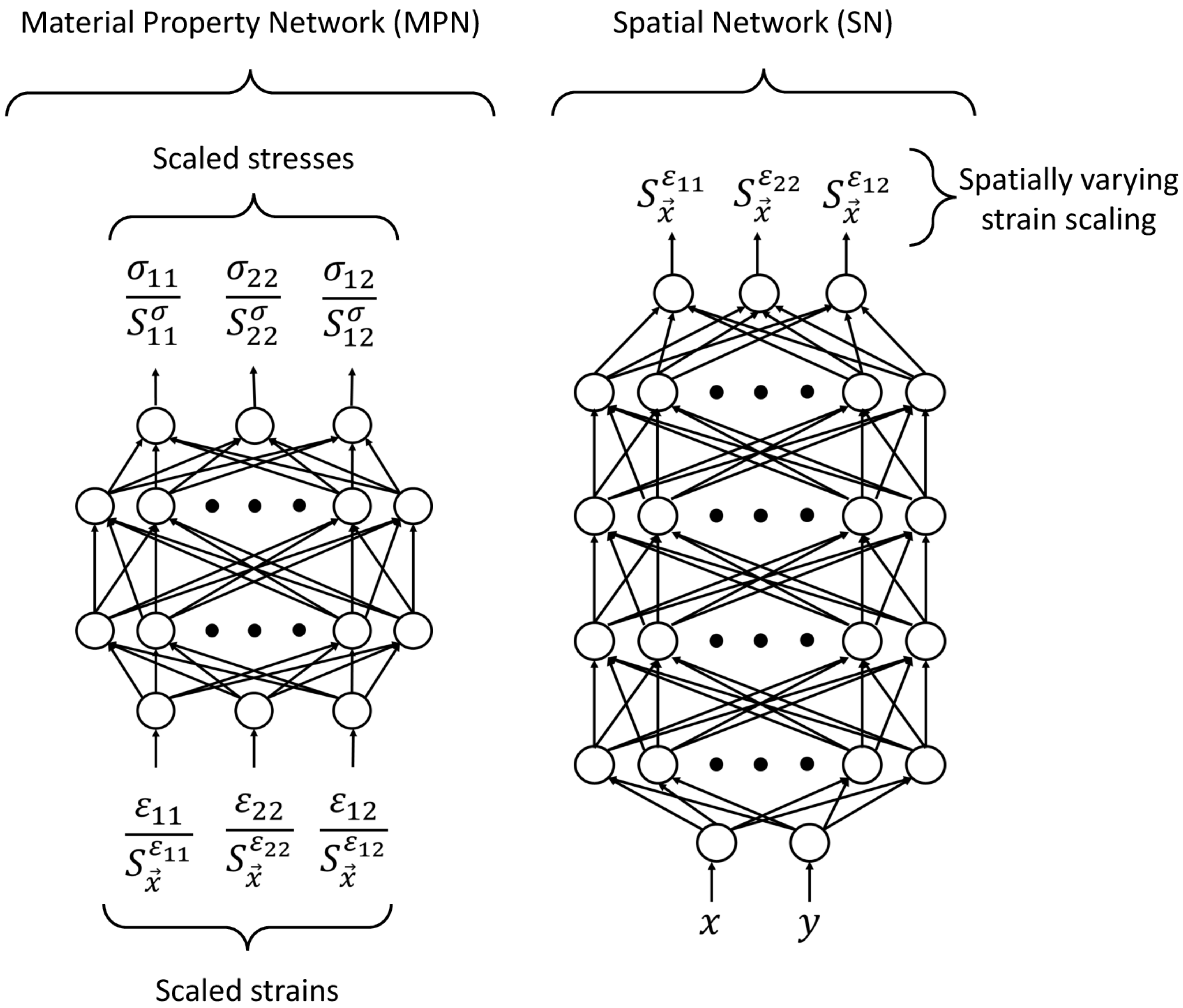}
		\caption{}
		\label{fig:cann_figure}
	\end{subfigure}
	\quad
	\begin{subfigure}[b]{.45\textwidth}
		\includegraphics[width=\textwidth]{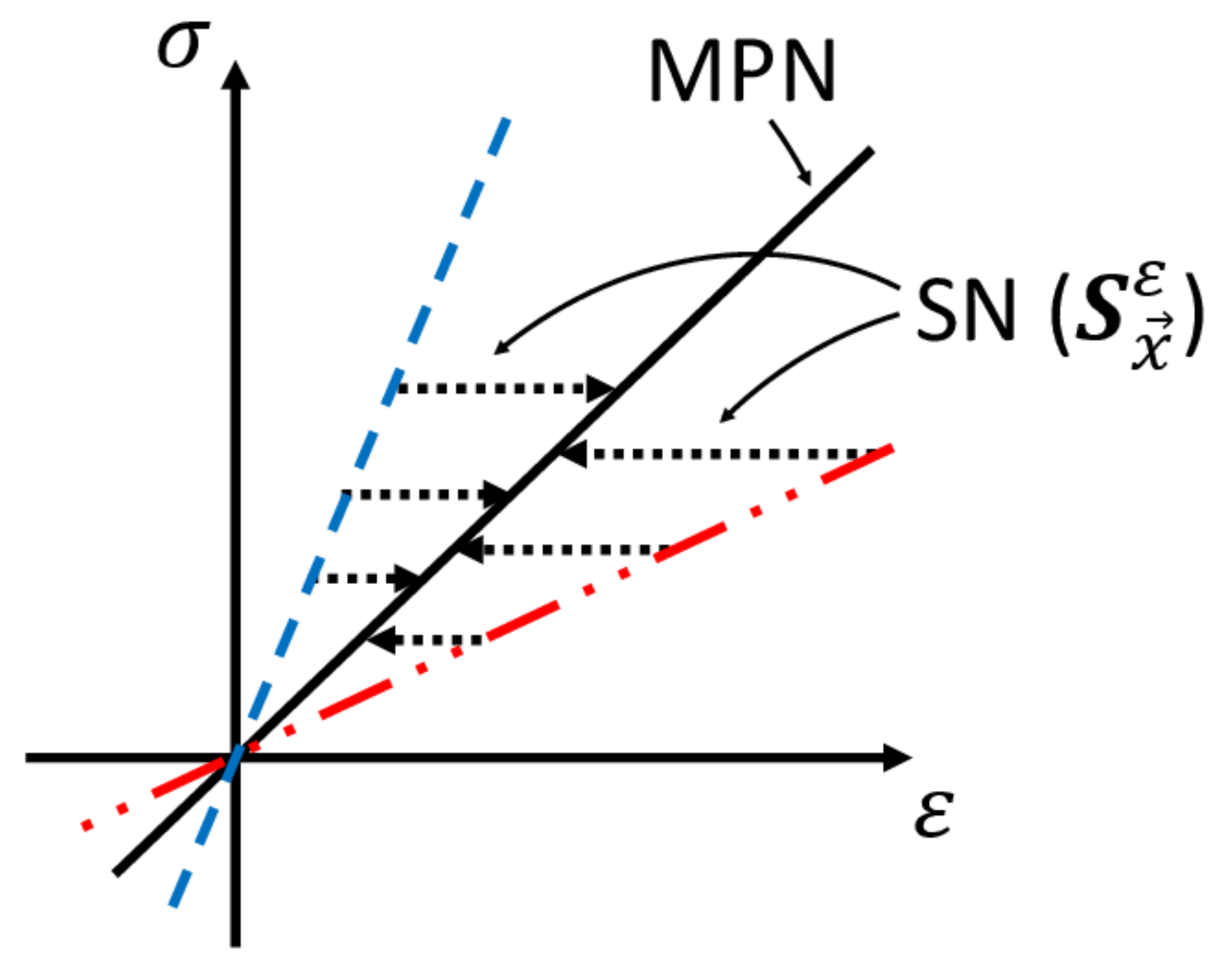}
		\caption{}
		\label{fig:cann_mapping}
	\end{subfigure}
	\caption{(a) Architecture of \canns. The material property network (left) accepts a scaled strain input vector and computes a scaled output stress vector. Previously, the strain scaling vectors were defined for the NNCM and were spatially invariant. Adding a spatial network (right) lets the strain scaling vary with position and allows the pair of networks to learn geometric information. For \canns, $S_{11}^{\sigma} = S_{22}^{\sigma} = S_{12}^{\sigma} = \mstressscale$ and does not change with position.
	(b) Visualization of interaction between MPN and SN. The MPN learns a reference material property (solid black line). Different spatial locations in the object exhibit different material properties (blue-dashed and red-dashed-dot-dot lines). At each location $\bm{x}$, the scaling value \spatscale\ output by the SN transforms the spatially varying material properties to the reference material.}
\end{figure}

\subsection{Cartesian Neural Network Constitutive Models}\label{sec:cann}
The NN structure should be designed to represent the input/output relationship and the intermediate computations that must be performed. For example, many previous implementations of AutoP constitutive modeling relied on fully-connected networks to represent a linear stress-strain relationship. Other NNCMs characterizing more complex behaviors, including nonlinearity and path-dependence, utilized a nested NN structure\cite{ghaboussi1998b} to depict the influence of previous stress and strain states on the current mechanical behavior. In our prior work using AutoP for elastography, a simple feed-forward, fully-connected architecture was sufficient (left side of Fig.~\ref{fig:cann_figure}). Here, NNCMs computed a stress vector output from a strain vector input. We accounted for the possible positive and negative input/output values --- and the sign symmetry between stresses and strains --- by using hyperbolic tangent activation functions. The choice of hyperbolic tangent also ensures that under zero strain, there is zero stress. To avoid saturating the input nodes and keep the NNCMs sensitive to changes in the input strains, a scaling vector \strainscale\ was selected to scale each component of the input strains within the $\pm 0.8$ range. Similarly, the output is bounded to $\pm 1.0$ whereas stresses can extend well beyond that range. As such, a different scaling vector \stressscale\ was selected to keep the output within $\pm 0.8$. The flow of data through the NNCM was therefore $\bm{\varepsilon}~\rightarrow~\bm{\varepsilon}/\mstrainscale~\rightarrow~ {\rm NNCM}~\rightarrow \bm{\sigma}/\mstressscale \rightarrow \bm{\sigma}$, where the vector divisions are element-by-element operations. We refer to this type of NNCM as a material property network (MPN).

Creating a NNCM to learn internal structure required spatial information be present somewhere in the flow of data through the network. Instead of increasing the number of inputs to include spatial coordinates, scaling factors are learned to alter the material properties characterized by the MPN. We consider the stress-strain behavior represented by the MPN with $\mspatscale = \bm{1}$ to be the ``reference'' material property (i.e., $R_m:\bm{\varepsilon}\rightarrow\bm{\sigma}$). For the case of linear-elastic materials, the stress-strain diagram is a straight line with a slope that is the Young's modulus as determined by MPN. Stresses are computed by simpling multiplying a strain vector by the stiffness matrix ${\bm C}$: $\bm{\sigma} = {\bm C \bm{\varepsilon}}$. Changing the strain scaling at a point $(x,y)$ alters the slope of the line, effectively changing the Young's modulus at that location. \canns\ accomplish this change by introducing a spatial network (SN) that computes \spatscale\ based on a coordinate input (right side of Fig.~\ref{fig:cann_figure}). A SN is the function $R_s:\bm{x}\rightarrow\mspatscale$. We also reduce the stress scaling vector to a single value: $S_{11}^{\sigma} = S_{22}^{\sigma} = S_{12}^{\sigma} = \mstressscale$. 

Details of how \spatscale\ and \stressscale\ are computed is covered in Sec.~\ref{sec:calc_spat_scale} where we show that the spatial scaling values produce a map of relative stiffnesses where larger \spatscale\ tend to correspond to softer regions. To understand why, consider the reference linear-elastic relationship learned by the MPN. Under uniaxial loading, a heterogeneous material exhibits large strains and small stresses in soft regions. In contrast, stiff regions produce large stresses for small strains. Therefore, larger \spatscale\ values in soft regions decrease the magnitude of the strain vector input to the MPN. Similarly, smaller scaling values in stiff regions result in larger strains after scaling. In broad terms, \spatscale\ acts as a function transformation for the relationship learned by the MPN: $\bm{\sigma} = f(\bm{\varepsilon}/\mspatscale)$. In the more specific case of linear-elastic materials, the SN acts as a spatially-varying matrix that operates on the strains before multiplying by the stiffness matrix: $\bm{\sigma} = \bm{CE}(x,y)\bm{\varepsilon}$. The interaction between the MPN and SN is illustrated in Fig.~\ref{fig:cann_mapping}. Together, the networks learn the mapping $R_m, R_s:\mstrainvec\rightarrow\mstressvec$

The SN also has a fully-connected, feed-forward architecture. Unlike the MPN counterpart, spatial networks use a mix of logistic and hyperbolic tangent activation functions. Considering that the output \spatscale\ is always positive, the logistic function is a natural choice for the output nodes. Conversely, the input $(x,y)$ can span the positive and negative range, but a vector of zeros at the input does not imply the output should also be zero. We thus use a logistic activation function for the first layer as well. All intermediate layers use a hyperbolic tangent. As with the MPN, care must be taken to bound the spatial network inputs and outputs: input values outside the $\pm 1.0$ can saturate the input nodes (and reduce sensitivity) while outputs not contained within $(0,1)$ cannot be achieved by a logistic function. We therefore scale the input $(x,y)$ values to within $\pm 1.0$. Preliminary tests showed that setting the coordinate origin to the center of the FE mesh produced the best results. A similar shifting and scaling to the $0.1-0.8$ range is performed for the output \spatscale\ values before training the spatial network.

Both the material property and scaling networks learn from the same set of stress-strain data. These data would be estimated in AutoP after applying force and displacement measurements in FEAs. Each network extracts different information from the same set of data. Splitting the material property and geometry problems allows each network to learn a simpler input-output relationship. Combining the two networks results in a cooperative \cann\ structure that captures both mechanical behavior and its geometric variation. 

\subsection{Calculating Spatial Values}\label{sec:calc_spat_scale}
Inputs to the SN --- Cartesian coordinates $\bm{x}$ --- are defined by the FE mesh and thus known \textit{a priori}. Given a trained MPN, spatially varying stresses \stressvec\, strains \strainvec\, and coordinates, the task of determining the target output of the SN remains. 

In preliminary studies, many methods of computing the spatial scaling values were evaluated. These methods all relied on calculating \spatscale\ using a pre-defined function of $\bm{x}$, \strainvec, and/or the stiffness matrix. Each method was successful to some degree, with the caveat that chosen functions influenced the material properties or geometry learned by the \cann. For example, a simple function we tested was computing the spatial scaling as the ratio of stress and strain magnitudes: $\mspatscale = ||\bm{\sigma}_{\bm{x}}||/||\bm{\varepsilon}_{\bm{x}}||$. This method would likely work for 1-D linear-elastic materials, but 2-D stress-strain distributions are more complex and the magnitude of stress to strain changes, even for the same material, based on applied load and location in the material. 

More importantly, a pre-defined function does not directly account for errors between the stress vector output from the MPN and the ``target'' stress. The MPN and SN are cooperative and therefore work together to minimize this error. Gradient-descent methods are utilized for computing \spatscale\ based on the difference between stress estimated by the MPN in response to a strain input and the target stress $\bm{\sigma}^t$ computed via FEA. Similar to the backpropagation algorithm for updating ANN connection weights, the error at the output of the material property network can be propagated back to the spatial scaling values. For simplicity, consider the stress $\bm{\sigma}^t(\bm{x})$ and strain $\bm{\varepsilon}(\bm{x})$ computed by FEA at a single location $\bm{x}$. The current value of $\mspatscale$ and $\bm{\varepsilon}(\bm{x})$ can be used to compute $\bm{\sigma}^{NN}(\bm{x})$, the value of stress predicted by the MPN: $R_m:\bm{\varepsilon}(\bm{x})\rightarrow\bm{\sigma}^{NN}(\bm{x})$. The goal is to minimize the objective function
\begin{flalign}
\bm{S}_{\bm{x}}^{\varepsilon} = \argmin_{\hat{\bm{S}}_{\bm{x}}^{\varepsilon} \in \mathbb{R}}\  \operatorname{f_m} (\bm{\sigma}^t(\bm{x}), \bm{\sigma}^{NN}(\bm{x})).\label{eq:spatial_obj}
\end{flalign}
The function $\operatorname{f_m}(\cdot)$ is the $L_2$ norm:
\begin{flalign}
\operatorname{f_m} = E &= \frac{1}{2}\sum_{i=1}^{3}(\sigma_i^t(\bm{x}) - \sigma_i^{NN}(\bm{x}))^2\label{eq:loss_func} \\
&= \frac{1}{2}\sum_{i=1}^{3}e_i^2,
\end{flalign}
where (dropping $(\bm{x})$ for brevity)
\begin{flalign}
\sigma_i^{NN} &= S_{\sigma}\sigma_{i}'^{,NN} \\
\varepsilon_j &= \SJi\varepsilon_j' \\
e_i &= \sigma_i^t - \sigma_i^{NN}.\label{eq:stress_error}
\end{flalign}
We define $\sigma_k^t$ as the $k^{th}$ component\footnote{There are three components in the stress vector for 2-D models, ordered as $[\sigma_{11},\ \sigma_{22},\ \sigma_{12}]$. The same ordering is used for the strain vector.} of ``target'' stress at $\bm{x}_i$, $\sigma_k^{NN}$ is the corresponding stress component predicted by the MPN in response to $\bm{\varepsilon}$, and $(i,j,k)$ have the range $(1,2,3)$. Scaled input and outputs of the MPN are denoted as $\varepsilon_j'$ and $\sigma_{i}'^{,NN}$, respectively. Calculating the partial derivate of the error with respect to $\SKi$\ is straightforward:

\begin{flalign}
\frac{\partial E}{\partial\SK} &= \frac{\partial}{\partial\SK}\frac{1}{2}\sum_{i=1}^3e_i^2 \nonumber\\
&= \sum_{i=1}^3 e_i \bigg[ \partialSK (\sigma_i^t - \sigma_i^{NN}) \bigg].
\end{flalign}
The partial derivative is distributed to the stress terms, noting that $\sigma_i^t$ was computed in a FEA and the partial derivative with respect to $\SKi$ is zero. For $\sigma_i^{NN}$ we invoke the chain rule:

\begin{flalign}
&= \sum_{i=1}^3 e_i \bigg[ -\sum_{j=1}^3 \frac{\partial\sigma_i^{NN}}{\partial\varepsilon_j} \frac{\partial\varepsilon_j}{\partial\SK} \bigg].
\end{flalign}
The term $\partial\sigma_i^{NN}/\partial\varepsilon_j = D_{ij}$ is the stiffness matrix relating stress to strain and can be calculated via the weights of the material property network\cite{hashash2004}. Computing the last factor in the braces:

\begin{flalign}
\frac{\partial\varepsilon_j}{\partial\SK} &= \partialSK (\SJ\varepsilon_j') \nonumber\\
&= \varepsilon_j'\delta_{kj},
\end{flalign}
where $\delta_{kj}$ is the Kronecker delta function. Finally, we arrive at the final expression for the error gradient:

\begin{flalign}
\frac{\partial E}{\partial\SK} &= -\sum_{i=1}^3 e_i \sum_{j=1}^3 D_{ij}\varepsilon_j'\delta_{jk}.
\end{flalign}
The update increment for $\mspatscalei$ is the negative of the gradient multiplied by the value $\eta$ to adjust the increment size:

\begin{flalign}
\Delta\SK &= \eta \sum_{i=1}^3 e_i \sum_{j=k} D_{ij}\varepsilon_j'.\label{eq:full_update}
\end{flalign}
In Eq.~\ref{eq:full_update}, the inner sum is approximately equal to the stress computed by the MPN  with all $j\neq k$ components of the scaled input strain vector set to zero. We call this stress vector $\hat{\bm \sigma}_i'^{,NN}$. While it is possible to calculate $D_{ij}$ directly, we can greatly reduce the computational load with this approximation, leading to a final equation for $\Delta\SK$:
\begin{flalign}
\Delta\SK \approx \eta\sum_{i=1}^3 e_i \hat{\sigma}_i'^{,NN}\label{eq:delta_approx}
\end{flalign}
Computing $\Delta\SKi$ using Eq.~\ref{eq:delta_approx} is not significantly different from Eq.~\ref{eq:full_update} and is nearly two orders of magnitude faster. Controlling the increment size with $\eta$ is equivalent to applying a learning rate in backpropagation. 

Eq.~\ref{eq:spatial_obj} attempts to minimize the stress error for a single stress-strain pair. However, many stress-strain pairs may exist at $\bm{x}$, meaning the stress error should be minimized in an average sense for all stress-strain pairs at $\bm{x}$. We do this by invoking Eq.~\ref{eq:delta_approx} for each data pair at $\bm{x}$ and computing the mean of $\Delta\mspatscalei$ before adding to $\mspatscale$.

A single application of Eq.~\ref{eq:delta_approx} is insufficient for updating $\mspatscale$. Alg.~\ref{alg:grad_iter} details the iterative process for computing a new \spatscale\ at location $\bm{x}$. $N$ corresponds to the number of gradient-descent iterations and $N^{\sigma}$ is the number of stress-strain pairs at $\bm{x}$.

\begin{algorithm}[!h]
	\caption{Iterations for computing $\mspatscale$}\label{alg:grad_iter}
	\begin{algorithmic}[1]
		\State{Given: current $\mspatscale$, $\bm{\sigma}_i,\ \bm{\varepsilon}_i$ at $\bm{x}$}
		\For{$n = 1, 2, ..., N$}
		\For{$k = 1, 2, 3$}
		\State $\Delta S = 0$
		\For{$i = 1, 2, ..., N^{\sigma}$}
		\State Compute $\bm{\sigma}_i^{NN}$ and (vector) $\hat{\bm{\sigma_i}}'^{,NN}$ using $\bm{\varepsilon_i}$
		\State Compute $\bm{e}_i = \bm{\sigma}_i^t - \bm{\sigma}_i^{NN}$
		\State Compute $\Delta S = \Delta S + \sum_{p=1}^3 e_p \hat{\sigma}_p'^{,NN}$
		\EndFor
		
		\State $\Delta\SK = \eta\Delta S/N^{\sigma}$
		\State $\SK = \SK + \Delta\SK$
		\EndFor
		\EndFor
	\end{algorithmic}
\end{algorithm}


Computing the stress scaling value \stressscale\ is far simpler. It is chosen to ensure all components of every stress vector falls within the $\pm 0.8$ range. Again, this follows from upper and lower bounds of the hyperbolic tangent activation function being $\pm 1.0$. In this study, setting $\mstressscale = 1.0$ is sufficient because the magnitude of every computed stress falls below 0.8.

\begin{figure}[!h]
	\captionsetup[subfigure]{justification=centering}
	\centering
	\begin{subfigure}[b]{0.25\textwidth}
		\centering
		\includegraphics[width=\textwidth]{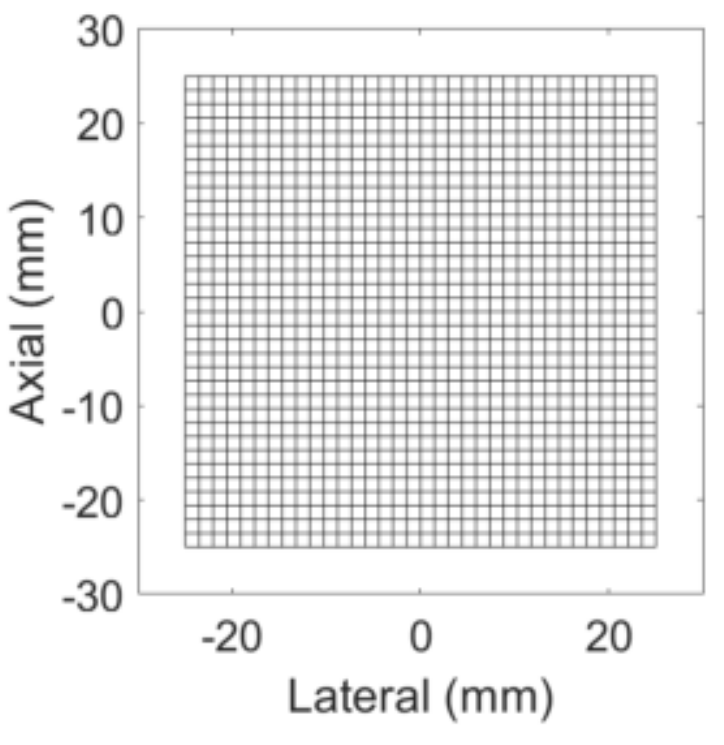}
		\caption{}
		\label{fig:rect_mesh}
	\end{subfigure}
	\quad
	\begin{subfigure}[b]{0.22\textwidth}
		\raisebox{7mm}{
		\includegraphics[width=\textwidth]{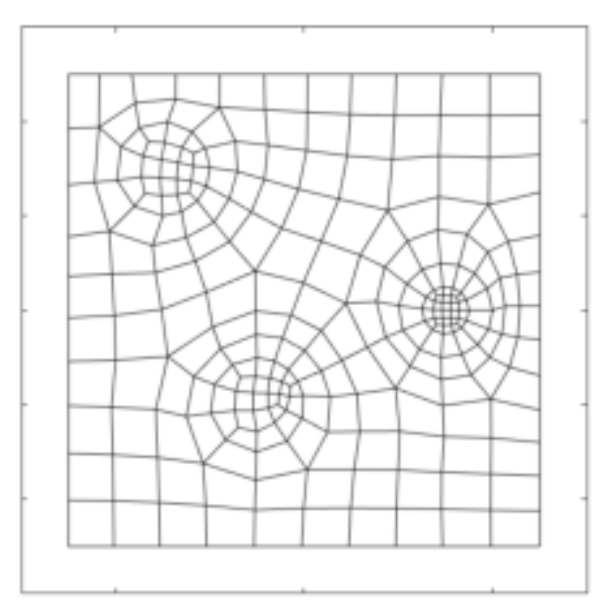}}
		\caption{}
		\label{fig:inc_mesh}
	\end{subfigure}
	\caption{(left) Rectilinear FE mesh (Mesh 1) with 35 nodes per edge. This mesh was used to compute stress and strain fields for all four models. (right) FE mesh that conforms to the geometry of the three inclusion model (Mesh 2). }
	\label{fig:mesh_figs}
\end{figure}

\subsection{Simulated Phantoms}

\canns\ learn material and geometric properties from stress-strain data. Here, we turn to simulated data to evaluate \canns\ under the ideal circumstance of stresses and strains being known exactly, and hence this study is confined to the dashed-line box in Fig.~\ref{fig:pathway}. Stress-strain data were generated by FEA (ABAQUS 6.13) with known Young's modulus distributions. A simple four-node, quadrilateral element rectilinear mesh with 35 nodes per 50 mm edge (Fig.~\ref{fig:rect_mesh}) was selected to demonstrate the independence of \canns\ to the FE mesh. In the FEA, we used a plane-stress, incompressible (Poisson's ratio $\mu = 0.5$) material model, the bottom surface of mesh was pinned to create a fixed boundary condition (BC), and a US probe was pressed into the top surface. The probe-phantom interface was modeled as frictionless to allow lateral motion of the top phantom surface during compression. Four equal compressive loads were applied by the probe up to 13.57 mN, leading to a minimum probe displacement of 0.98 mm and a maximum of 2.23 mm depending on the material property distribution. 

\begin{figure*}[!h]
	\captionsetup[subfigure]{justification=centering}
	\centering
	\begin{subfigure}[b]{0.24\textwidth}
		\includegraphics[width=\textwidth]{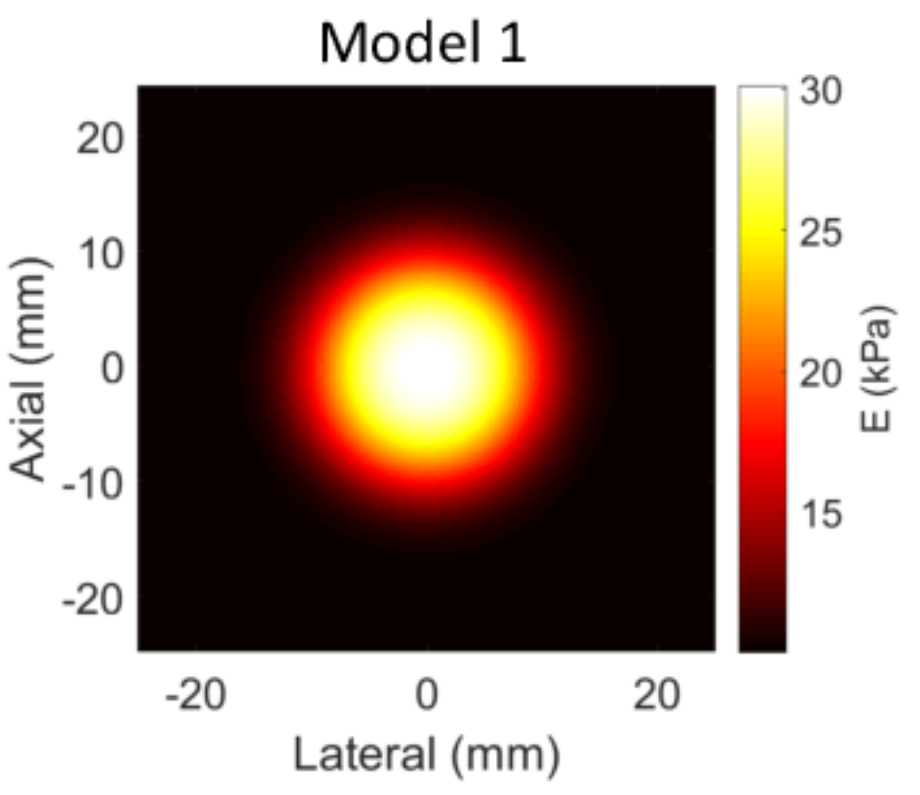}
		\caption{}
		\label{fig:model_1}
	\end{subfigure}
	\hfill
	\begin{subfigure}[b]{0.24\textwidth}
		\includegraphics[width=\textwidth]{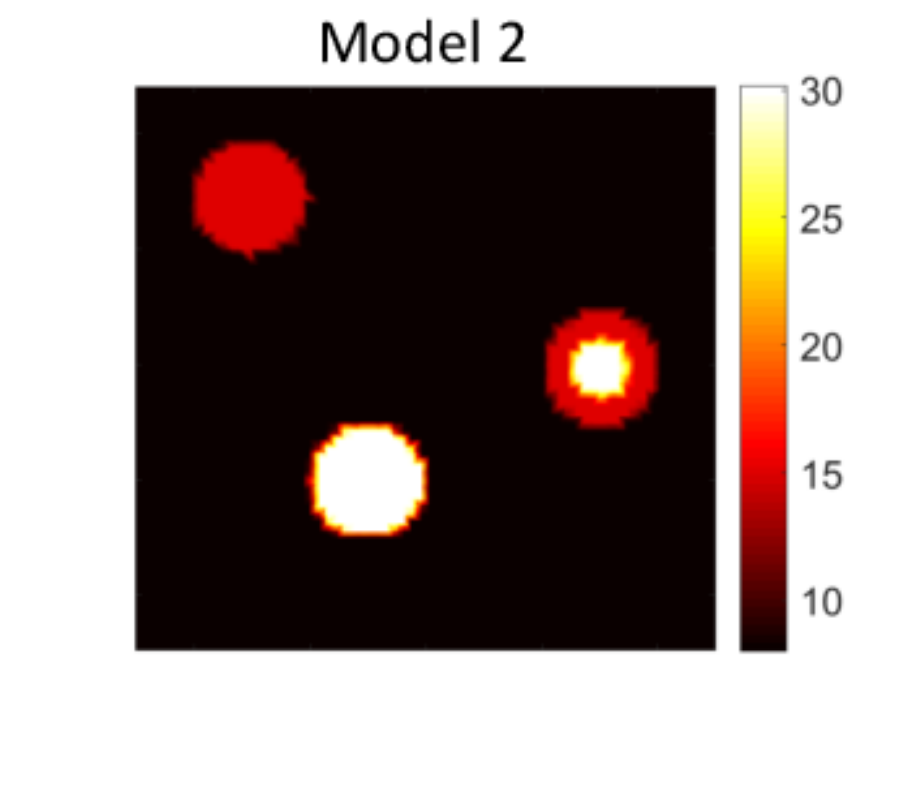}
		\caption{}
		\label{fig:model_2}
	\end{subfigure}
	\hfill
	\begin{subfigure}[b]{0.24\textwidth}
		\includegraphics[width=\textwidth]{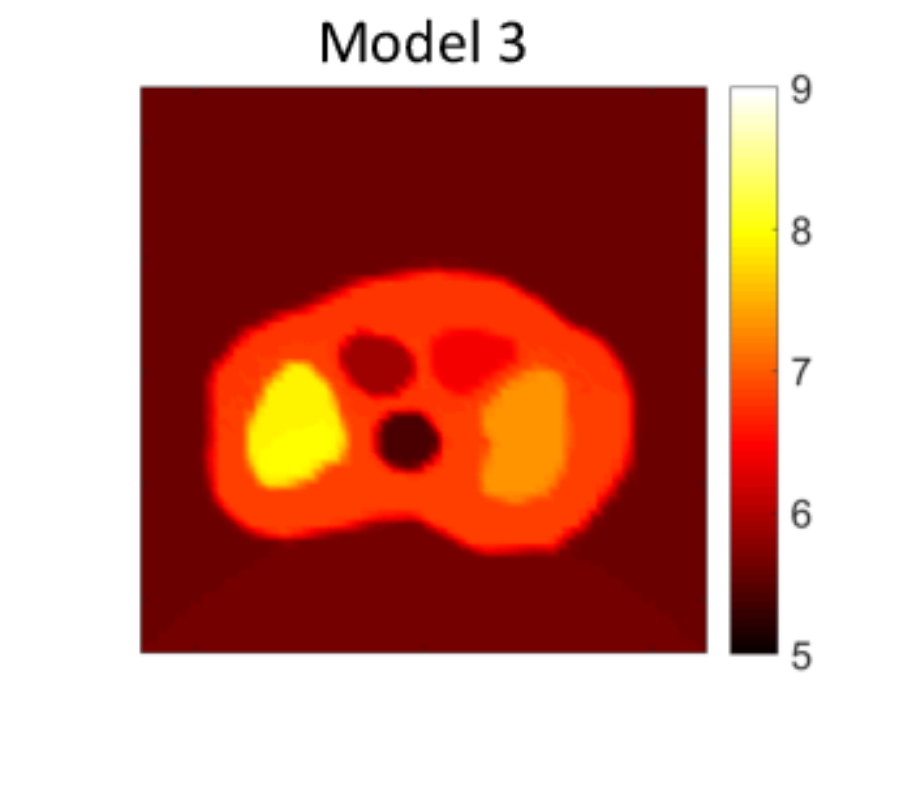}
		\caption{}
		\label{fig:model_3}
	\end{subfigure}
	\hfill
	\begin{subfigure}[b]{0.24\textwidth}
		\includegraphics[width=\textwidth]{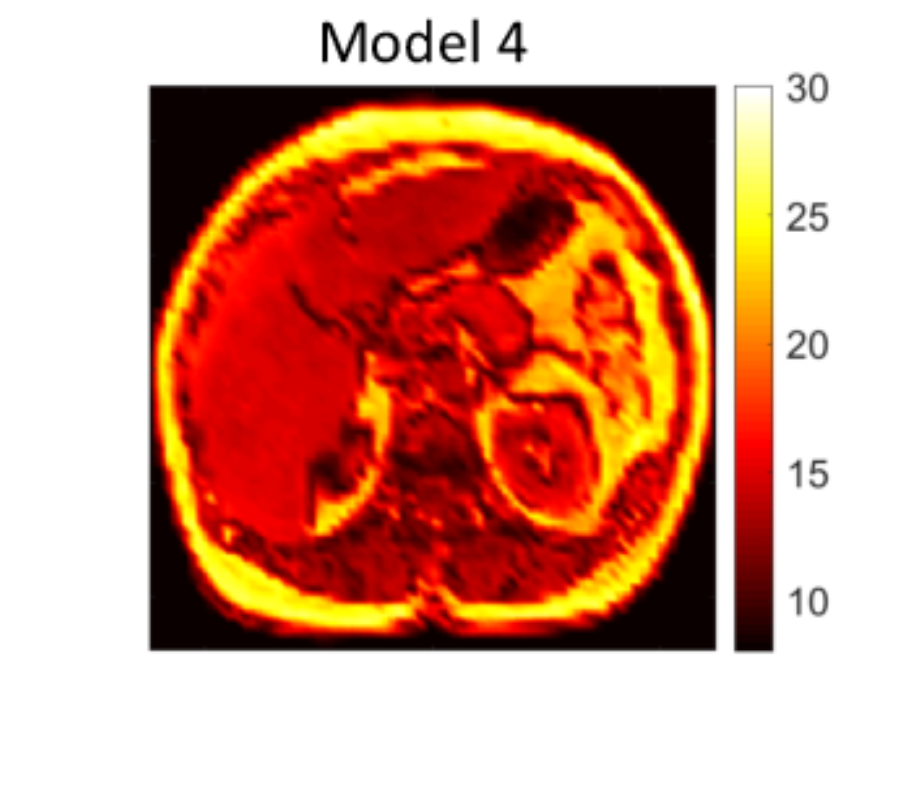}
		\caption{}
		\label{fig:model_4}
	\end{subfigure}
	\vfill
	
	\begin{subfigure}[b]{0.24\textwidth}
		\includegraphics[width=\textwidth]{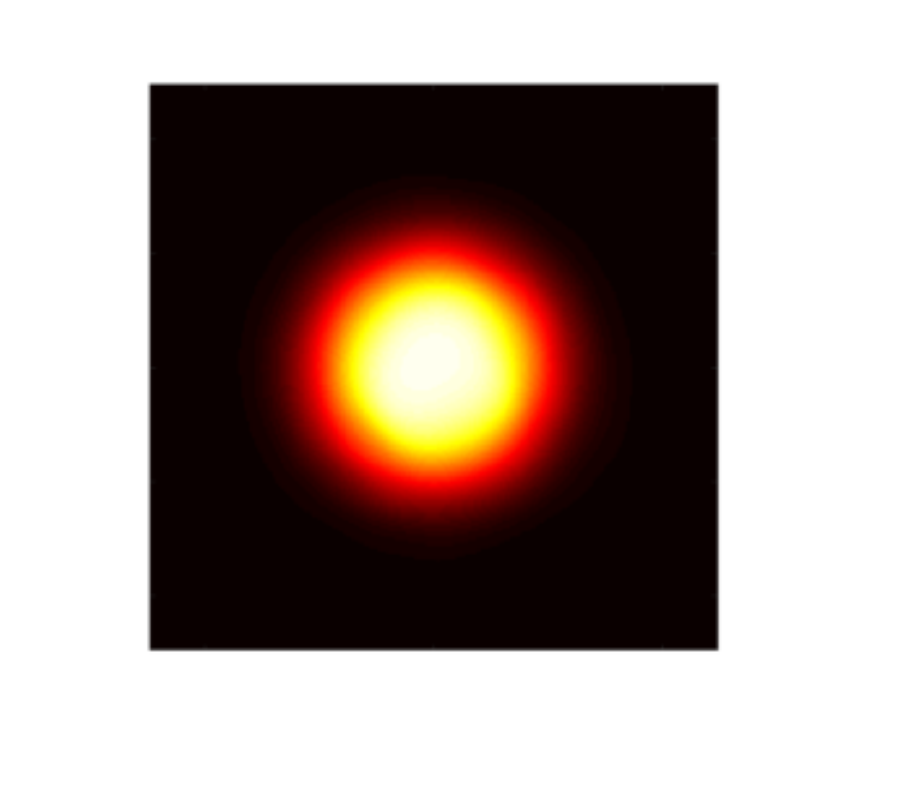}
		\caption{}
	\end{subfigure}
	\hfill
	\begin{subfigure}[b]{0.24\textwidth}
		\includegraphics[width=\textwidth]{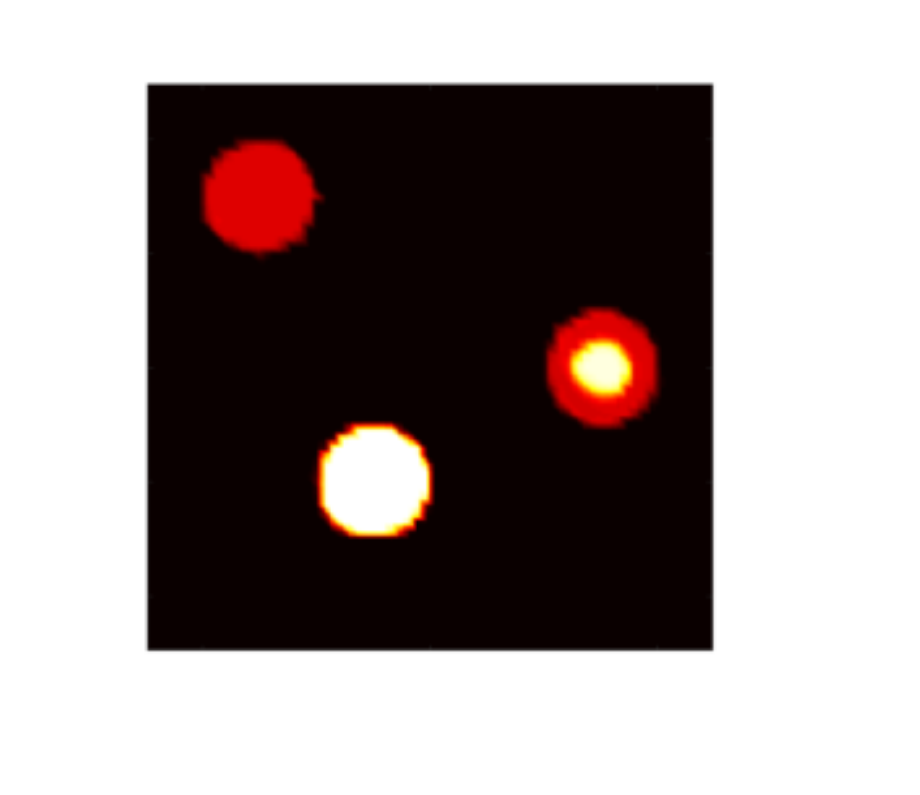}
		\caption{}
	\end{subfigure}
	\hfill
	\begin{subfigure}[b]{0.24\textwidth}
		\includegraphics[width=\textwidth]{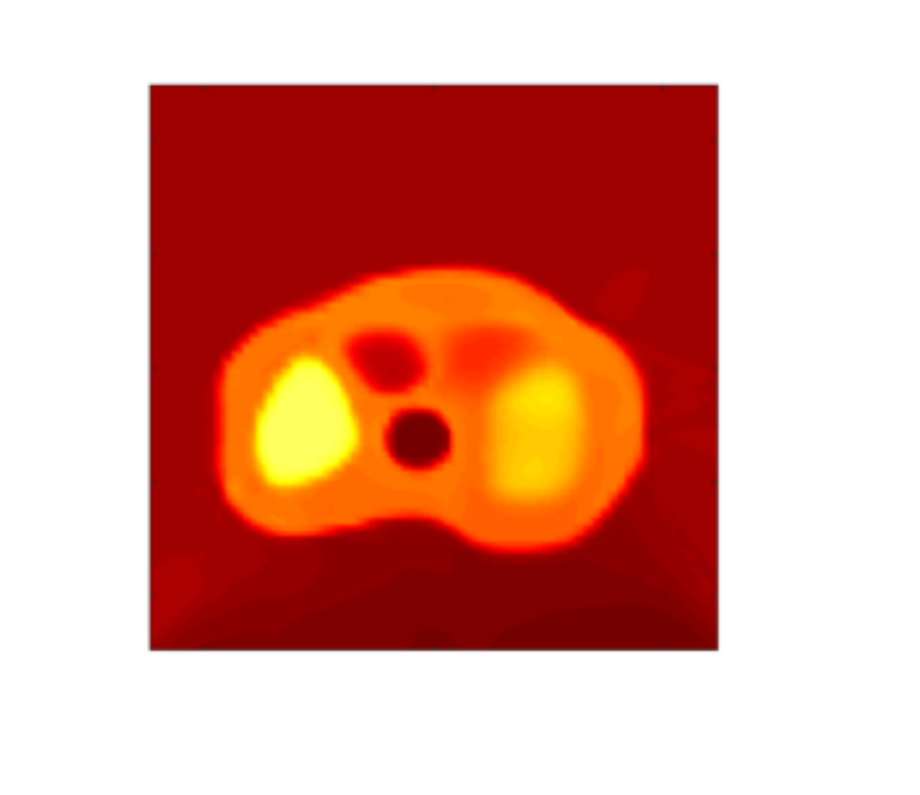}
		\caption{}
	\end{subfigure}
	\hfill
	\begin{subfigure}[b]{0.24\textwidth}
		\includegraphics[width=\textwidth]{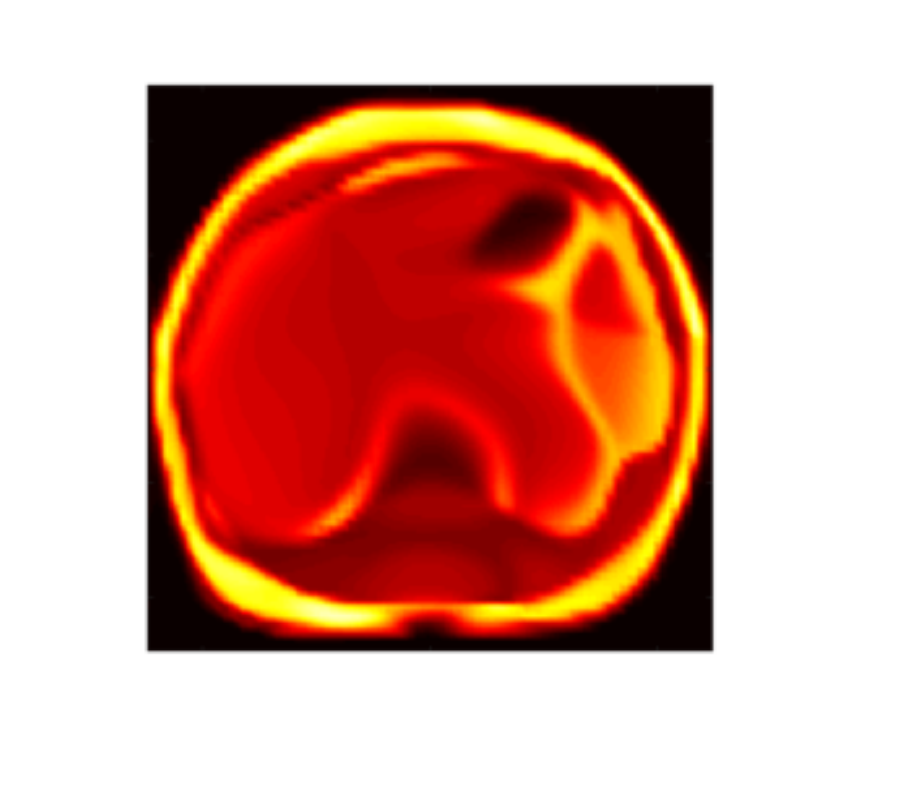}
		\caption{}
		\label{fig:cann_1_4}
	\end{subfigure}
	\vfill
	
	\begin{subfigure}[b]{0.24\textwidth}
		\includegraphics[width=\textwidth]{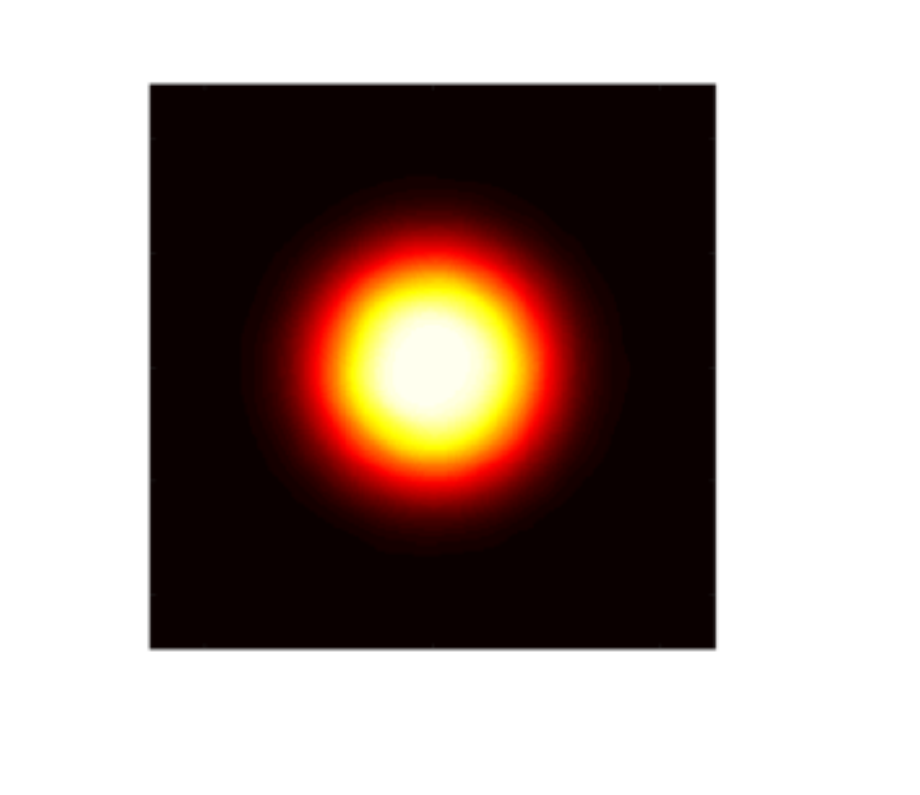}
		\caption{}
	\end{subfigure}
	\hfill
	\begin{subfigure}[b]{0.24\textwidth}
		\includegraphics[width=\textwidth]{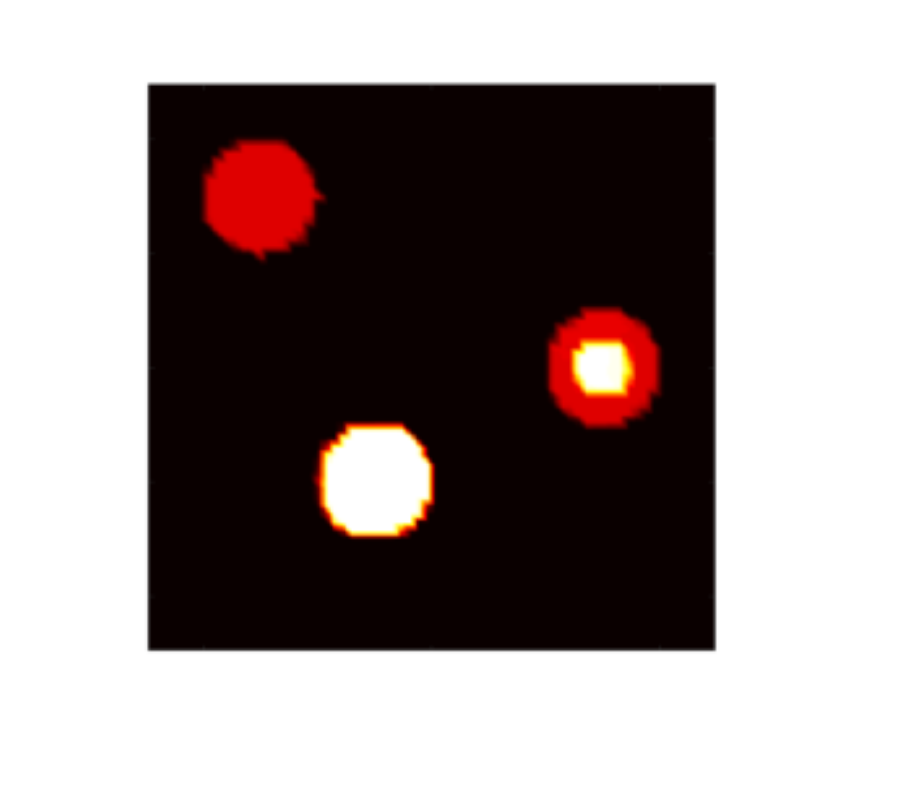}
		\caption{}
	\end{subfigure}
	\hfill
	\begin{subfigure}[b]{0.24\textwidth}
		\includegraphics[width=\textwidth]{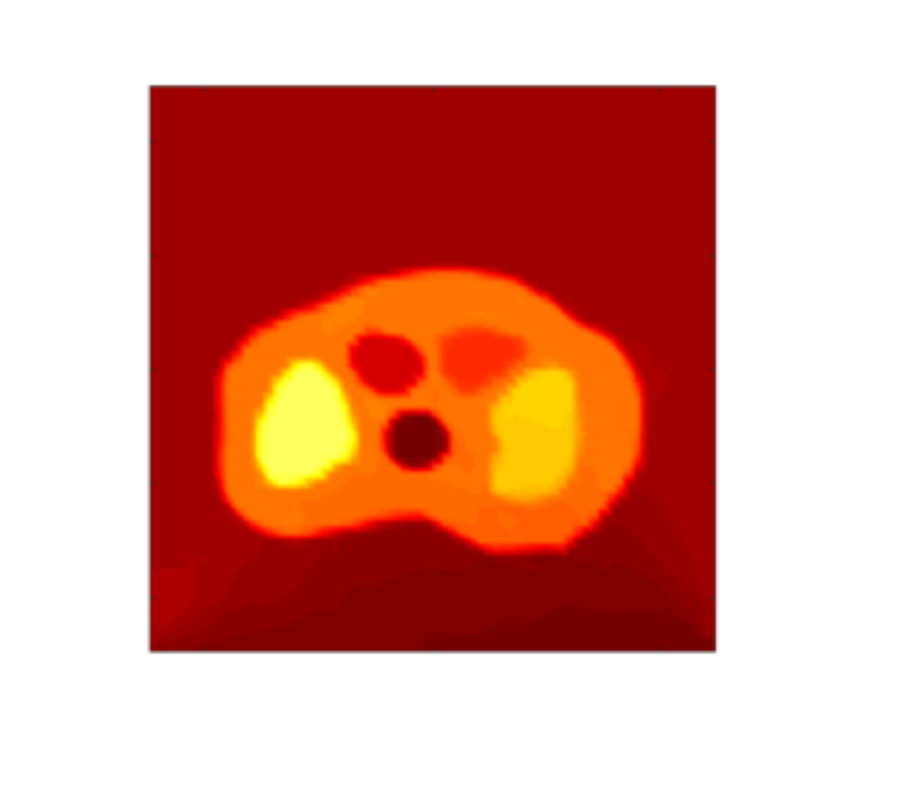}
		\caption{}
	\end{subfigure}
	\hfill
	\begin{subfigure}[b]{0.24\textwidth}
		\includegraphics[width=\textwidth]{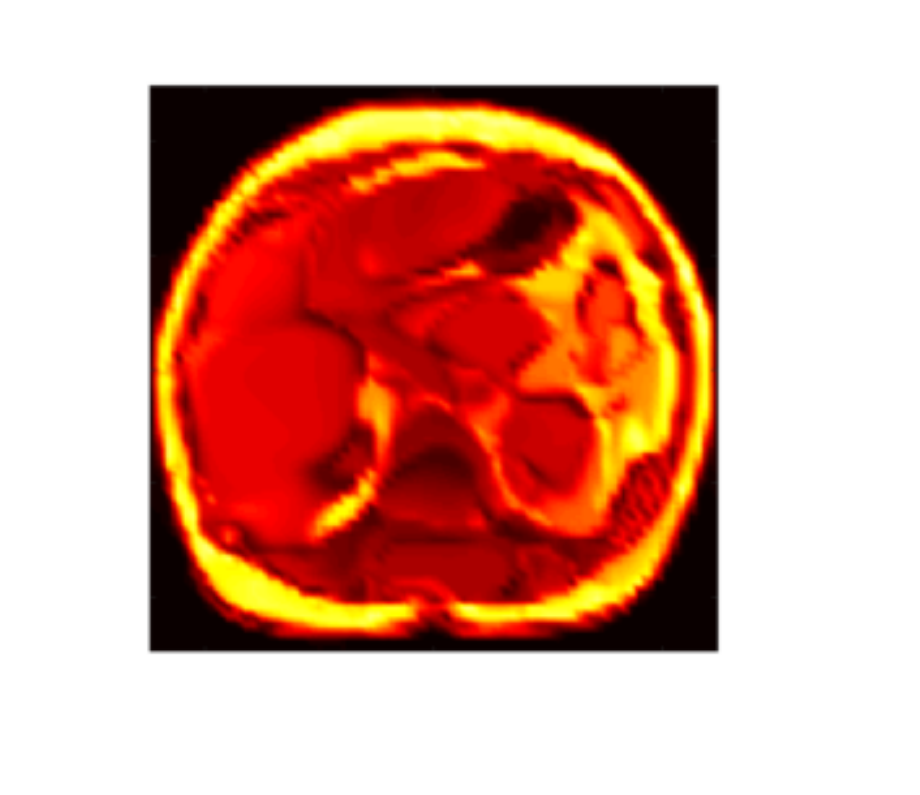}
		\caption{}
		\label{fig:cann_2_4}
	\end{subfigure}
	
	\caption{(Top row) Target Young's modulus distributions for the four 50x50 mm models. Model~4 was created by converting the gray scale values from an abdominal MR image to Young's modulus values. It was included only as an extreme case of spatial complexity that represented actual human physiology and is not a real case of elasticity imaging. (Middle row) Young's modulus images reconstructed by \canns\ after spatial scaling update and training using the stress-strain data generated by a forward FEA. Spatial networks were trained using Test 1 parameters. (Bottomw row) Young's modulus images from after SNs trained with Test~2 parameters.}
	\label{fig:youngs_figs}
\end{figure*}

Four different simulated phantom models, displayed in the top row of Fig.~\ref{fig:youngs_figs}, were selected to test different aspects of \canns. Model 1 (Fig.~\ref{fig:model_1}) is a stiff, Gaussian-shaped inclusion embedded in a soft background. The peak Young's modulus of the inclusion was 30 kPa and smoothly transitioned into the 10 kPa background. We chose this model to demonstrate the ability of \canns\ to capture smooth, continuous material property distributions, a feat not achievable in our prior work with NNCMs~\cite{hoerig2017}. Models 2 and 3 have abrupt transitions in the material property distributions. Model 2 contains three stiff inclusions (15 kPa and 30 kPa) in a 8 kPa background. Model 3 represents a rabbit kidney embedded in a block of gelatin. We previously performed this experiment and trained seven NNCMs in AutoP with the force-displacement measurements~\cite{hoerig2017b}. The Young's modulus values chosen for this simulated phantom correspond to the moduli estimated from those seven NNCMs. Model 4 was selected as an extreme case of complex spatial geometry. To generate this model, the gray-scale values of an abdominal MRI scan were scaled to the 8-30 kPa range of Young's modulus values. Model 4 does not represent a real case of elasticity imaging nor do we claim any translational use of \canns\ to MRI. The image was only chosen for its geometric complexity while also representing actual human physiological structure.

Two additional data sets were generated with Model 3 after adding noise to the target Young's modulus distribution. Uniformly distributed random values up to $\pm10\%$ and $\pm30\%$ the local Young's modulus value ($\approx 27.2$~dB and $\approx 17.6$~dB peak signal-to-noise ratio, respectively) were added to the target distribution of Model 3. We generated two different corrupted target distributions for each test, performing the FEA with the loads and BCs previously specified, and compiled stresses from one analysis and strains from the other. Diagrams describing the data generation techniques are provided in Fig.~\ref{fig:app_1} of the Appendix~A.

We created Mesh 2 (Fig.~\ref{fig:inc_mesh}) that conforms to the geometry of Model 2. The same force loads and BCs described above were applied in a FEA of this model to generate the stress-strain data. While \canns\ are independent of the internal FE mesh structure, the learned distribution of stresses and strains will be affected. For example, the inner inclusion of Model 2 fills four whole elements in Mesh 1 and partially fills twelve adjacent elements. Conversely, 16 whole elements in Mesh 2 comprise the same nested inclusion. Furthermore, there are 1156 total elements in Mesh 1 and 235 is Mesh 2. The ratio of generated data from that one inclusion to all other points greatly increases and more accurately captures the geometry. 

Before the spatial scaling values could be computed for each model, it was necessary to pretrain a MPN. Without a trained MPN there is no reference for updating \spatscale. The material property network consisted of two hidden layers with six nodes per layer. Weights were initialized by drawing from a uniform distribution in the range [-0.2, 0.2]. We generated 5000 strain vectors, whose components were also randomly generated uniformly in the range [-0.2, 0.2], and computed the corresponding stress vectors using a plane-stress model with a Young's modulus value of 10 kPa and Poisson's ratio $\nu = 0.5$. Note that the initialization range for the weights and the strain vectors do not have to match. Previous results suggested this range performed well for the MPN. As for the strains, we chose a range that extends beyond the magnitude of the strain vectors generated for the aforementioned models.  Before weight update via the resilient backpropagation (RPROP) algorithm~\cite{riedmiller1993}, frame invariance of the stresses and strains was enforced by rotating the data $90^{\circ}$ and appending the new rotated data to the original set, doubling the total number of training pairs. This rotation was done by simply swapping the axial and lateral components of the data and was implemented in our previous study with AutoP~\cite{hoerig2017}. Finally, we trained the MPN over 50 epochs with $\mspatscale = \bm{1}$ and $\mstressscale = 1$.

After generating all data sets and pretraining the MPN, we used the stress-strain data from all four load increments in each set to compute new spatial scaling values using Alg.~1 ($N = 150$, $N^{\sigma} = 8$ due to frame invariance, $\eta = 2.5$). Spatial networks were trained for each model using the newly computed \spatscale. Each network was comprised of five hidden layers with 25 nodes per hidden layer. Training was split into iterations to mimic what occurs in AutoP. For example, instead of simply training the SN over 12000 epochs, we split training into iterations where fewer epochs were used. 10 iterations of 300 epochs were used in Test~1 whereas 30 iterations of 600 epochs were used in Test~2. These were equivalent to training for 3000 and 12000 epochs in a single iteration, respectively. Training for the SN was implemented in TensorFlow using He initialization of the weights~\cite{he2015delving}, the Adam optimizer~\cite{kingma2014} with default settings, and a learning rate of 0.03. 

The trained spatial network for each model was then paired with the pretrained material property network to form a \cann\ and used to reconstruct the Young's modulus image. Reconstruction using only a \cann\ is done by setting a constant strain vector $\bm{\varepsilon} = [0.003\ 0.005\ 0.0001]$ and varying $(x,y)$ over the domain of the mesh. At each $(x,y)$, a corresponding stress $\bm{\sigma}$ was computed. Axial and lateral components of the input strain vector ($\varepsilon_{11}$ and $\varepsilon_{22}$, respectively) and the axial component of the computed stress ($\sigma_{22}$) were used to compute the spatially varying Young's modulus $E(x,y)$ by inverting the plane-stress equation:
\begin{flalign}
E(x,y) = \frac{\sigma_{22}(1-\nu^2)}{\nu\varepsilon_{11} + \varepsilon_{22}}
\end{flalign}
where $\nu = 0.5$. The choice of strain vector is arbitrary so long as it resides within the range of training data. Selecting small values for each component ensured the strain was within said range.

\begin{figure}
	\captionsetup[subfigure]{justification=centering}
	\centering
	\begin{subfigure}{0.23\textwidth}
		\includegraphics[width=\textwidth]{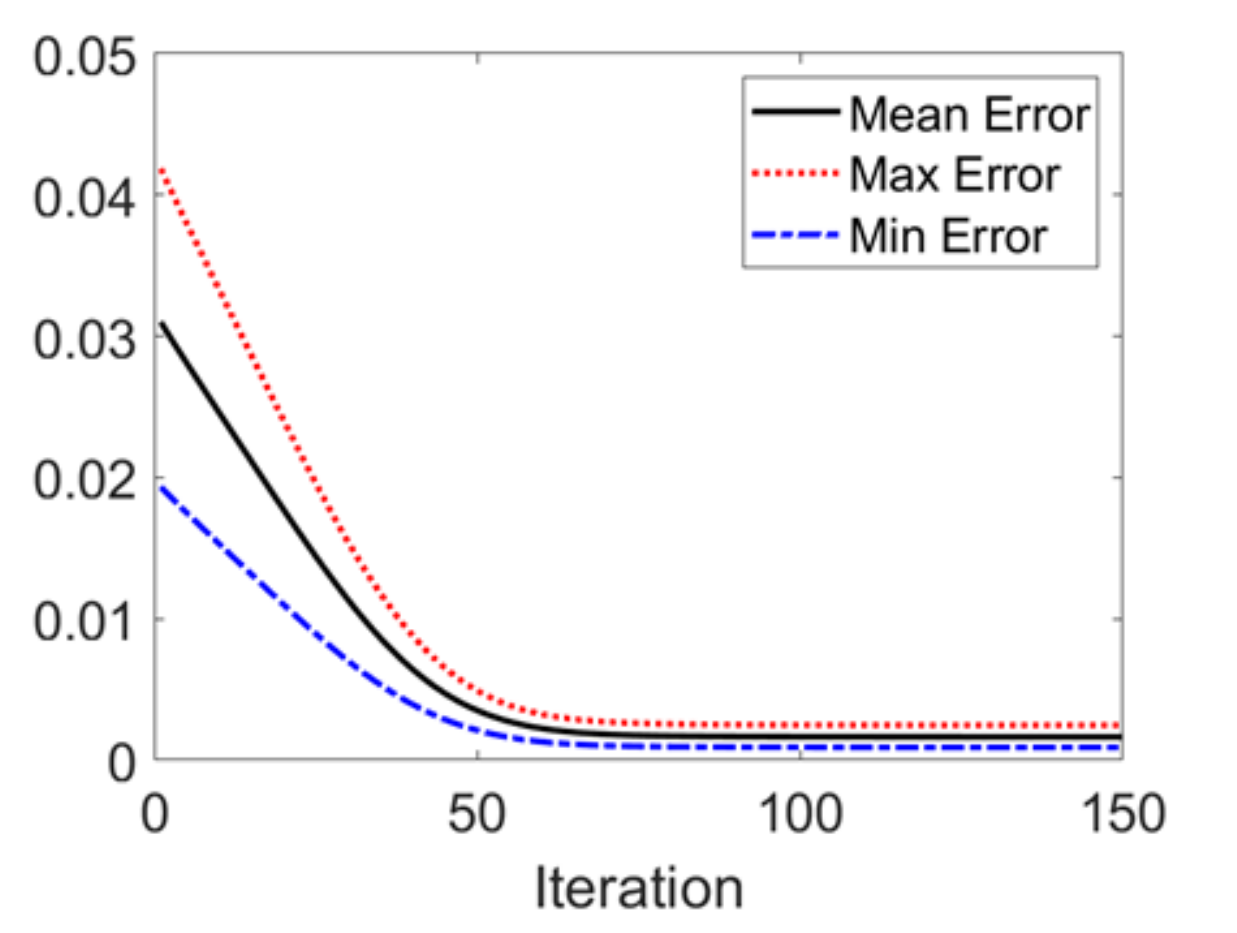}
	\end{subfigure}
	\hfill
	\begin{subfigure}{0.23\textwidth}
		\includegraphics[width=\textwidth]{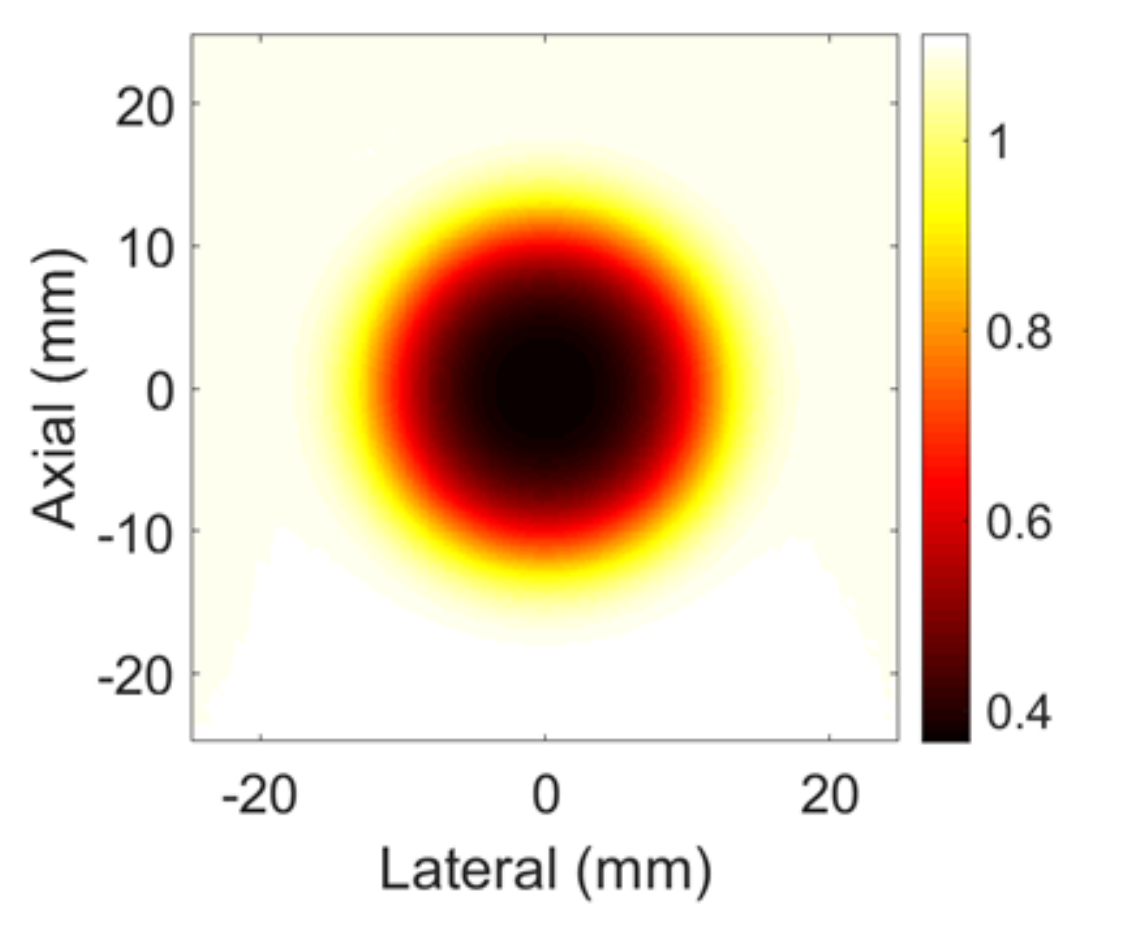}
	\end{subfigure}
	\hfill
	\begin{subfigure}{0.23\textwidth}
		\includegraphics[width=\textwidth]{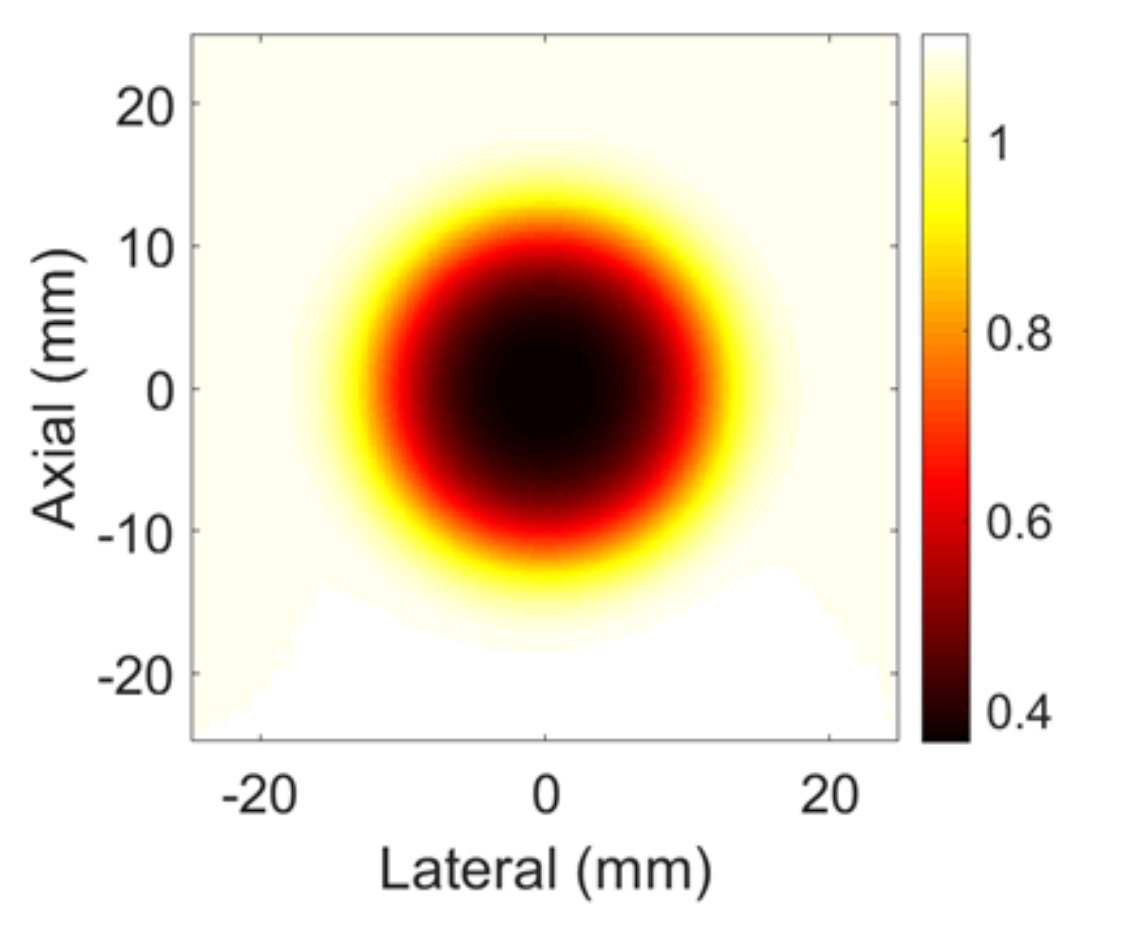}
	\end{subfigure}
	\hfill
	\begin{subfigure}{0.23\textwidth}
		\includegraphics[width=\textwidth]{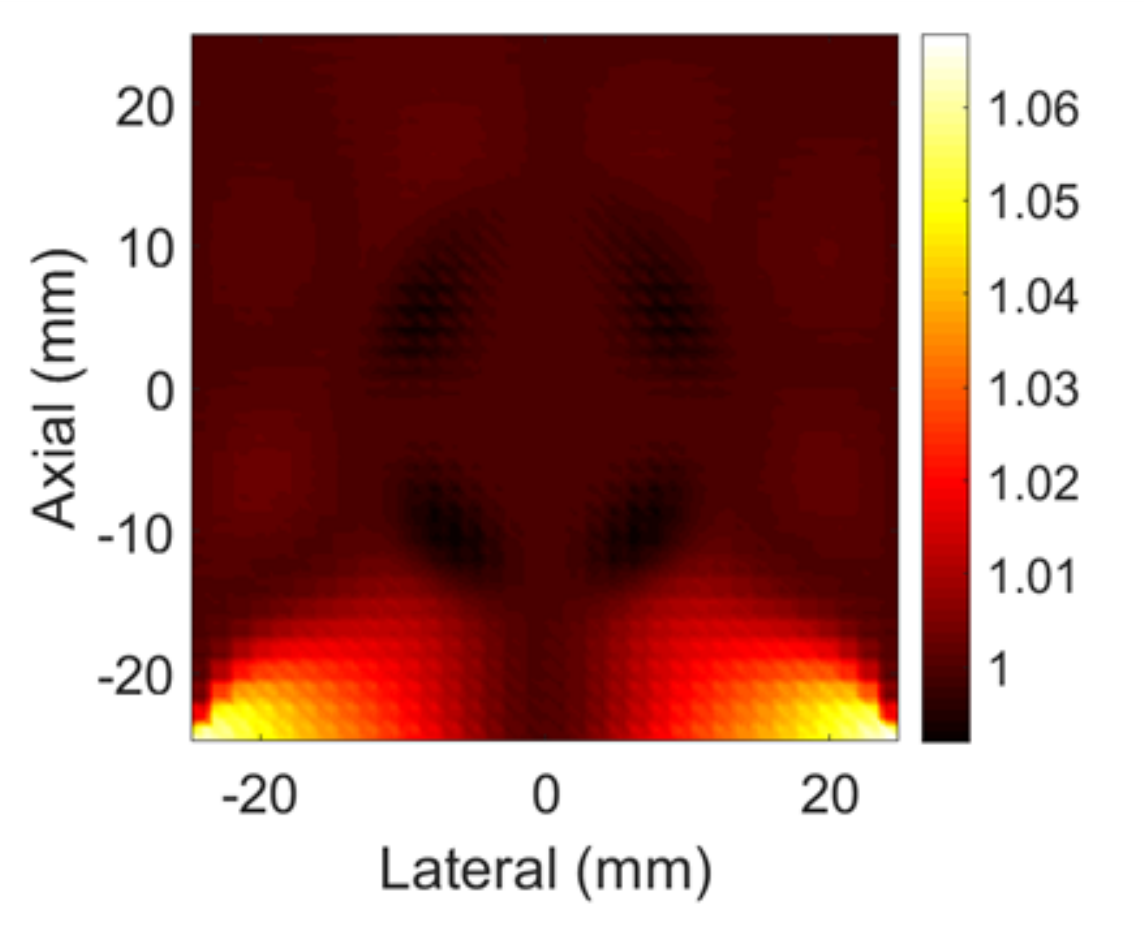}
	\end{subfigure}
	\vfill
	\begin{subfigure}{0.23\textwidth}
		\includegraphics[width=\textwidth]{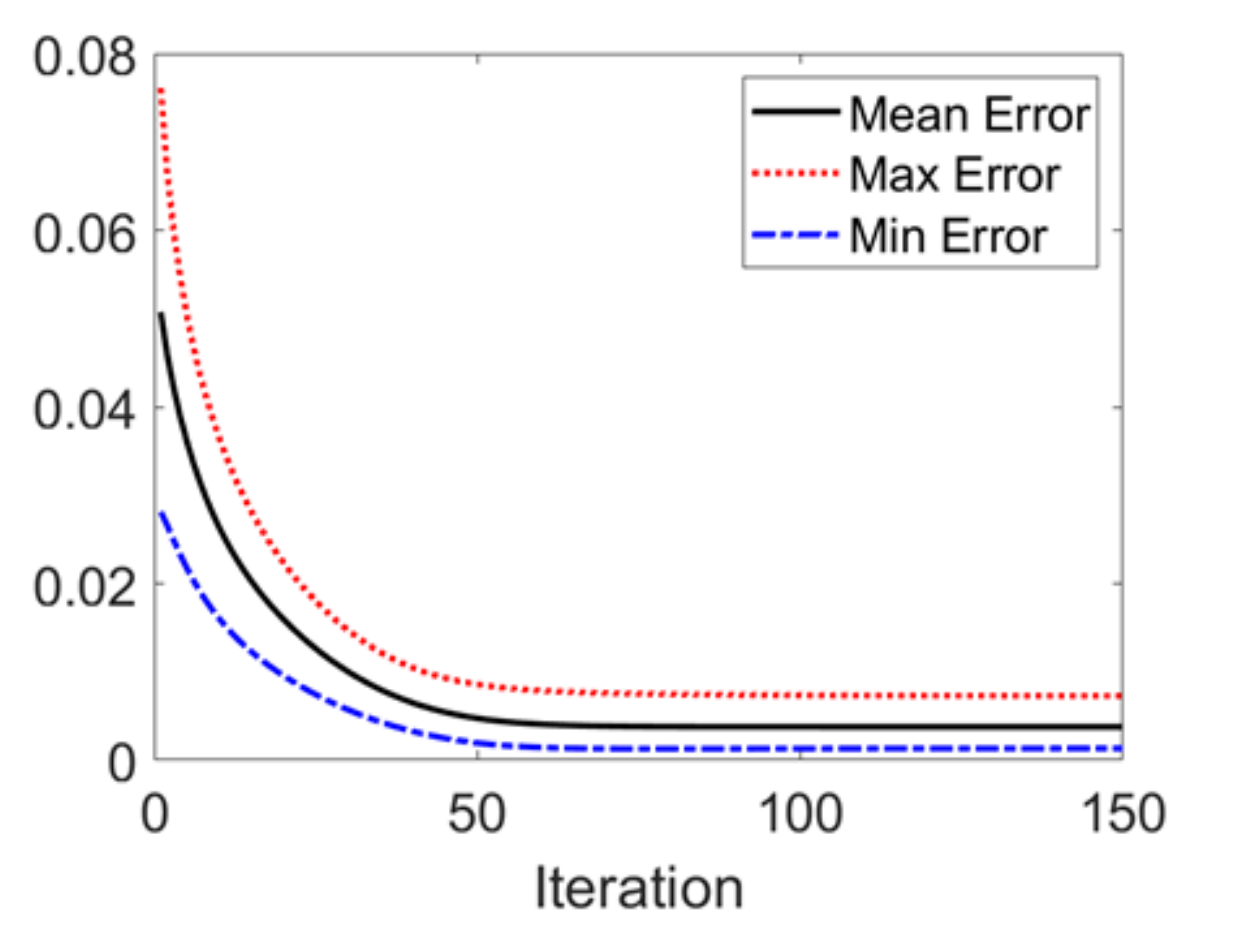}
	\end{subfigure}
	\hfill
	\begin{subfigure}{0.23\textwidth}
		\includegraphics[width=\textwidth]{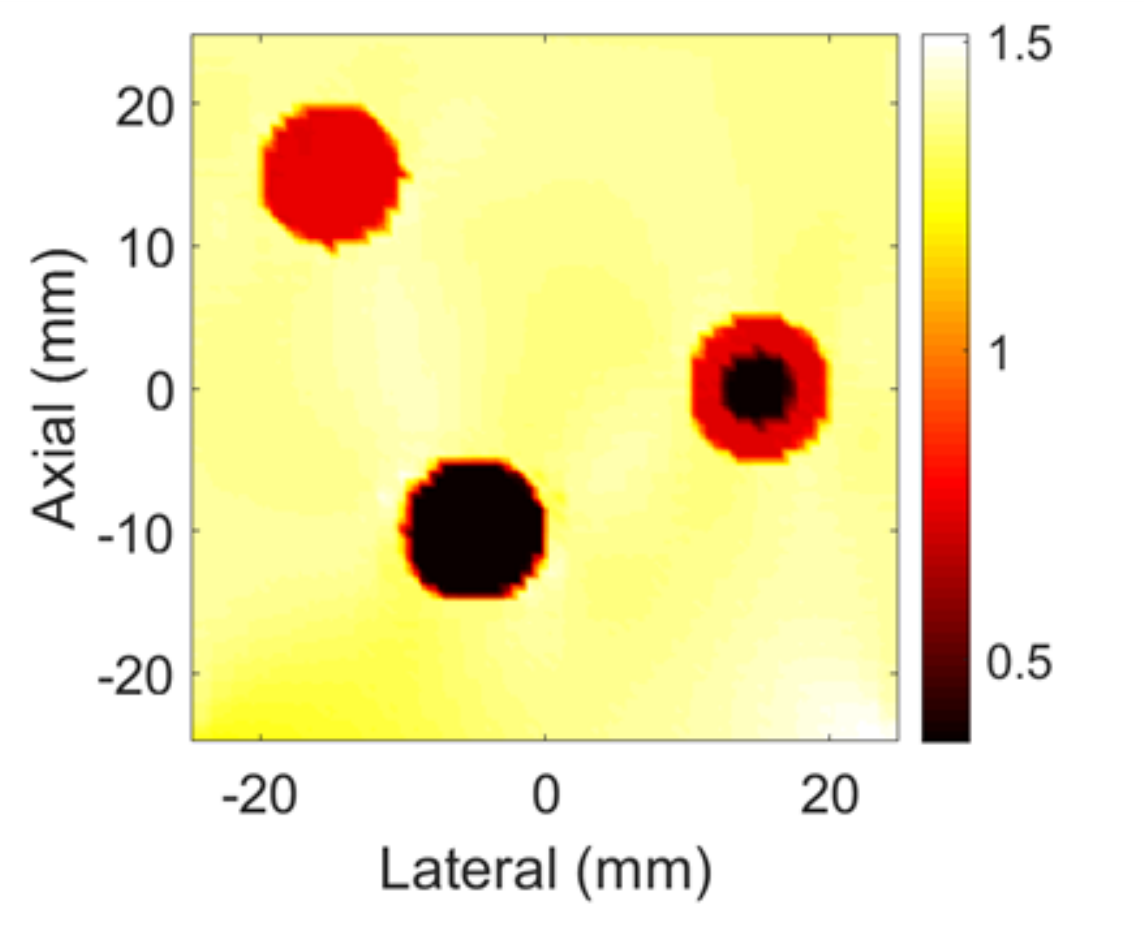}
	\end{subfigure}
	\hfill
	\begin{subfigure}{0.23\textwidth}
		\includegraphics[width=\textwidth]{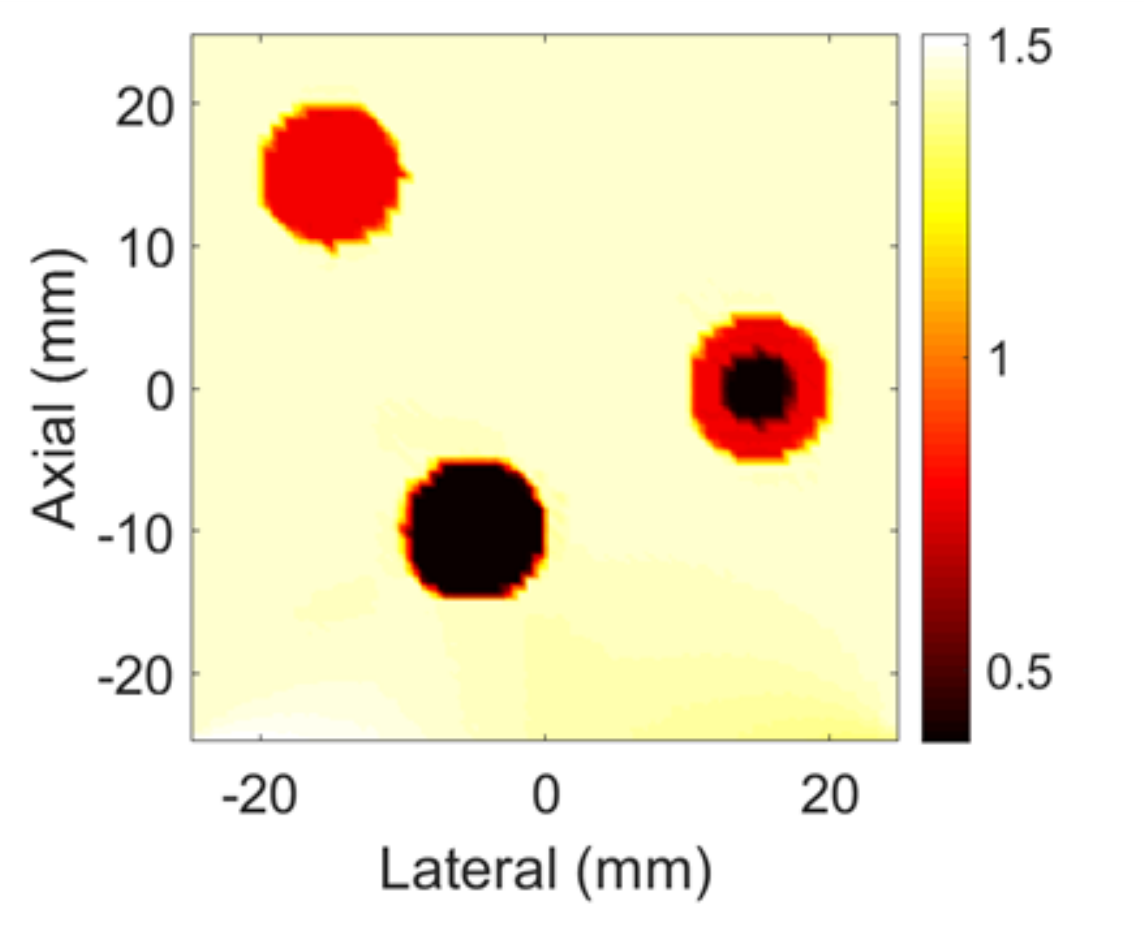}
	\end{subfigure}
	\hfill
	\begin{subfigure}{0.23\textwidth}
		\includegraphics[width=\textwidth]{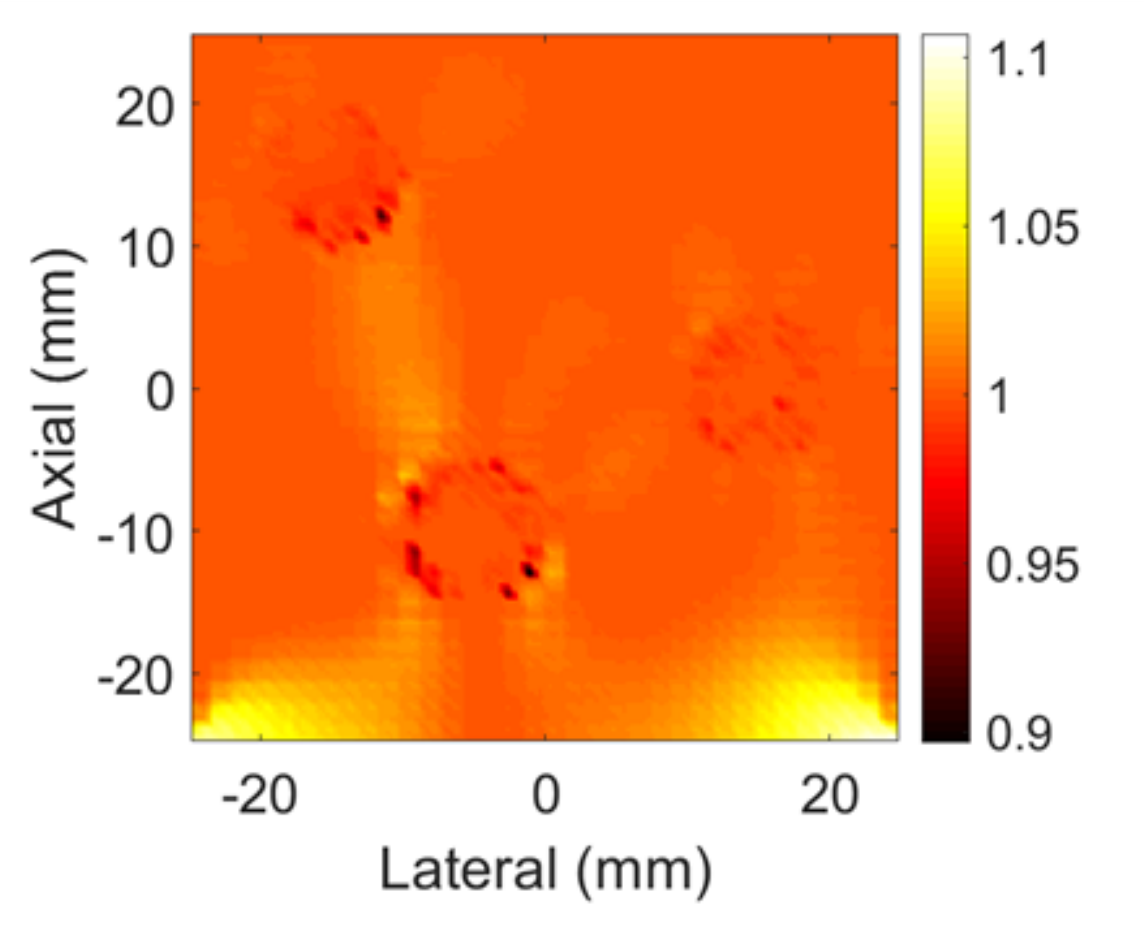}
	\end{subfigure}
	\vfill
	\begin{subfigure}{0.23\textwidth}
		\includegraphics[width=\textwidth]{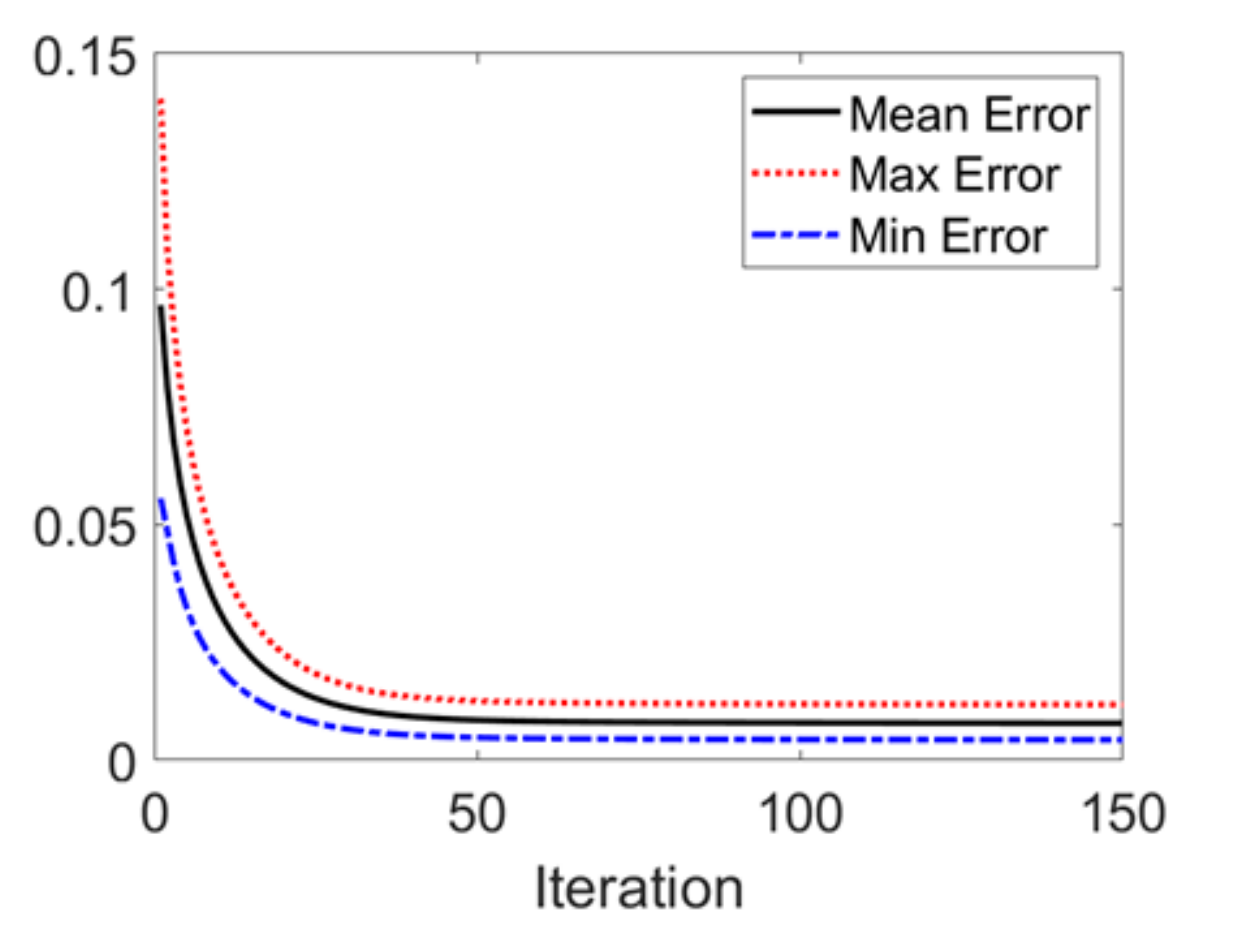}
	\end{subfigure}
	\hfill
	\begin{subfigure}{0.23\textwidth}
		\includegraphics[width=\textwidth]{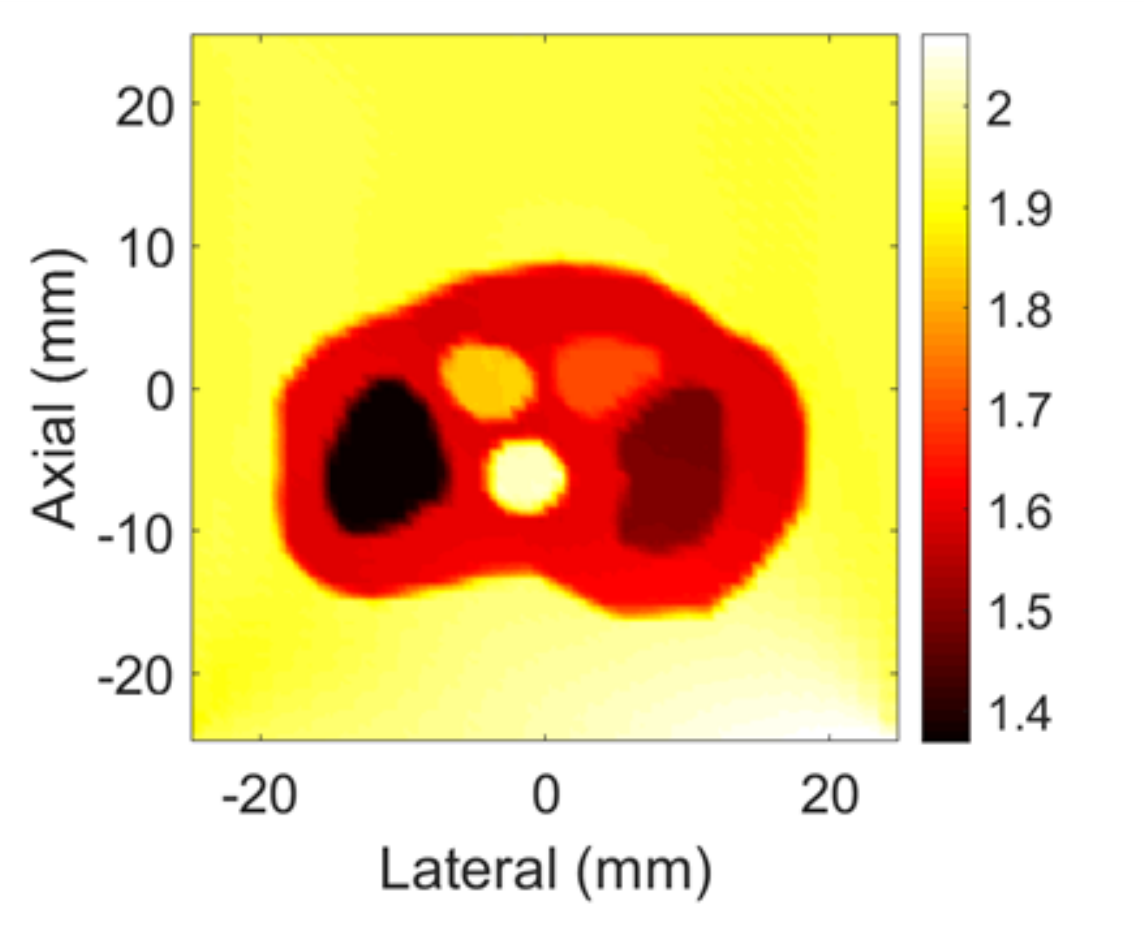}
	\end{subfigure}
	\hfill
	\begin{subfigure}{0.23\textwidth}
		\includegraphics[width=\textwidth]{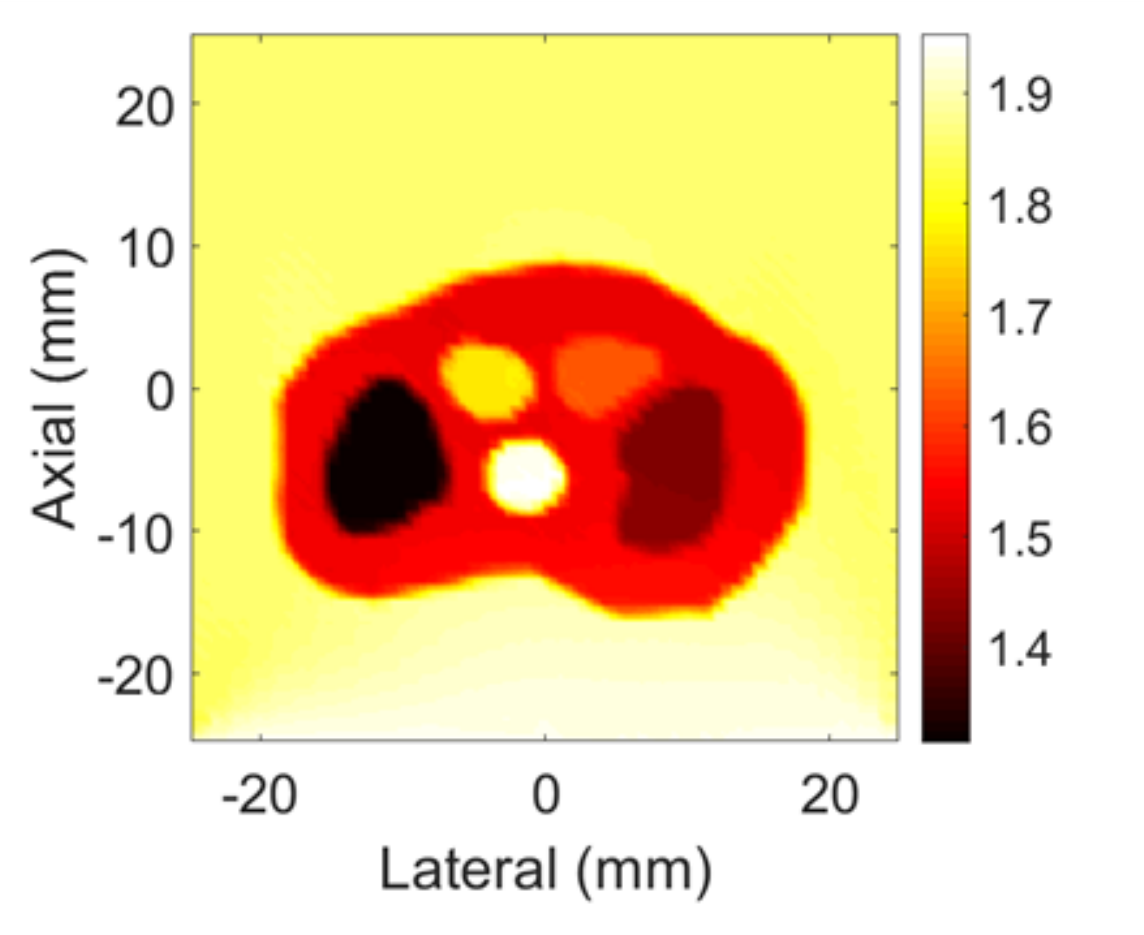}
	\end{subfigure}
	\hfill
	\begin{subfigure}{0.23\textwidth}
		\includegraphics[width=\textwidth]{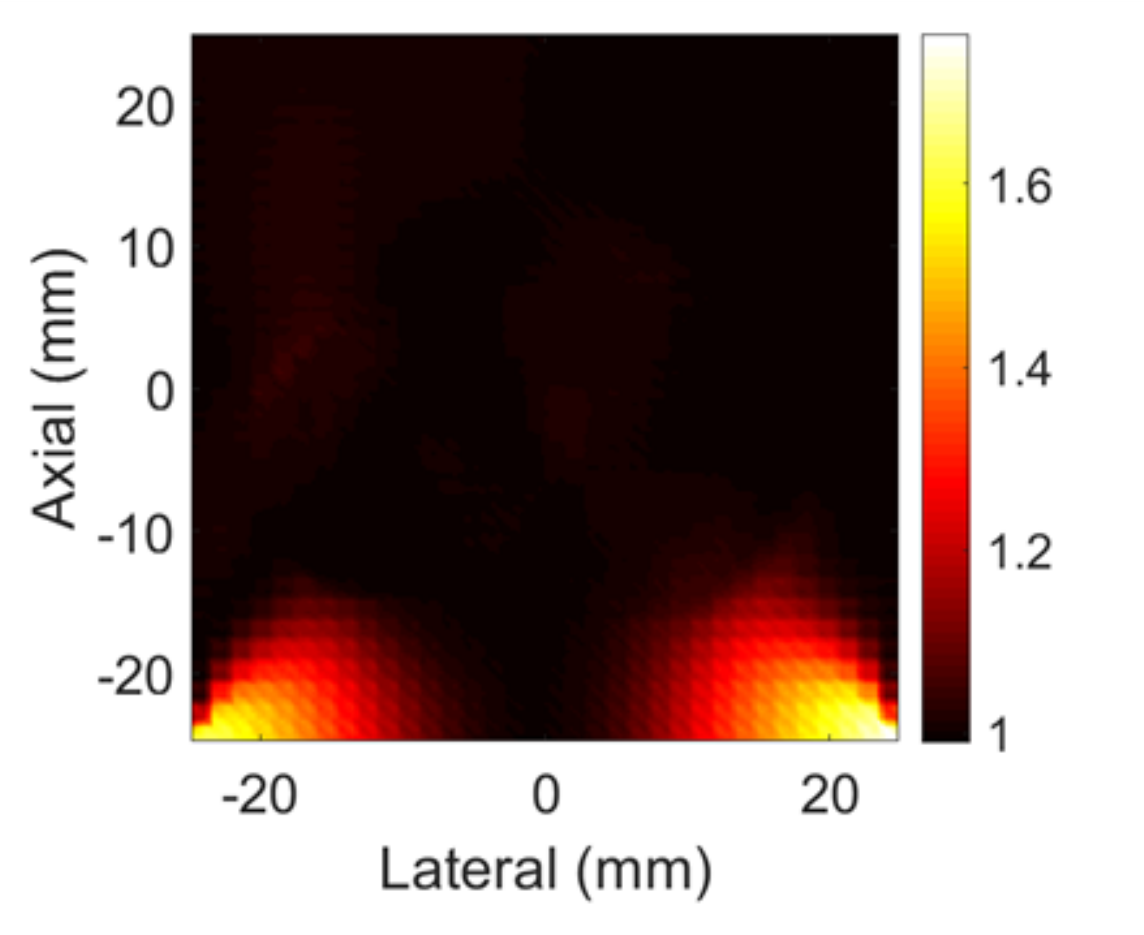}
	\end{subfigure}
	\vfill
	\begin{subfigure}{0.23\textwidth}
		\includegraphics[width=\textwidth]{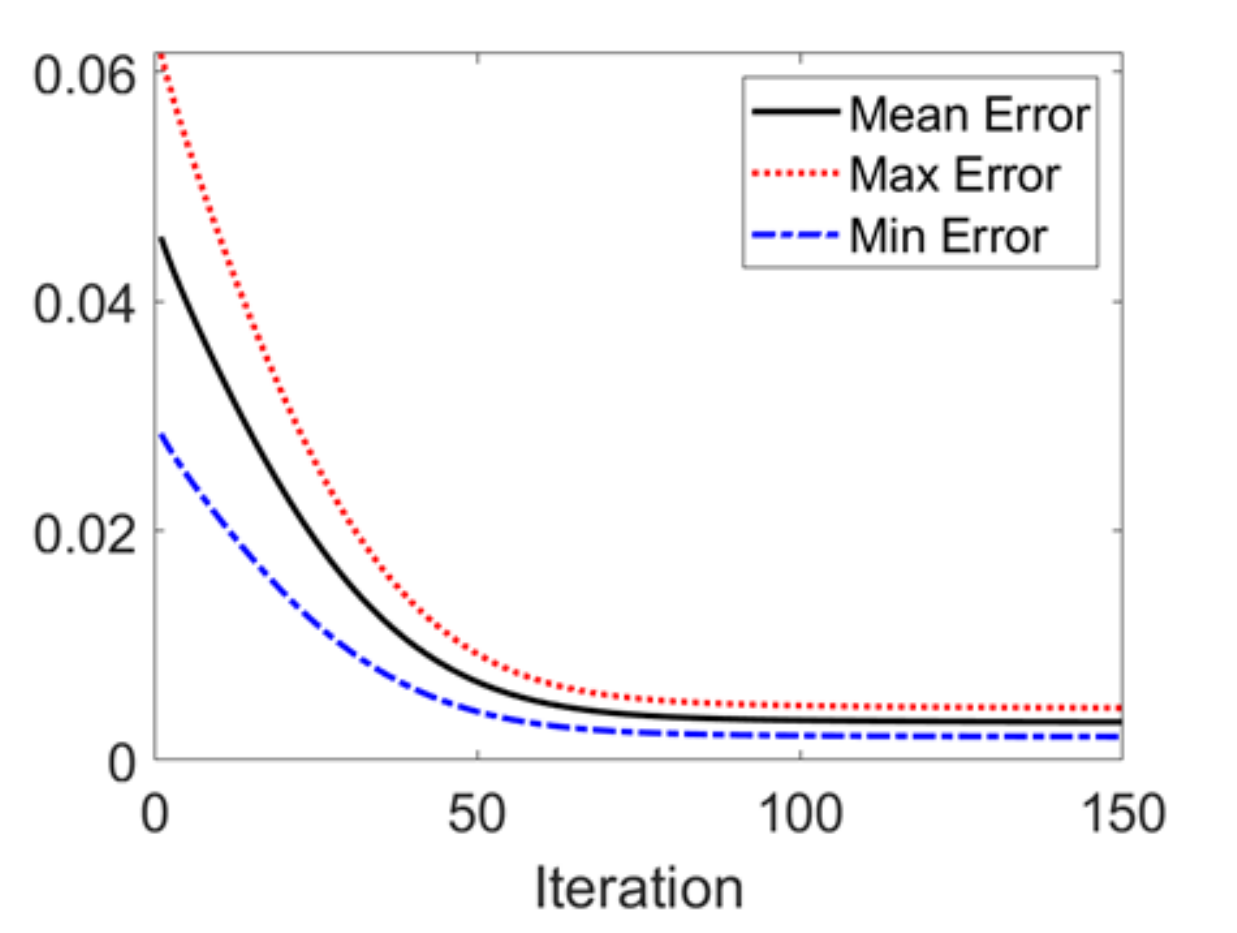}
	\end{subfigure}
	\hfill
	\begin{subfigure}{0.23\textwidth}
		\includegraphics[width=\textwidth]{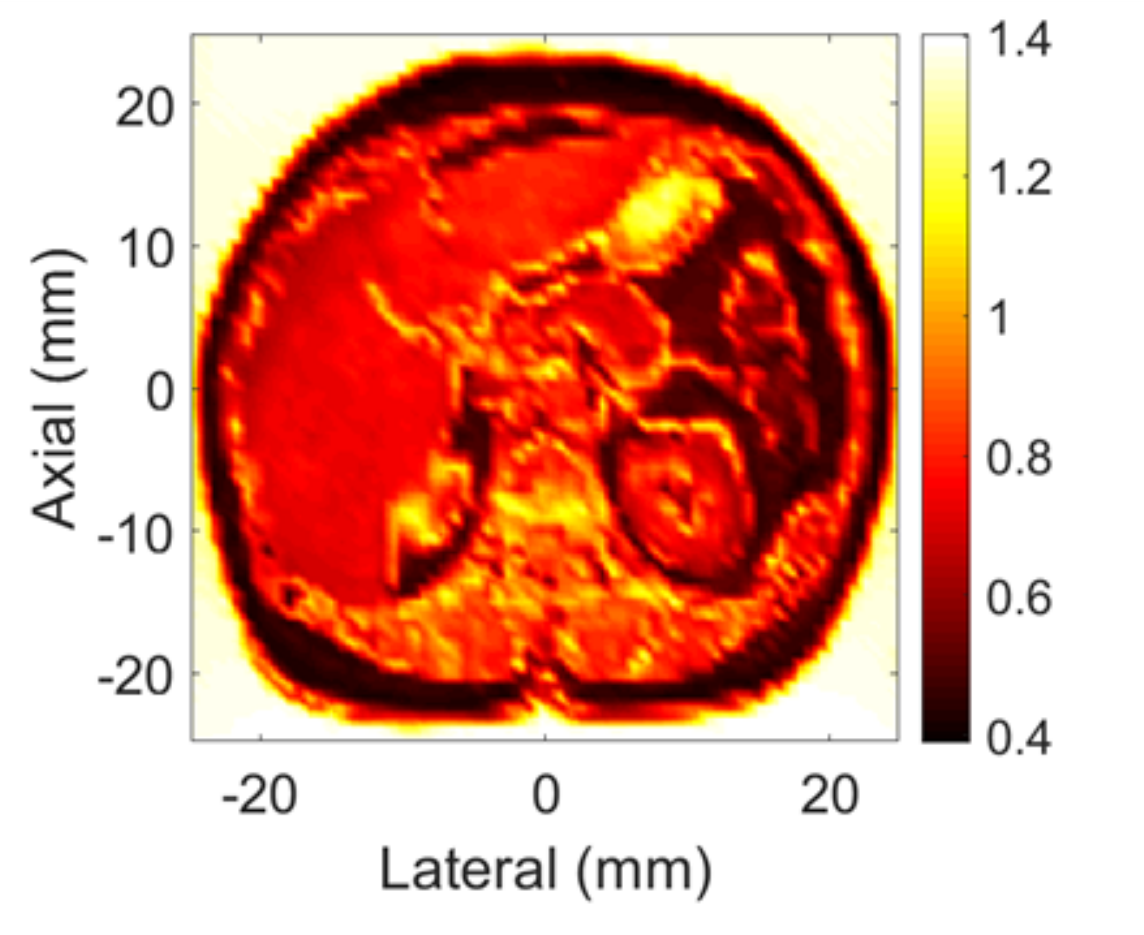}
	\end{subfigure}
	\hfill
	\begin{subfigure}{0.23\textwidth}
		\includegraphics[width=\textwidth]{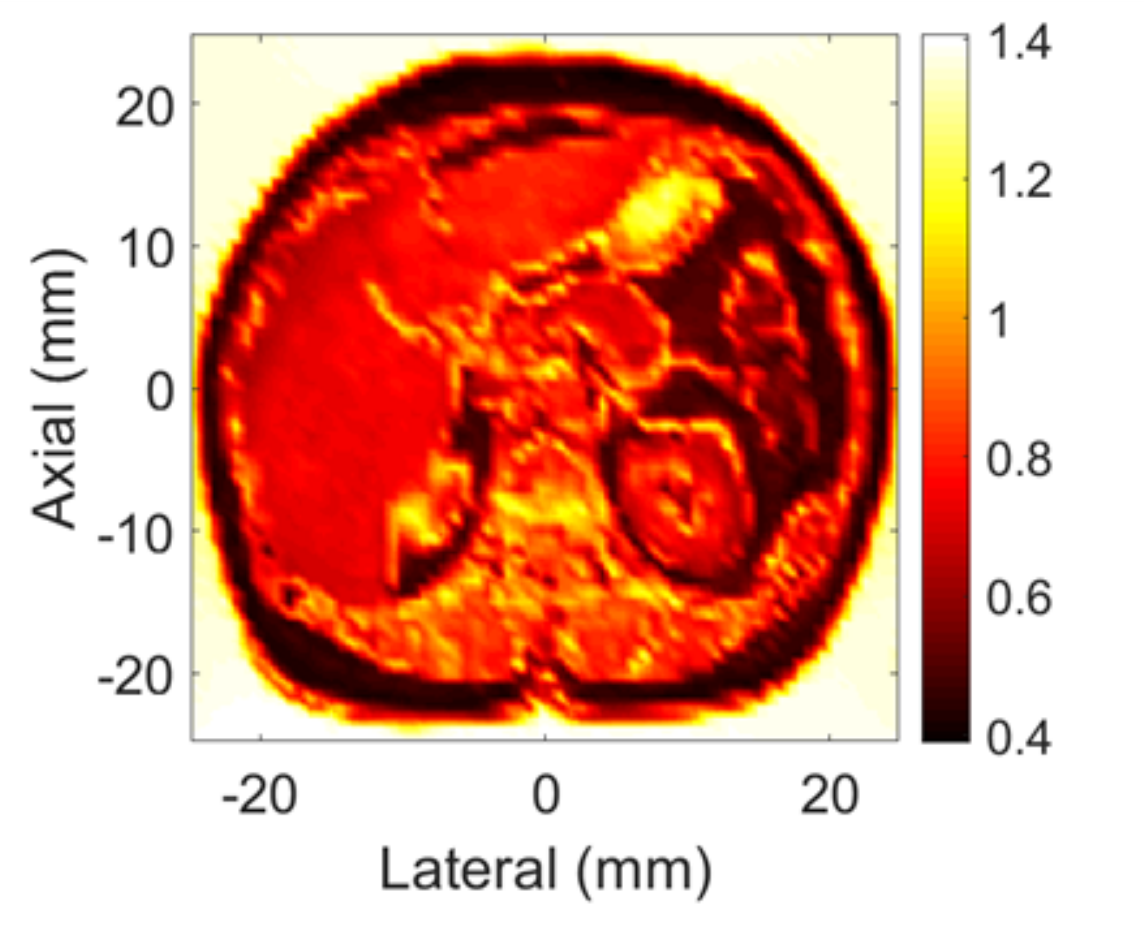}
	\end{subfigure}
	\hfill
	\begin{subfigure}{0.23\textwidth}
		\includegraphics[width=\textwidth]{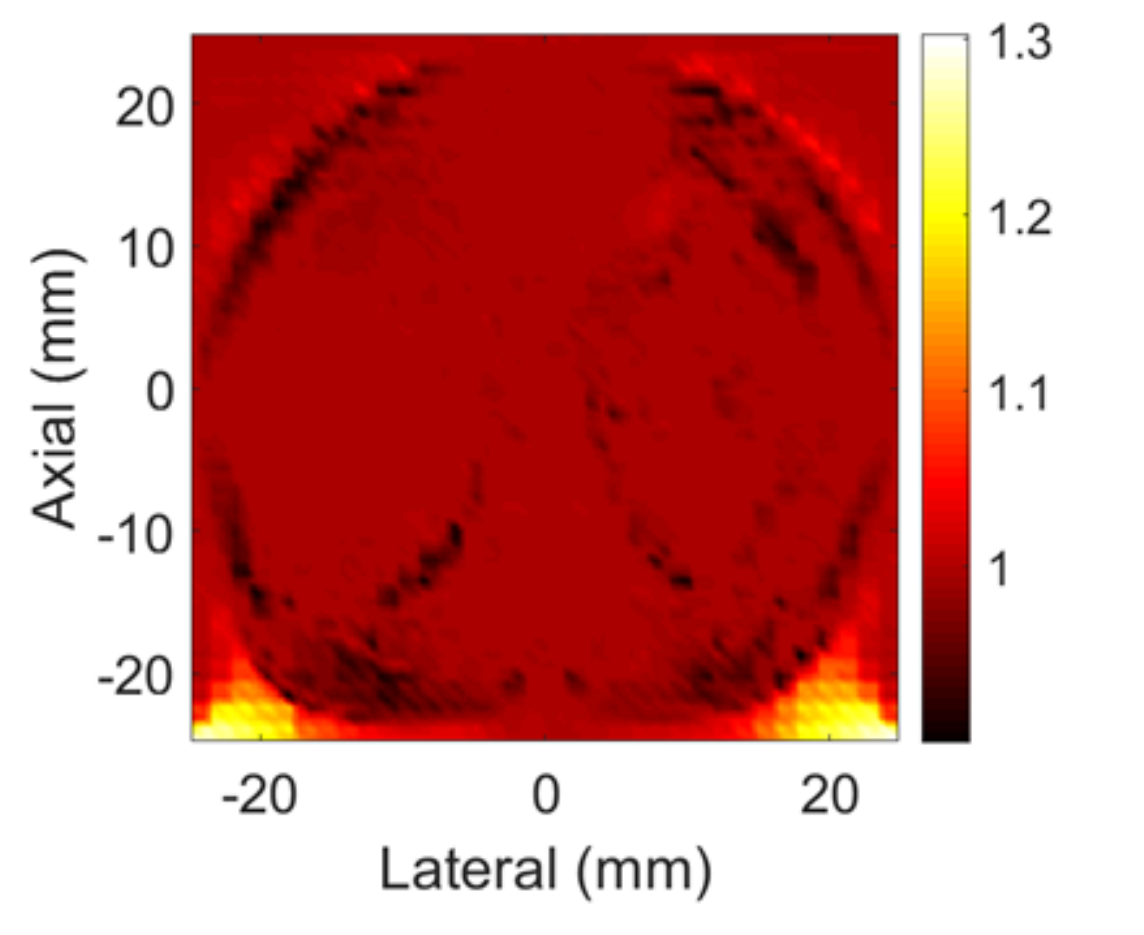}
	\end{subfigure}
	\vfill
	\begin{subfigure}{0.23\textwidth}
		\includegraphics[width=\textwidth]{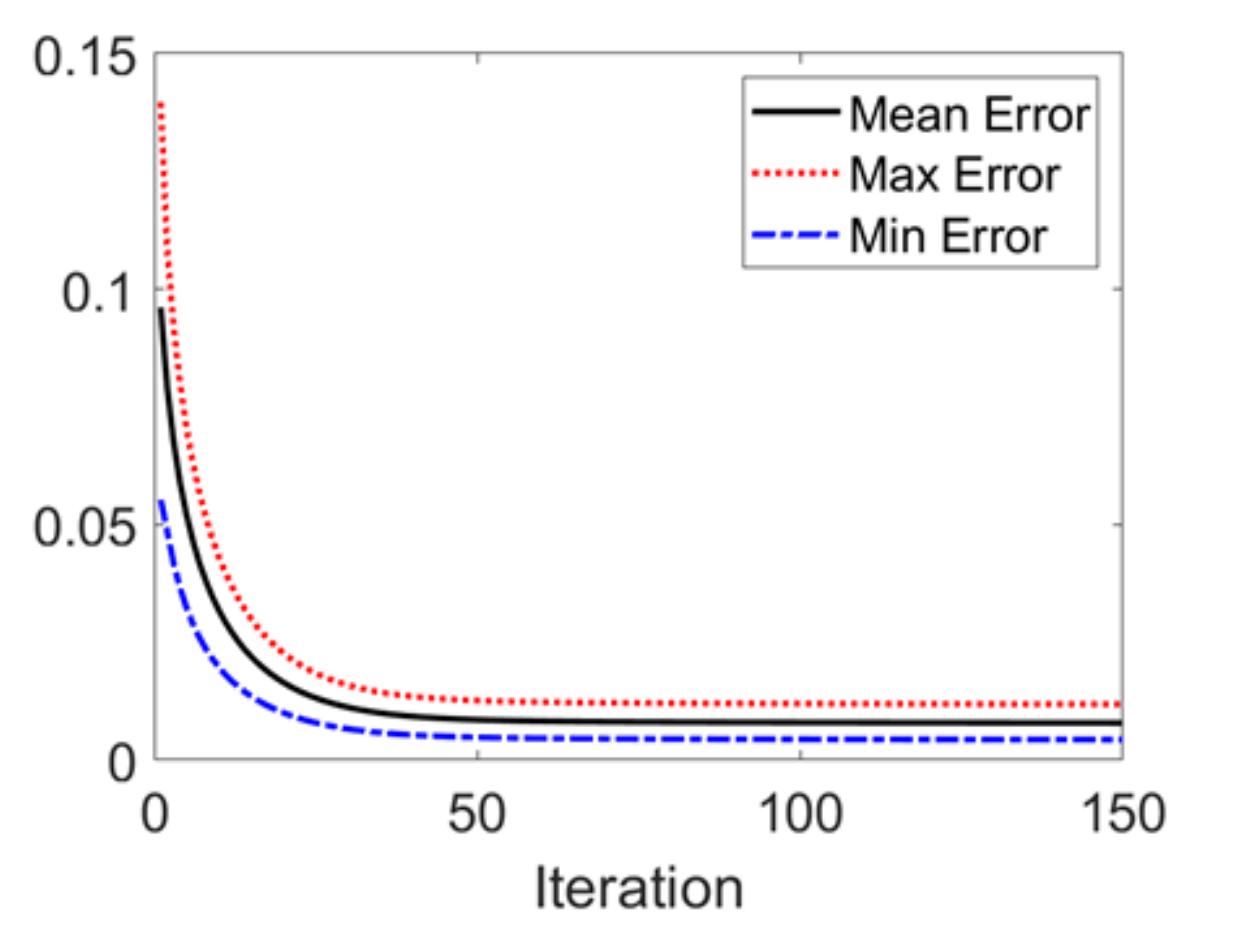}
	\end{subfigure}
	\hfill
	\begin{subfigure}{0.23\textwidth}
		\includegraphics[width=\textwidth]{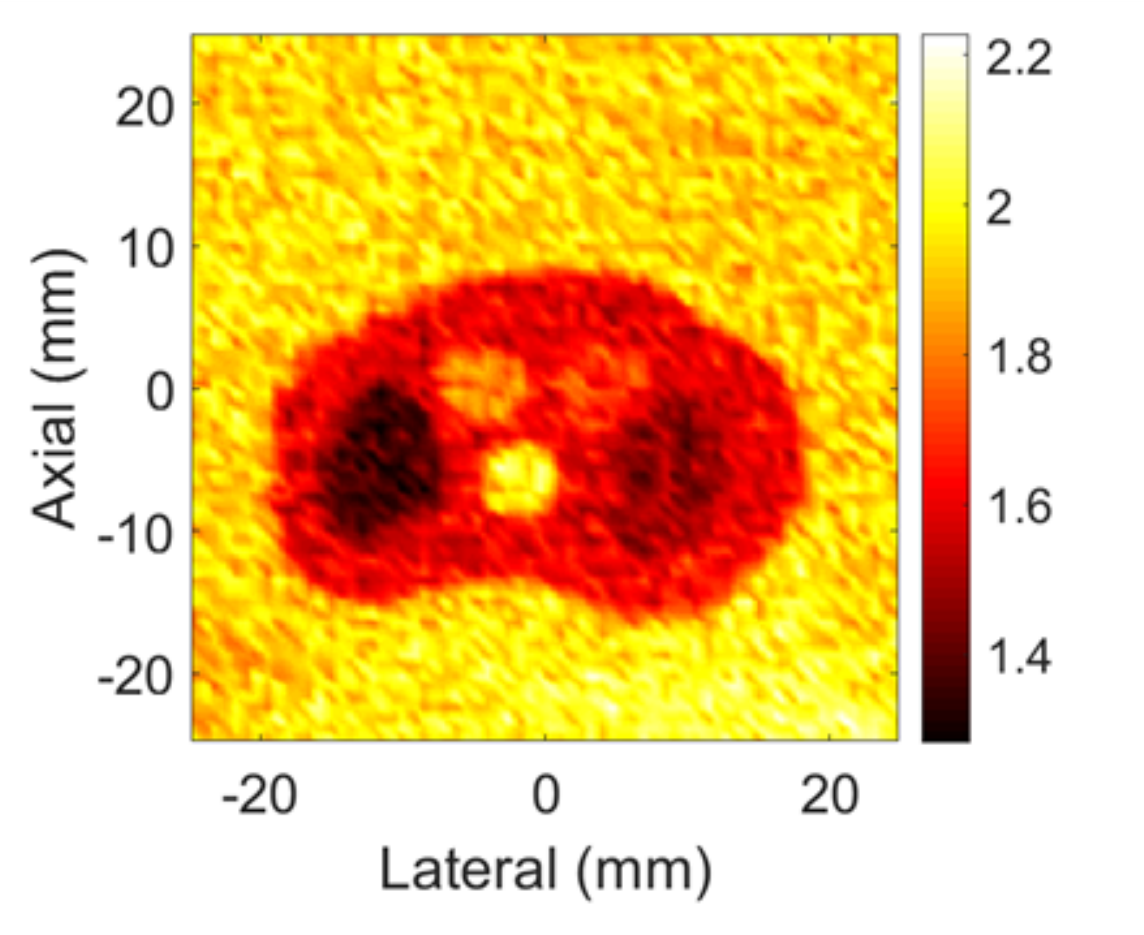}
	\end{subfigure}
	\hfill
	\begin{subfigure}{0.23\textwidth}
		\includegraphics[width=\textwidth]{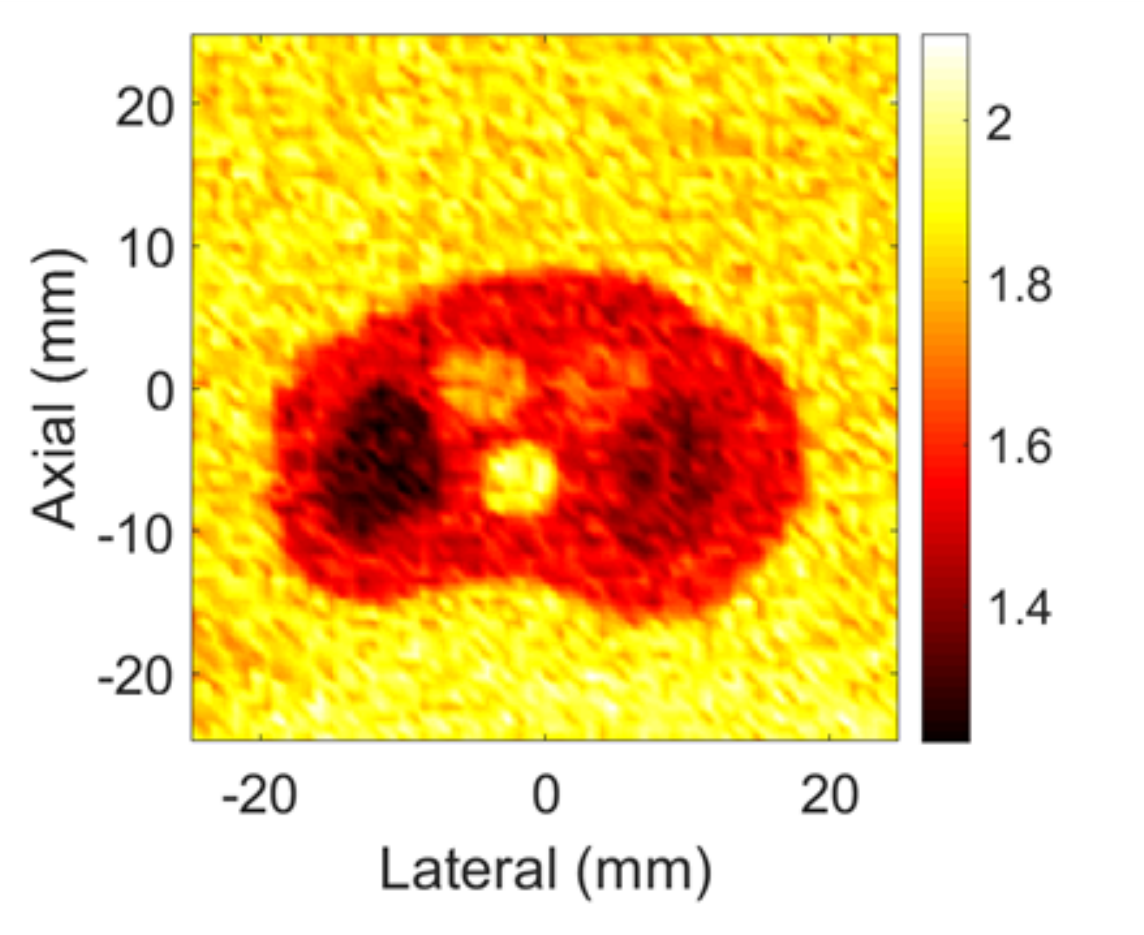}
	\end{subfigure}
	\hfill
	\begin{subfigure}{0.23\textwidth}
		\includegraphics[width=\textwidth]{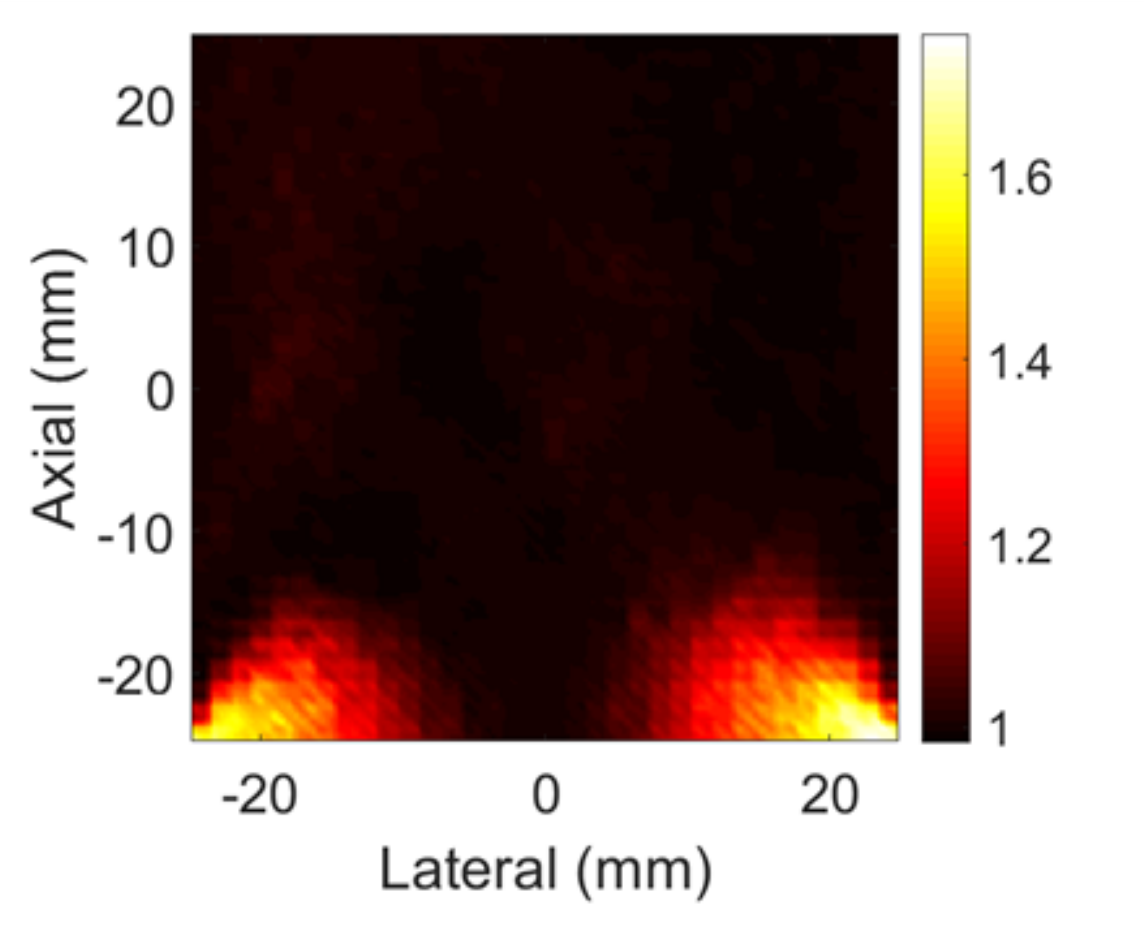}
	\end{subfigure}
	\vfill
	\begin{subfigure}{0.23\textwidth}
		\includegraphics[width=\textwidth]{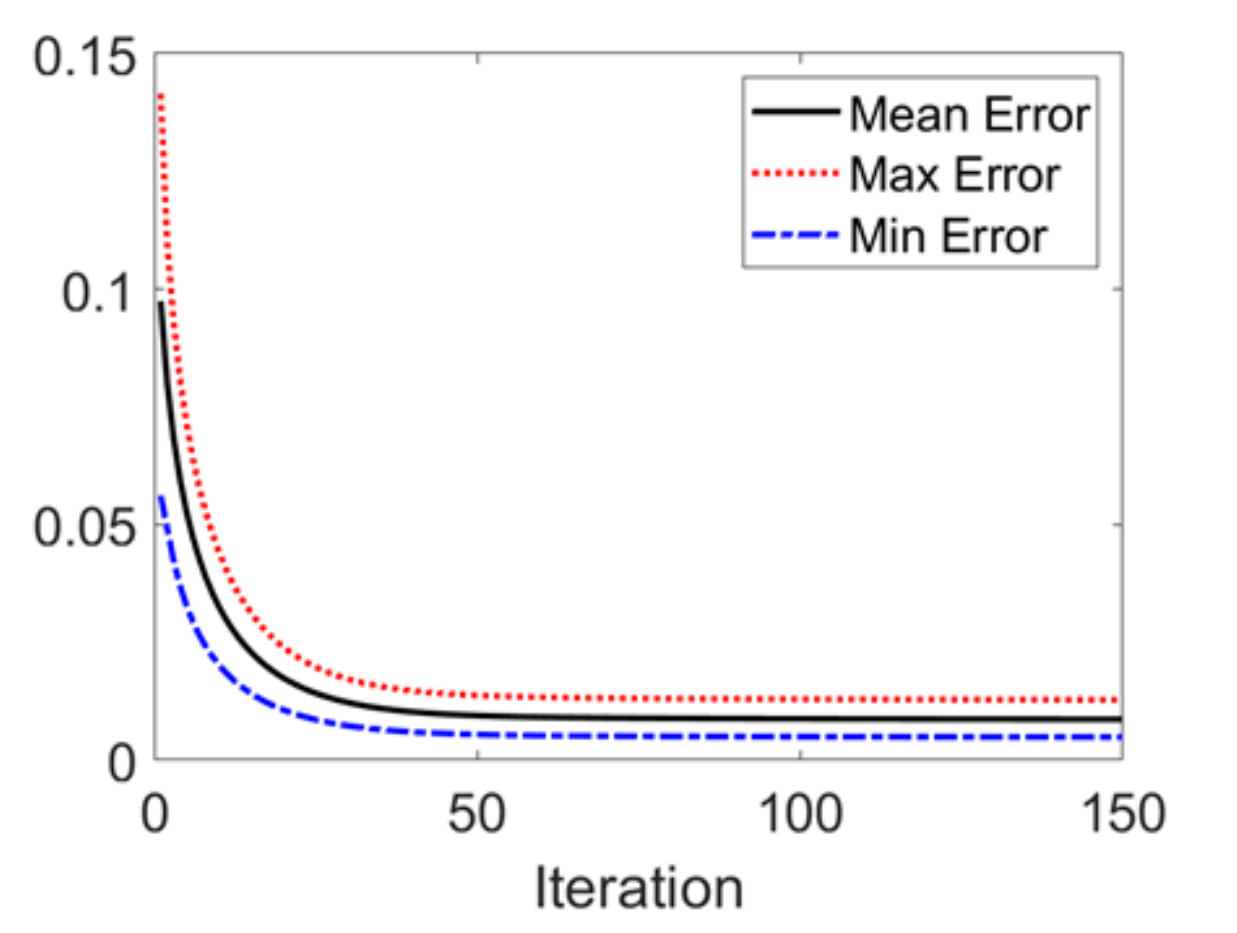}
	\end{subfigure}
	\hfill
	\begin{subfigure}{0.23\textwidth}
		\includegraphics[width=\textwidth]{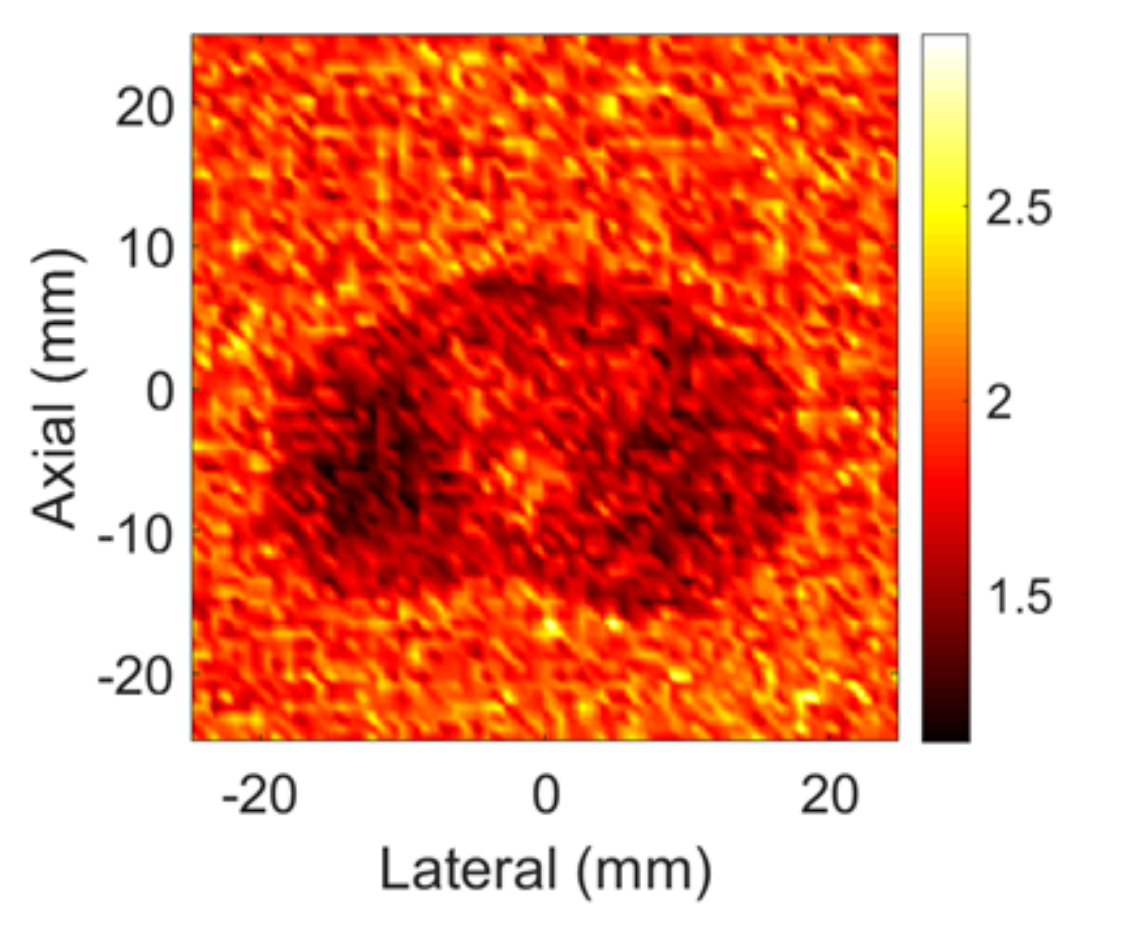}
	\end{subfigure}
	\hfill
	\begin{subfigure}{0.23\textwidth}
		\includegraphics[width=\textwidth]{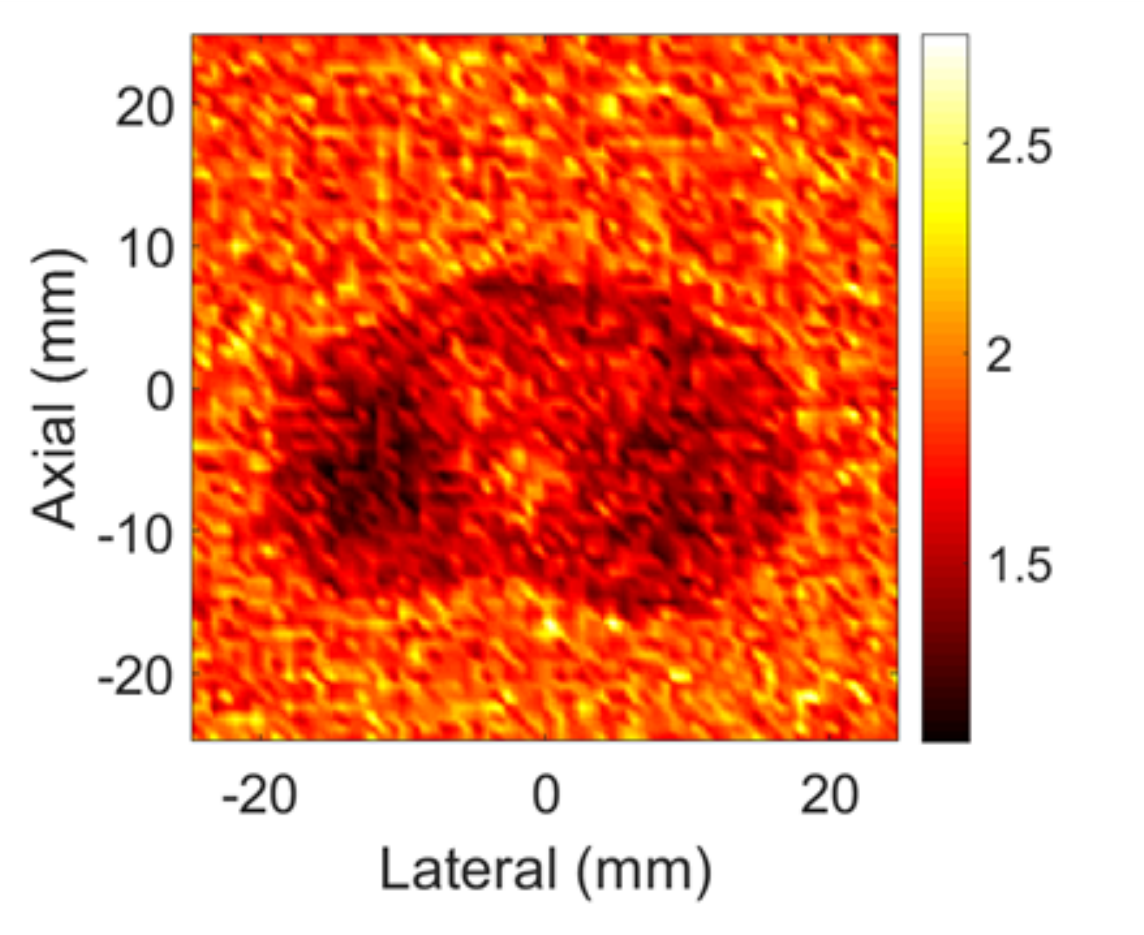}
	\end{subfigure}
	\hfill
	\begin{subfigure}{0.23\textwidth}
		\includegraphics[width=\textwidth]{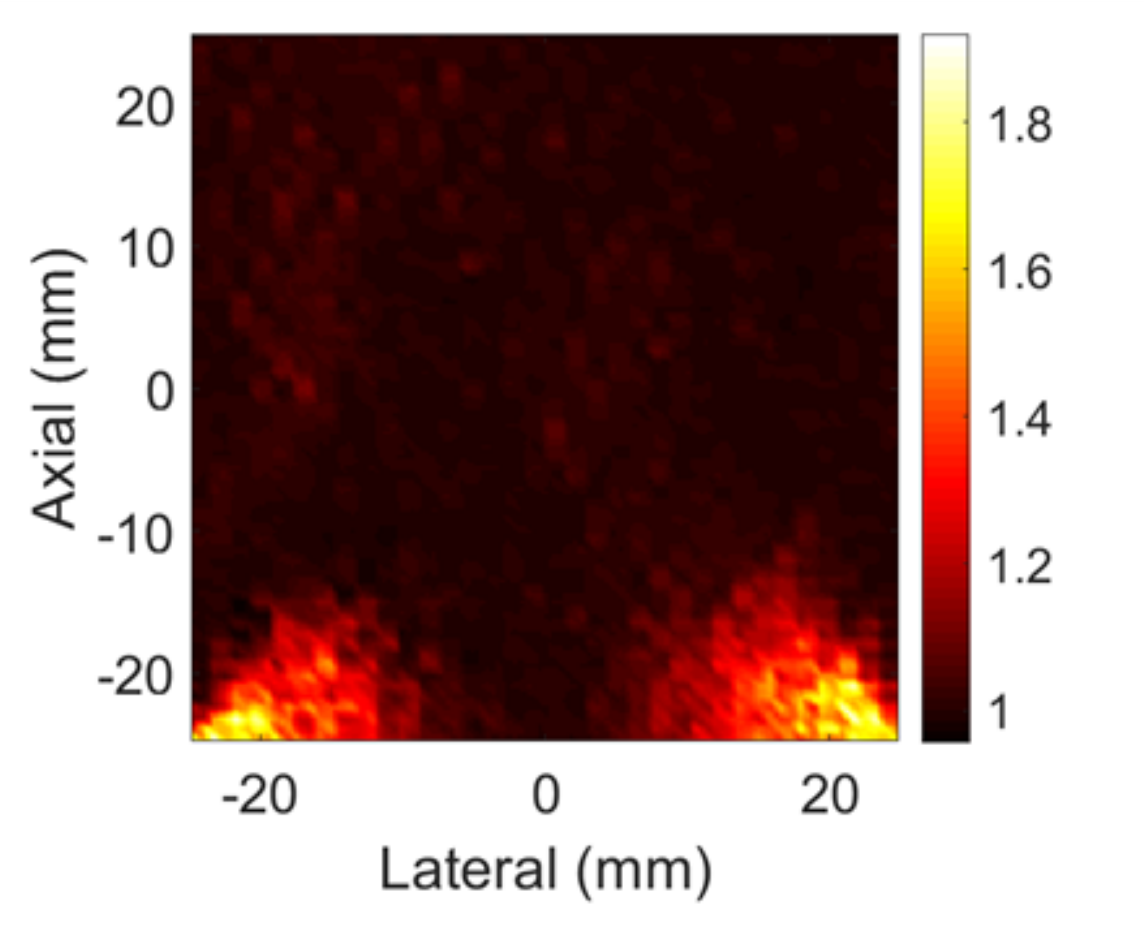}
	\end{subfigure}
	\vfill
	\begin{subfigure}{0.23\textwidth}
		\includegraphics[width=\textwidth]{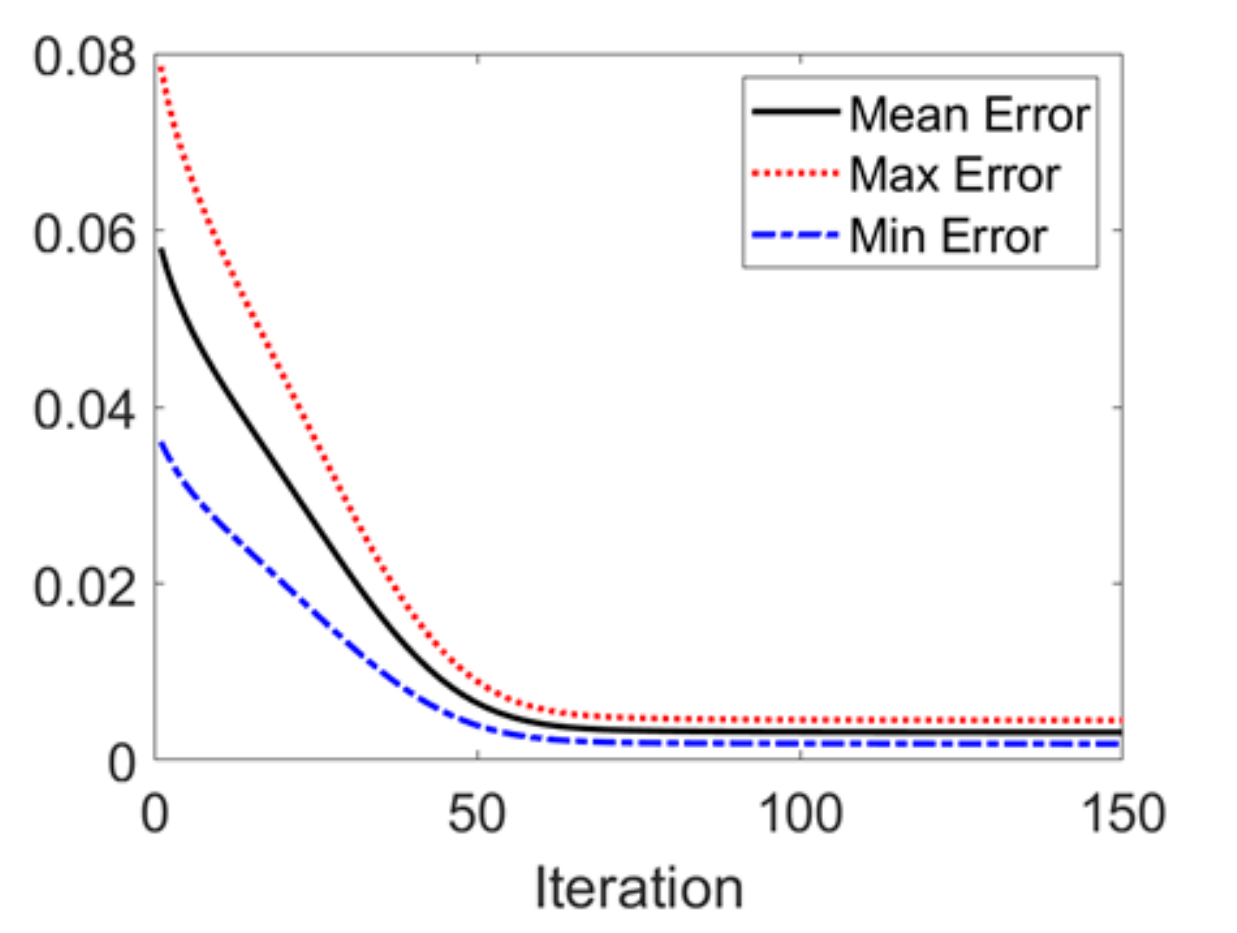}
	\end{subfigure}
	\hfill
	\begin{subfigure}{0.23\textwidth}
		\includegraphics[width=\textwidth]{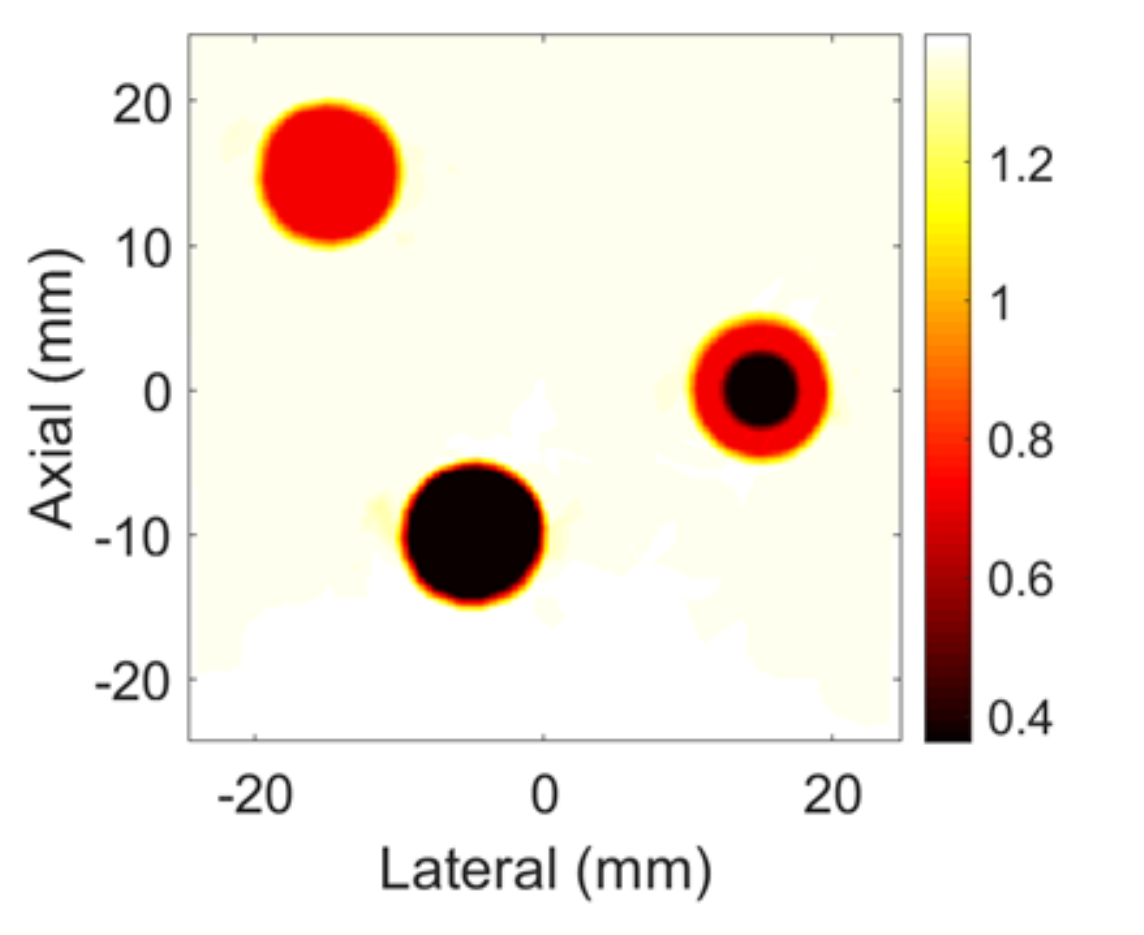}
	\end{subfigure}
	\hfill
	\begin{subfigure}{0.23\textwidth}
		\includegraphics[width=\textwidth]{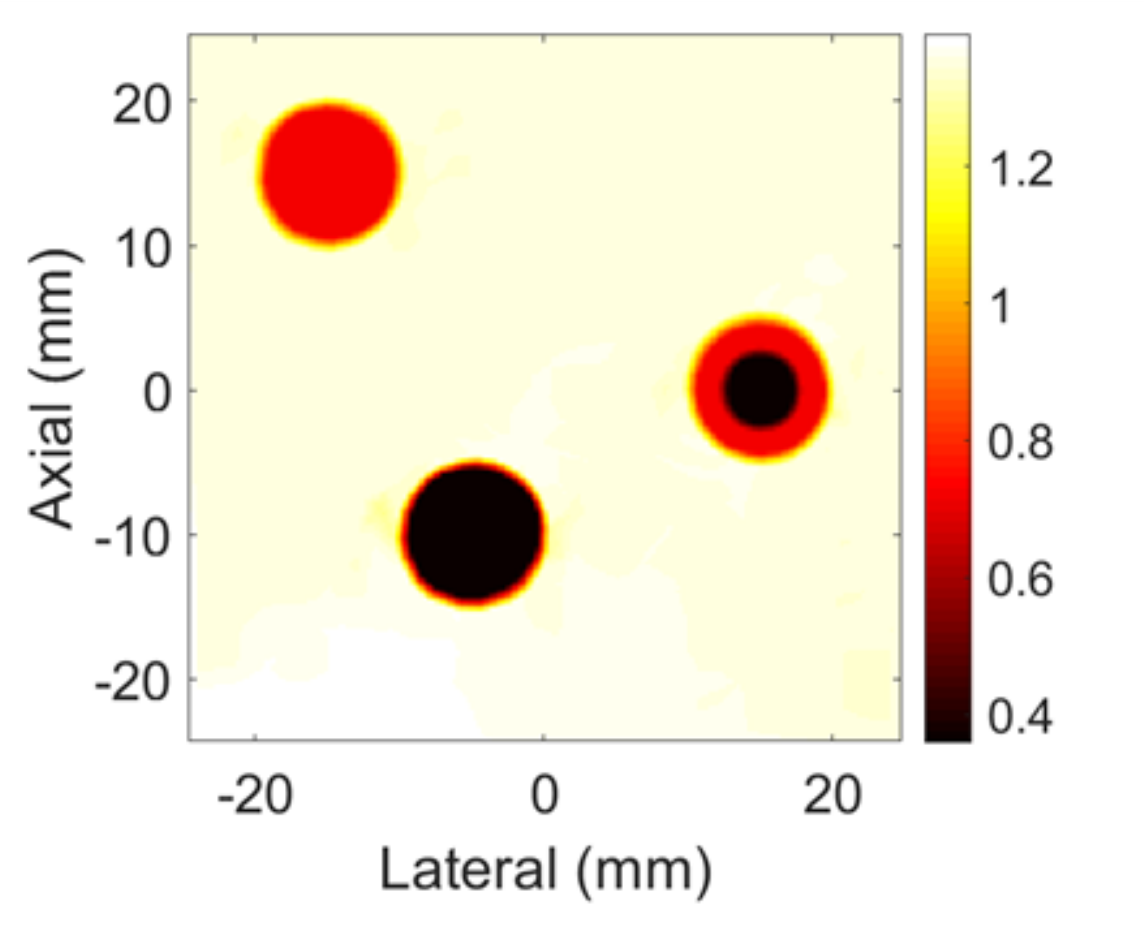}
	\end{subfigure}
	\hfill
	\begin{subfigure}{0.23\textwidth}
		\includegraphics[width=\textwidth]{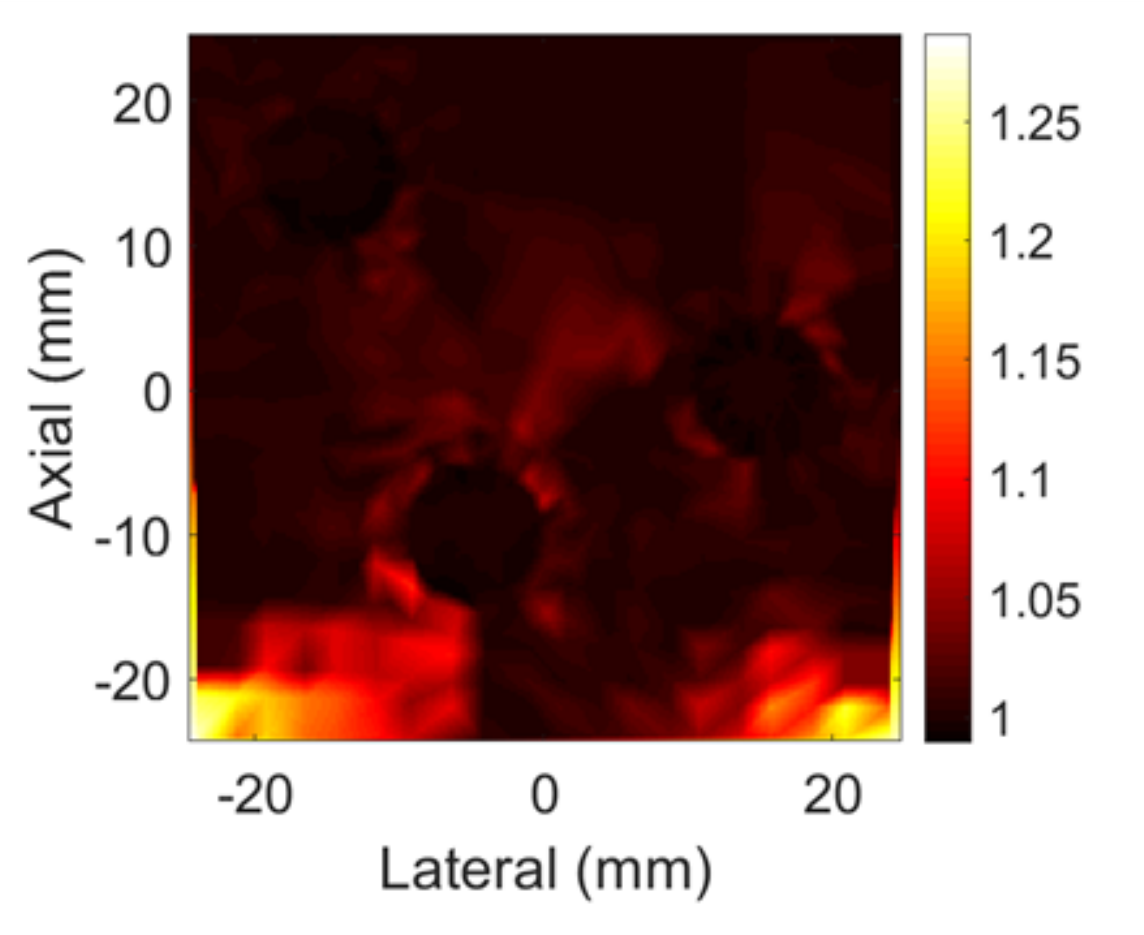}
	\end{subfigure}
	\caption{The left-most column contains the minimum, maximum, and mean RMS error curves from Alg.~1 while computing the values of \spatscale\ for each of the seven cases. Columns 2-4 are the resulting $S_{\bm{x}}^{\varepsilon_1}$, $S_{\bm{x}}^{\varepsilon_2}$, and $S_{\bm{x}}^{\varepsilon_3}$ distributions, respectively.}
	\label{fig:computed_scaling}
\end{figure}

\section{Results}
Fig.~\ref{fig:computed_scaling} contains the results of computing the spatial scaling values for all seven cases in this study. Plots in the left-most column are the mean, minimum, and maximum error in each iteration of Alg.~1. Errors are the RMS of the difference between $\bm{\sigma}^t$ and $\sigma_i^{NN}$ over all stress-strain pairs for the model. Columns 2, 3, and 4 are the maps of $S_{\bm{x}}^{\varepsilon_1}$, $S_{\bm{x}}^{\varepsilon_2}$, and $S_{\bm{x}}^{\varepsilon_3}$, respectively. 

The error curves provide insight on the number of iterations required in Alg.~1. When implemented in AutoP, the spatial scaling values will be recomputed many times, meaning there is a trade-off between computation speed and error. From these curves, 50 iterations appears to be sufficient. Images of the computed spatial scaling values, on the other hand, offer intuition on what information is contained with \spatscale. We observe the scaling values for the axial and lateral strains are inversely proportional to the Young's modulus. In the case of linear-elastic materials, the spatial scaling values, and thus the SN, contain information about distribution of the relative stiffness.

\begin{figure}[!h]
	\captionsetup[subfigure]{justification=centering}
	\centering
	\begin{subfigure}{0.23\textwidth}
		\includegraphics[width=\textwidth]{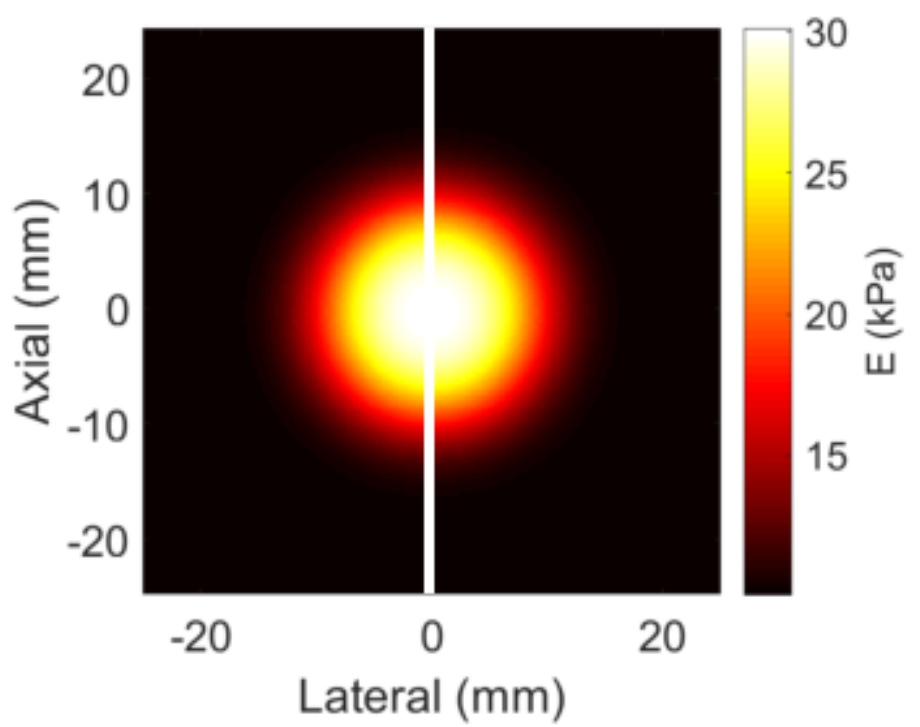}
		\caption{}
	\end{subfigure}
	\hfill
	\begin{subfigure}{0.23\textwidth}
		\includegraphics[width=\textwidth]{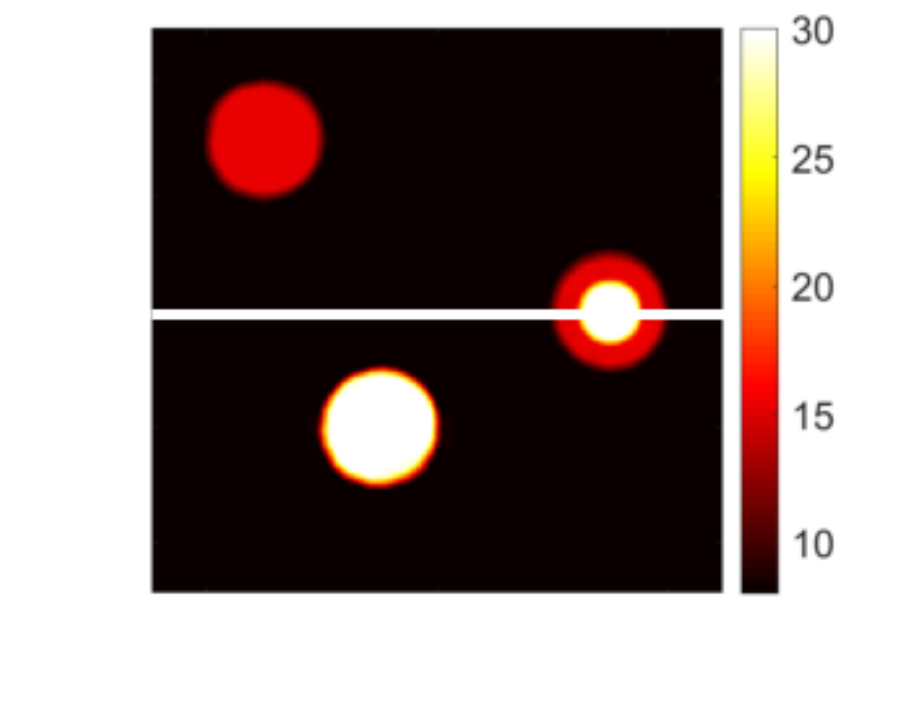}
		\caption{}
		\label{fig:three_line}
	\end{subfigure}
	\begin{subfigure}{0.23\textwidth}
		\includegraphics[width=\textwidth]{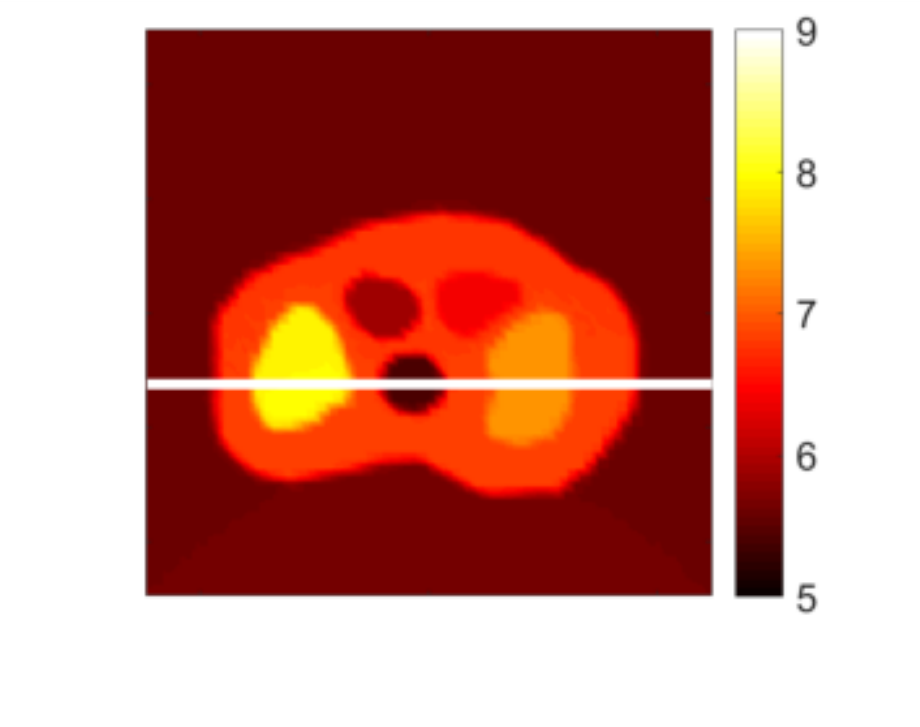}
		\caption{}
		\label{fig:kidney_line}
	\end{subfigure}
	\hfill
	\begin{subfigure}{0.23\textwidth}
		\includegraphics[width=\textwidth]{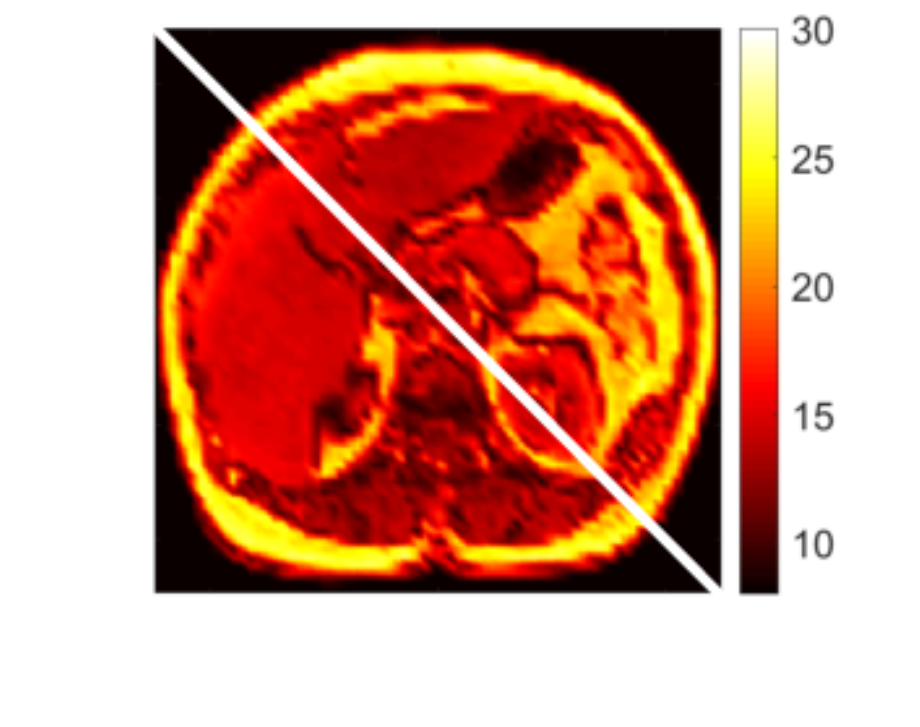}
		\caption{}
		\label{fig:mri_line}
	\end{subfigure}
	\vfill
	\begin{subfigure}{0.23\textwidth}
		\includegraphics[width=\textwidth]{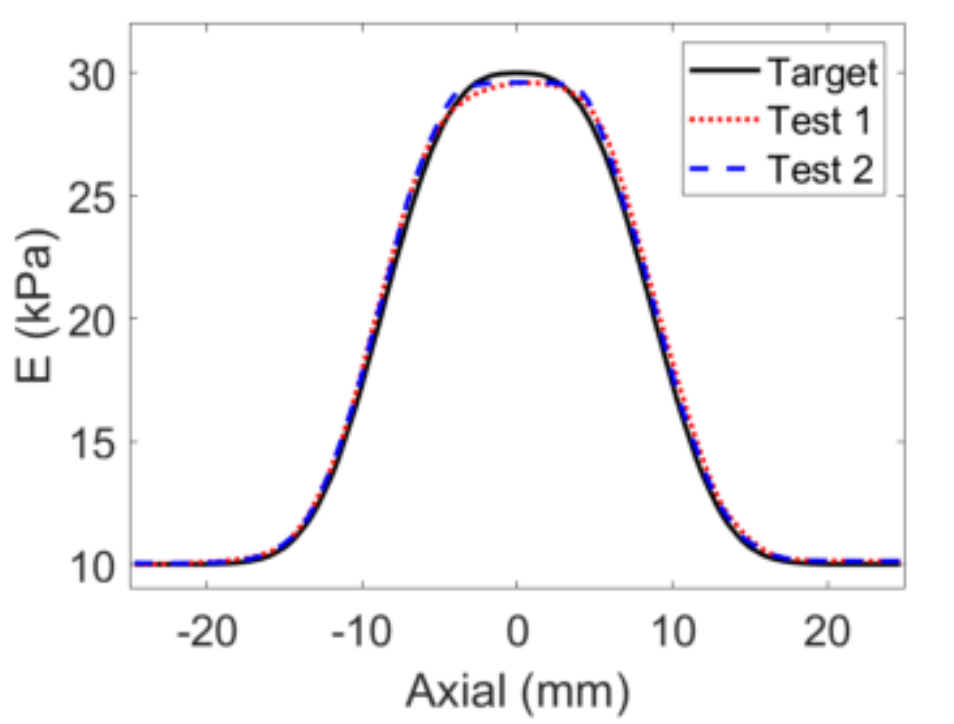}
		\caption{}
	\end{subfigure}
	\hfill
	\begin{subfigure}{0.23\textwidth}
		\includegraphics[width=\textwidth]{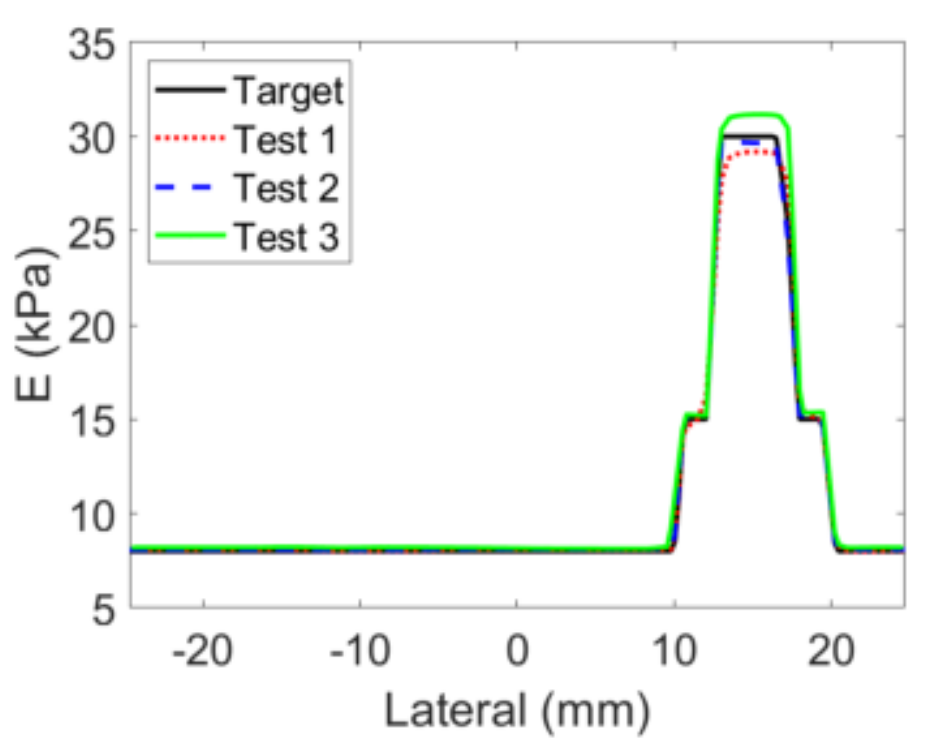}
		\caption{}
		\label{fig:3inc_line_comp}
	\end{subfigure}
	\hfill
	\begin{subfigure}{0.23\textwidth}
		\includegraphics[width=\textwidth]{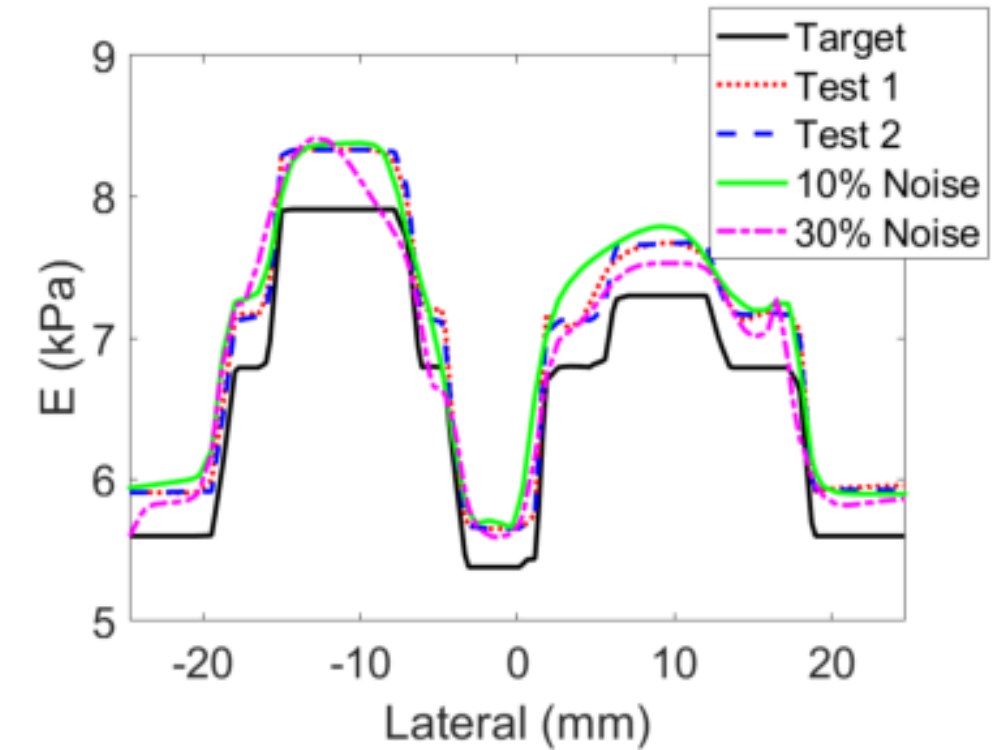}
		\caption{}
		\label{fig:kidney_line_comp}
	\end{subfigure}
	\begin{subfigure}{0.23\textwidth}
		\includegraphics[width=\textwidth]{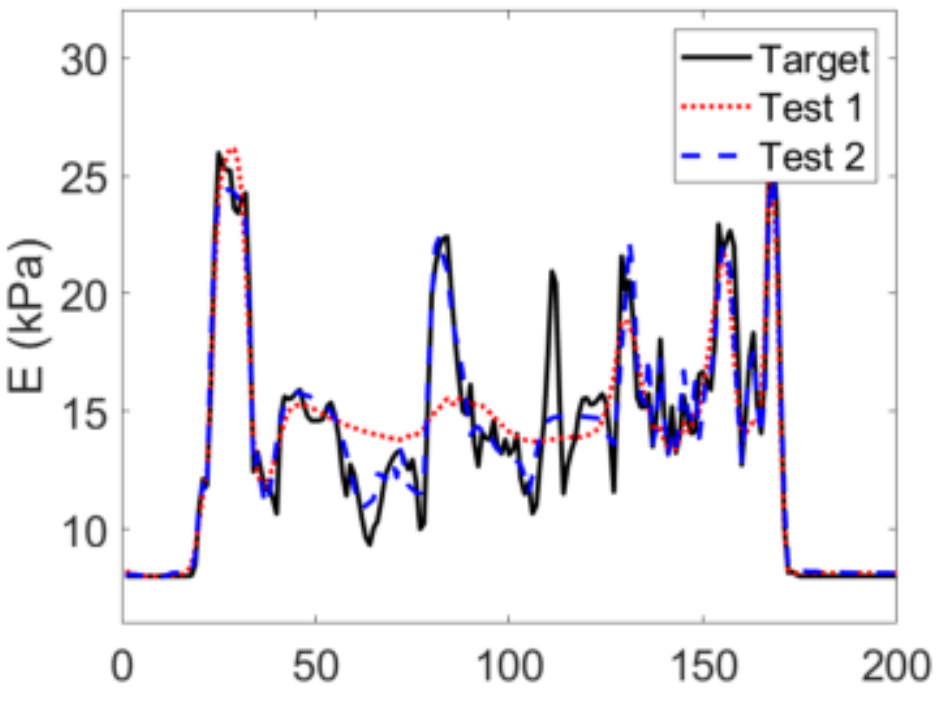}
		\caption{}
		\label{fig:mri_line_comp}
	\end{subfigure}
	\caption{(a-d) The white lines indicate the locations where Young's modulus cross-section comparisons occur. (e) Target and \cann\ Young's modulus along $x = 0$ for single inclusion model. (f) Young's modulus values along $y = 0$ for three inclusion model. Test~3 refers to the case where Mesh~2 and Test~1 parameters were used. (g) Target and \cann\ modulus estimates for no added noise, 10\% added noise, and 30\% added noise along $y = -6.3$ of the kidney model. (h) Young's modulus value for abdominal MRI model along the diagonal.}
	\label{fig:line_comp}
\end{figure}

Young's modulus images reconstructed with \canns\ trained for the four models (no added noise and data generated on Mesh~1) are displayed in the middle and bottom rows of Fig.~\ref{fig:youngs_figs}. A comparison of the target and \cann{-}estimated Young's modulus along cross-sections of these two models are shown in Fig.~\ref{fig:line_comp}. In the case of Model~1, there is no significant difference between the Young's modulus estimates between Test~1 and Test~2. Similarly, there is only a marginal difference in the results for Model~2. The effect of increased training iterations/epochs is far more pronounced for Models~3 and 4. In the case of the former, the boundaries of the regions become sharper. For Model~4, the internal structures only become distinguishable under Test~2 training. 

Images reconstructed by \canns\ are expected to improve when the number of training epochs increases. That Fig.~\ref{fig:cann_2_4} better matches the target distribution than Fig.~\ref{fig:cann_1_4} is no surprise. What these results do show is that 1) \canns\ are capable of learning fairly complex material property distributions and 2) in the absence of noise, the chosen training parameters do not result in over-training the SN.

\begin{figure}[!h]
	\captionsetup[subfigure]{justification=centering}
	\centering
	\begin{subfigure}{0.24\textwidth}
		\includegraphics[width=\textwidth]{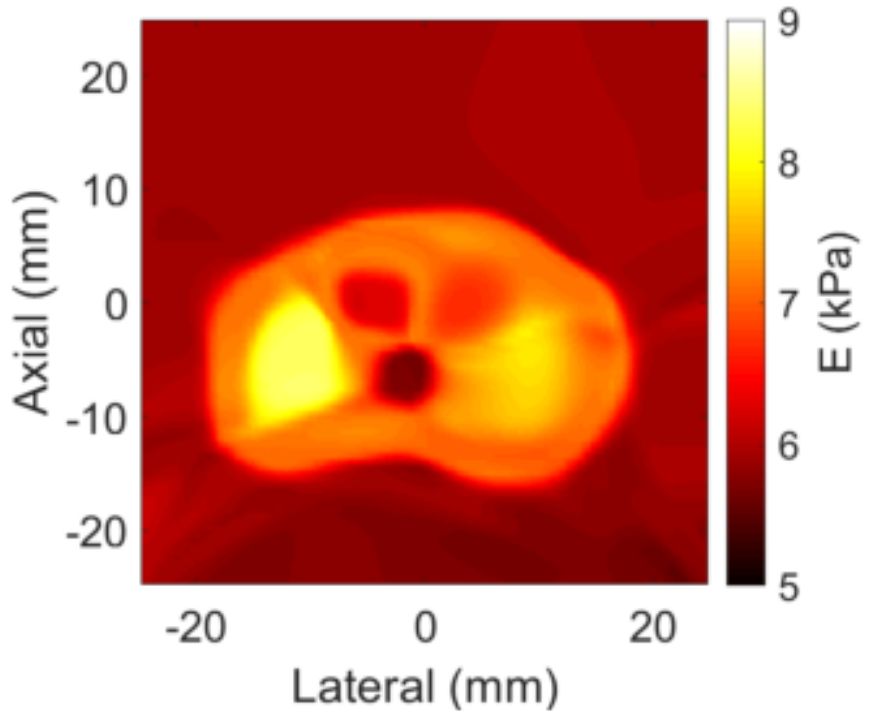}
		\caption{}
		\label{fig:10perc_noise}
	\end{subfigure}
	\hfill
	\begin{subfigure}{0.24\textwidth}
		\includegraphics[width=\textwidth]{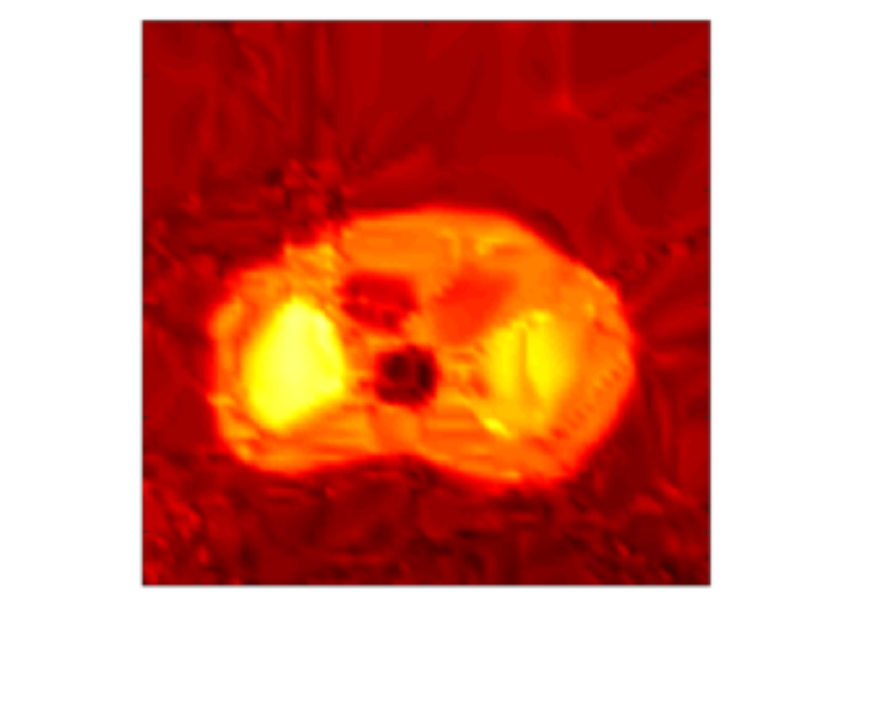}
		\caption{}
		\label{fig:10perc_noise_2}
	\end{subfigure}
	\hfill
	\begin{subfigure}{0.24\textwidth}
		\includegraphics[width=\textwidth]{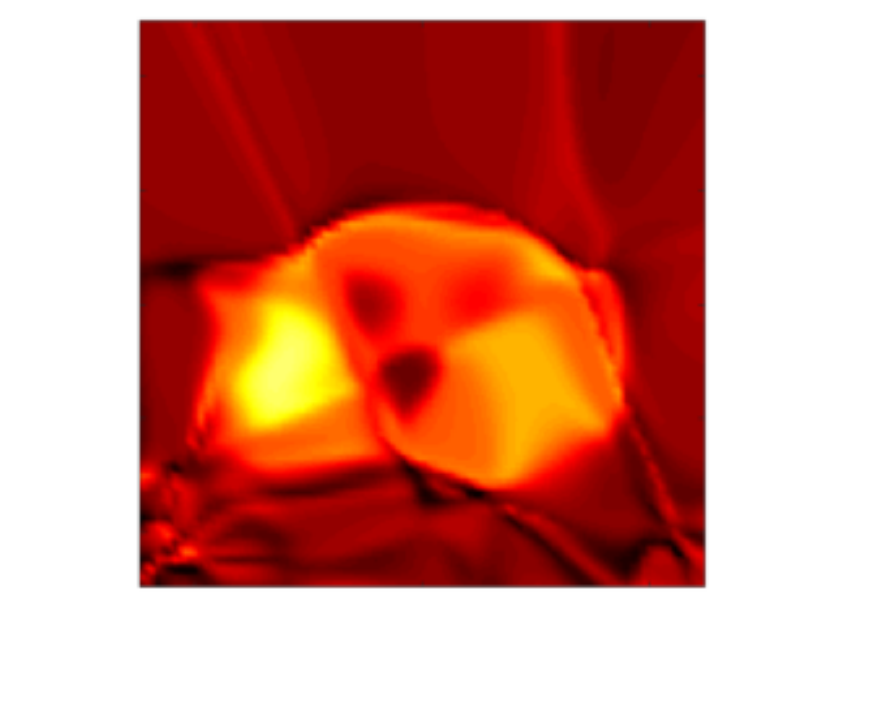}
		\caption{}
		\label{fig:30perc_noise}
	\end{subfigure}
	\hfill
	\begin{subfigure}{0.24\textwidth}
		\includegraphics[width=\textwidth]{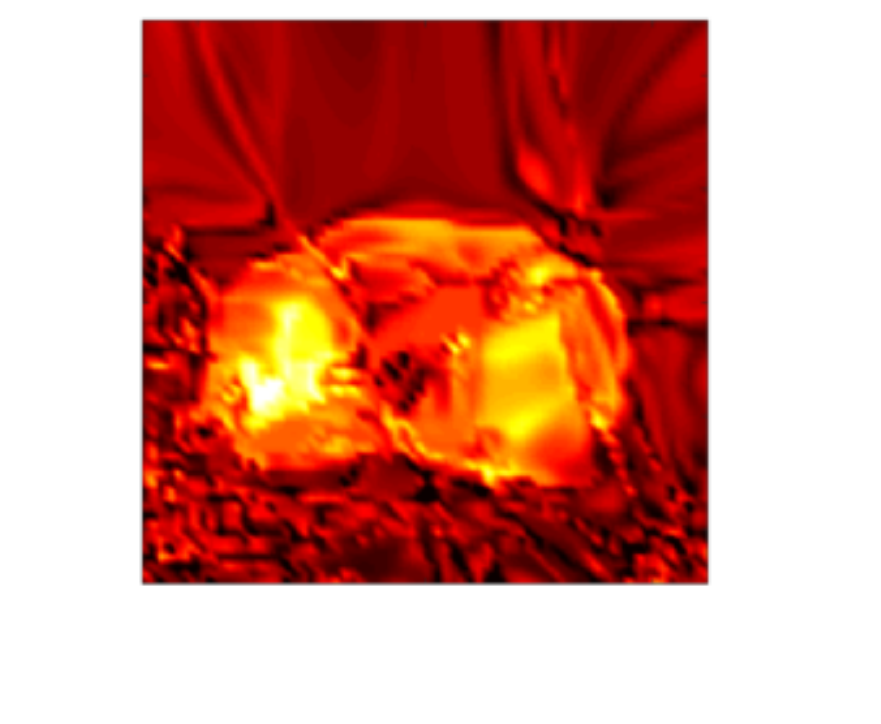}
		\caption{}
		\label{fig:30perc_noise_2}
	\end{subfigure}
	\caption{Young's modulus image after training \cann\ with stress-strain data from kidney model with added noise. (a,b) Test~1 and Test~2 training parameters for model with 10\% added noise, respectively. (c,d) Test~1 and Test~2 training parameters for model with 30\% added noise, respectively.}
	\label{fig:noise_e_maps}
\end{figure}

Image reconstructions with \canns\ trained on data with $10\%$ and $30\%$ additive noise are displayed in Fig.~\ref{fig:noise_e_maps}. Modulus values estimated along the line indicated in Fig.~\ref{fig:kidney_line} by these \canns\ (trained with Test~1 parameters) are included in the curve comparison in Fig.~\ref{fig:kidney_line_comp}. Finally, the Young's modulus values computed by the \cann\ trained on Model~2, Mesh~2, and Test~1 parameters along the line $y = 0$ are included in Fig.~\ref{fig:3inc_line_comp}.

Table~\ref{table:error_table} contains quantitative comparisons between trained \canns\ and the target Young's modulus images. Training times for the SN are also included. For each point $(x,y)$ in the mesh, the error between the target and \cann\ Young's modulus estimate was computed using Eq.~\ref{eqn:youngs_error}.

\begin{flalign}
e^{E}_{(x,y)} = \frac{|E^{target}_{(x,y)} - E^{NN}_{(x,y)}|}{E^{target}_{(x,y)}}\label{eqn:youngs_error}
\end{flalign}
where $E_{(x,y)}^{target}$ is the target Young's modulus value at $(x,y)$ and $E^{NN}_{(x,y)}$ is the modulus estimated by the corresponding trained \cann.

We observe that as the complexity of the model geometry increases and training parameters remain the same, the mean error in the \cann{-}reconstructed Young's modulus image also increases. As expected, increasing the number training iterations and epochs in Test~2 generally reduced the error between target and \cann{-}estimated values. Our expectation was not met for Model~2 and Model~3 with 30\% added noise. The increase in error for Model~2, Test~2 is likely a statistical artefact. Comparing the curves in Fig.~\ref{fig:3inc_line_comp} reveals the \cann\ trained with Test~2 parameters better approximates the target curve. Furthermore, the images were interpolated to a new rectilinear grid before computing the error. It is possible that the interpolation procedure led the the slightly increased error. 

Over-training led to the increased error for Model~3 with 30\% added noise. Comparing Figs.~\ref{fig:30perc_noise} and \ref{fig:30perc_noise_2} reveals the corruption caused by the SN fitting the noise. Similarly, over-training may be also responsible for the increased error in Model~2 when Mesh~2, Test~1 parameters were used. The ``Test 3'' curve in Fig.~\ref{fig:3inc_line_comp} refers to this case and shows a slight bias in the Young's modulus estimate. A particularly large over-estimate occurs for the small, stiff inclusion embedded in the softer inclusion.

Biased errors also occur for Model~3 where all the \cann{-}estimated Young's modulus values in Fig.~\ref{fig:kidney_line_comp} lie above the line denoting target modulus values. The same bias appears in all training cases for this model. While not displayed, this bias can be removed by pretraining the MPN as a 5~kPa linear-elastic material instead of 10~kPa. We do not expect this to be an issue when \canns\ are implemented in AutoP because the MPN may be retrained multiple times. Retraining the MPN with updated stress-strain data should alleviate these types of issues by adjusting the ``reference'' material to a form more suitable for the generated data.

\begin{table}[!h]
	\caption{Young's modulus errors and SN training times for the four models shown in Fig.~\ref{fig:youngs_figs}.
		No asterisk indicates Mesh 1, Test 1 parameters whereas single asterisk ($^\ast$) indicates Test 2 parameters. The double asterisk ($^{\ast\ast}$) specifies the use of Mesh 2, Test 1 parameters .}
	\centering
	\begin{tabular}{|l|c|c|}
		\hline
		\multicolumn{1}{|c|}{\multirow{2}{*}{Model}} & \multicolumn{1}{|c|}{Modulus Error} &
		{\multirow{2}{*}{\shortstack{Training \\Time (s)}}}\\
		\cline{2-2}
		& Mean $\pm$ STD &  \\ \hline
		1 & $0.0166 \pm 0.0099$ & 71 \\ 
		1$^\ast$ & $0.0146 \pm 0.0076$ & 271 \\ \hline
		
		2 & $0.0131 \pm 0.0172$ & 71  \\
		
		2$^\ast$ & $0.0140 \pm 0.0190$ & 271  \\
		
		2.$^{\ast\ast}$  & $0.0258 \pm 0.0334$ & 56 \\

		\hline		 
		
		3 & $0.0504 \pm 0.0142$ & 72  \\
		3$^\ast$ & $0.0493 \pm 0.0131$ & 272 \\
		
		3 ($10\%$ noise) & $0.0534 \pm 0.0198$ & 71  \\ 
		3$^\ast$ ($10\%$ noise) & $0.0487 \pm 0.0230$ & 272 \\ 
		
		3 ($30\%$ noise) & $0.0418 \pm 0.0252$ & 71  \\ 
		3$^\ast$ ($30\%$ noise) & $0.0549 \pm 0.0399$ & 271  \\ \hline
		
		4 & $0.0658 \pm 0.0755$ & 71  \\
		4$^\ast$  & $0.0485 \pm 0.0518$ & 270   \\ \hline
		
	\end{tabular}\label{table:error_table}
\end{table}

\section{Discussion}
We have demonstrated the ability of \canns\ to learn both the material properties and geometry of 2-D linear-elastic, isotropic materials under plane-stress loading when the stress-strain data are known. \canns\ utilize two cooperating NNs to minimize the error between known and predicted stresses after a forward propagation of a strain vector and a $(x,y)$ coordinate. The material property network learns a general stress-strain relation --- here, the linear-elastic response of a 10 kPa homogeneous material --- whereas the spatial network learns a map of relative stiffness. Developing the two-network cooperating architecture followed from attempts to simply add two extra inputs to the material property network for (scaled) Cartesian x- and y-coordinates. These early \canns\ performed poorly and were unable to accurately learn geometric information regardless of the number of hidden layers, nodes per layer, and/or training parameters. We believe the failure was caused in large part by an incompatibility between the hyperbolic tangent activation function and the geometry-material property relationship. In short, a hyperbolic tangent is symmetric in sign so that a negative valued input produces a negative valued output. Use of this function for the MPN works because stresses and strains exhibit the same relationship. Furthermore, a zero-valued input vector should produce a zero-valued output vector (i.e., zero strain means zero stress). But, the same does not hold for spatial location and material property. Reformulating the problem as adjusting a reference material property learned by the MPN with some auxiliary function led to incorporating the spatial network.

Adding the spatial network required developing a method of computing its outputs. We use a gradient-descent based approach by backpropagating the output error all the way to the MPN inputs/spatial network outputs. The spatial scaling values effectively encode stiffness relative to the reference material property. For example, $S_{\bm{x}}^{\varepsilon_2}$ calculated for Model~1 is $\approx 0.3657$ at the 30 kPa inclusion peak and $\approx 1.1$ for the 10 kPa background, maintaining the ratio $30/10 = 1/3 \approx 1.1/0.3657$. Similarly, Model 4 values span 8-30 kPa and the resulting $S_{\bm{x}}^{\varepsilon_2}$ are in the range $[0.3891,\ 1.4014]$. We must emphasize that the spatial scaling values were not specifically computed to behave in this manner; this property emerged from minimizing the error function defined in Eq.~\ref{eq:loss_func}.

From our previous study, we found only two hidden layers were necessary for the material property network to learn a linear-elastic relationship for 2-D and 3-D materials. Spatial networks, though, require a larger network. We chose to increase the number of hidden layers instead of vastly increasing the number of nodes per layer. The spatial network must be larger because the mapping from $(x,y)$ to \spatscale\ is more complicated than a linear stress-strain relationship and thus requires a network with larger capacity. While there is no strict rule for determining NN size, five hidden layers comprised of 25 nodes each was sufficient for all four models in this study. However, the larger size of the SN increases the risk of over-training, as observed in Model~2 with 30\% added noise.

Keeping the spatial network size constant, we could improve the Young's modulus reconstruction by changing the training parameters or the mesh. In the former case, increasing the number of epochs and training iterations produced better results by simply allowing more weight updates to occur. But increasing training epochs is not the best choice if training time is to be minimized. For the cases where Test 1 parameters were used, training time was $\approx 71$s on a quad-core CPU operating at 2.7 GHz. Test 2 parameters increased the time to $\approx 271$s on the same computer.

Conversely, changing from Mesh 1 to Mesh 2 and resorting back to Test 1 parameters only required $\approx 56$s of training time. Changing the mesh reduced the total number of data points from 4624 to 940, in turn reducing the training time. However, maintaining the same number of training epochs for the reduced amount of training data did not improve the resulting Young's modulus estimates. There was arguably a qualitative improvement due to changes in the data sampling distribution. With the rectilinear mesh, the edges of the inclusions are coarsely sampled and the number of data points pulled from said inclusions are small compared to the soft background material. Changing the mesh altered this sampling distribution so that data are better sampled around edges. 

Finally, while the addition of noise does not appear to significantly affect the ability of \canns\ to learn material properties, the geometry is corrupted. Unfortunately, it is difficult to extrapolate these results to AutoP because the noise will appear in force-displacement measurements and then propagate through the stress-strain calculations in non-straightforward ways. Implementation in AutoP paper will reveal how robust \canns\ are to the noise encountered in real measurement data.

\section{Conclusion}
Cartesian neural network constitutive models can simultaneously learn material property and geometric information. Unlike previous machine-learning methods, \canns\ are able to capture continuous material property distributions. Furthermore, these networks can resolve fine structures with minor adjustments to training or the finite element mesh, the latter of which changes the distribution of available training data. \canns\ are fairly robust to noise and can still produce accurate estimates of Young's modulus for linear-elastic materials at the cost geometric distortion. \canns\ are a novel NN architecture that, once incorporated into the Autoprogressive Method, will offer a machine-learning alternative to elasticity imaging.


%

\section*{Acknowledgment}
Research reported in this publication was supported by NCI and NIBIB of the National
Institutes of Health under Award Numbers R01 CA168575 and R21 EB023402. The content is solely the responsibility
of the authors and does not necessarily represent the official views of the National Institutes of
Health. 


\bibliographystyle{abbrv}
\bibliography{references,references_autop}
\newpage
\section*{Special Terms and Acronyms}
\begin{description}[labelwidth=0.8cm]
	\item[$\bm{\sigma}_{\rm i}$] $i$th stress vector
	\item[$\sigma_{\rm i}$] $i$th component of stress vector
	\item[$\bm{\sigma}^t$] Target stress computed in FEA
	\item[$\bm{\sigma}^{NN}$] Stress computed by MPN from an input strain vector
	\item[$\hat{\bm{\sigma}}_j^{NN}$] Stress vector computed by MPN from an input strain vector with all $j\neq k$ components equal to zero
	\item[\spatscale] The spatial scaling vector at $\bm{x}$
	\item[$\SK$] The $k$th component of \spatscale
	\item[MPN] Material property network
	\item[SN] Spatial network
	\item[Test 1] RPROP training over 10/300 iterations/epochs
	\item[Test 2] RPROP training over 30/600 iterations/epochs
\end{description}

\section*{Appendix A}
\setcounter{figure}{0}
\renewcommand{\thefigure}{S\arabic{figure}}

\begin{figure}
	\captionsetup[subfigure]{justification=centering}
	\centering
	\begin{subfigure}{0.40\textwidth}
		\includegraphics[width=\textwidth]{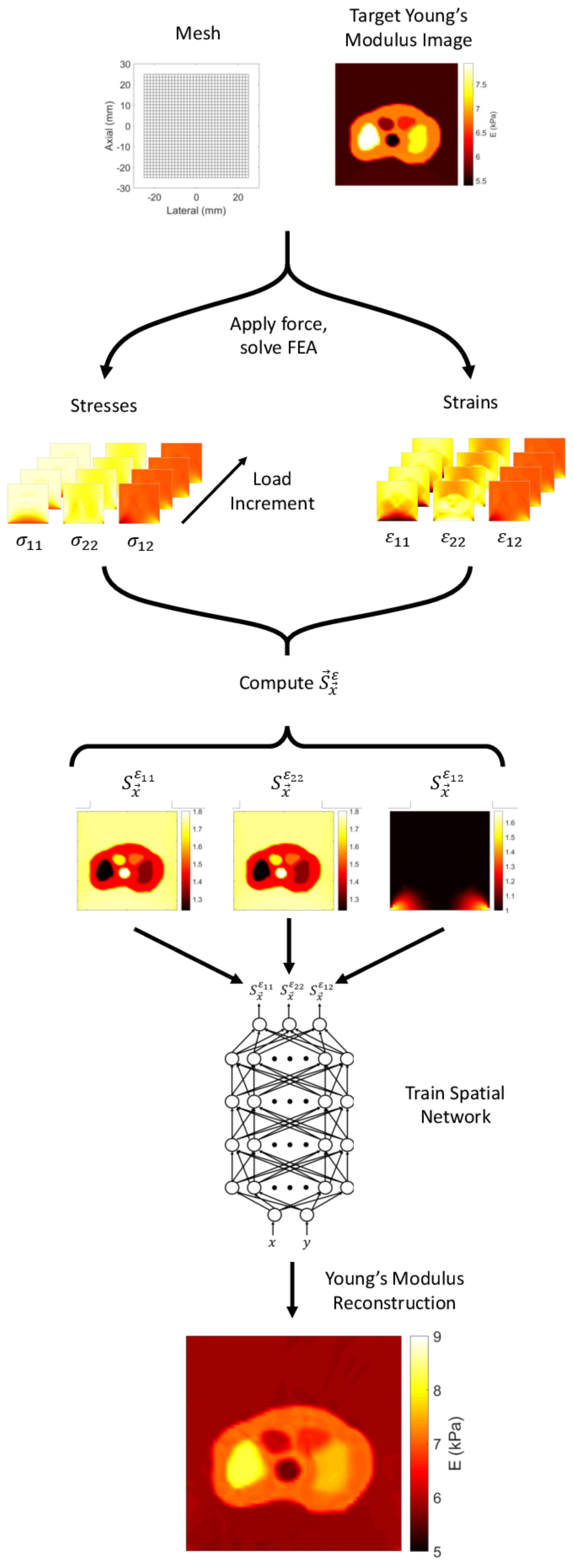}
		\caption{}
		\label{fig:app_method1}
	\end{subfigure}
	\hfill
	\begin{subfigure}{0.42\textwidth}
		\includegraphics[width=\textwidth]{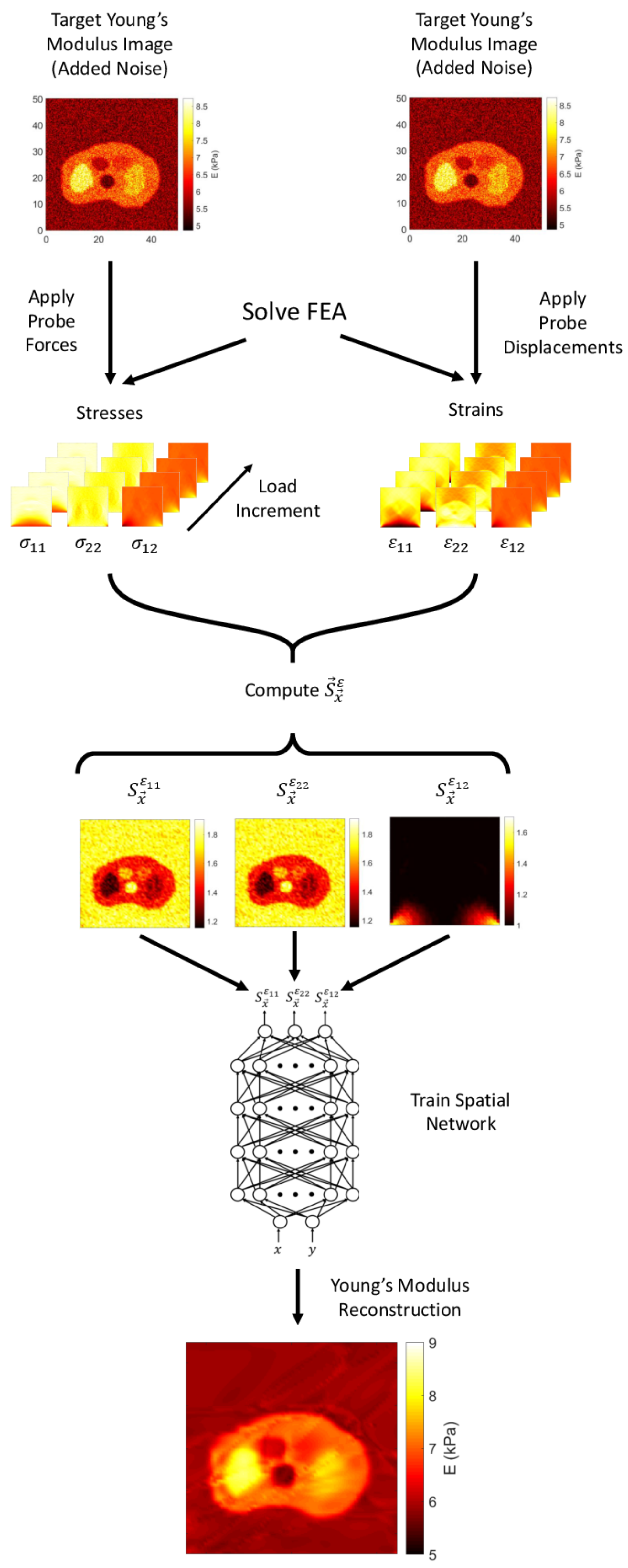}
		\caption{}
		\label{fig:app_method2}
	\end{subfigure}
	\caption{Diagram of experimental method. (a) No noise added to target Young's modulus distribution. (b) Uniformly distributed noise added to target distribution.}
	\label{fig:app_1}
\end{figure}

Figures~\ref{fig:app_method1} and \ref{fig:app_method2} are provided to clarify the experimental procedure. When no noise was added to the target Young's modulus distribution, stresses and strains were both computed in a single FEA. Spatial scaling values were then computed using Alg.~1 and the pre-trained MPN. Conversely, two separate FEAs were solved to compute stresses and strains separately when noise was present. Noise was added by sampling from a zero-mean uniform distribution with maximum magnitude equal to either 10\% or 30\% of the Young's modulus at each $\bm{x}$. For example, when 10\% noise was added, the target Young's modulus at each point was $E'(x,y) = E(x,y) + E(x,y)*p$, where $p~\in~[-0.1*E(x,y),~0.1*E(x,y)]$. Stresses and strains were computed in separate FEAs to avoid correlating the noise (i.e., the same value $p$ was not used to compute $E'(x,y)$ at each location).

\end{document}